\documentclass{article}

\usepackage[nonatbib,final]{neurips_2020}

\usepackage[utf8]{inputenc} 
\usepackage[T1]{fontenc}    
\usepackage{hyperref}       
\usepackage{url}            
\usepackage{booktabs}       
\usepackage{amsfonts}       
\usepackage{nicefrac}       
\usepackage{microtype}      
\usepackage{comment}
\usepackage{graphicx}
\usepackage{amsmath}
\usepackage{amssymb}
\usepackage{mathtools}
\usepackage{xcolor}
\usepackage{mynotation}
\usepackage{float}
\usepackage{algorithm}
\usepackage{algpseudocode}
\usepackage{graphicx,wrapfig,lipsum}
\usepackage[caption=false]{subfig}

\newcommand{\parsection}[1]{\noindent\textbf{#1:} }

\usepackage{xspace}
\makeatletter
\DeclareRobustCommand\onedot{\futurelet\@let@token\@onedot}
\def\@onedot{\ifx\@let@token.\else.\null\fi\xspace}

\def\eg{\emph{e.g}\onedot} 
\def\ie{\emph{i.e}\onedot}

\def\wrt{w.r.t\onedot} 
\def\etal{\emph{et al}\onedot}
\def\dicor{GOCor\xspace}
\makeatother

\graphicspath{{../}}

\title{GOCor: Bringing Globally Optimized Correspondence Volumes into Your Neural Network}

\author{%
    Prune Truong\thanks{Both authors contributed equally} \ , Martin Danelljan$^*$, Luc Van Gool, Radu Timofte\\ 
    \texttt{\{prune.truong, martin.danelljan, vangool, radu.timofte\}@vision.ee.ethz.ch} \\
    Computer Vision Lab, ETH Zurich, Switzerland\\
}

\begin{document}

\maketitle
\vspace{-3mm}
\begin{abstract}
    
The feature correlation layer serves as a key neural network module in numerous computer vision problems that involve dense correspondences between image pairs. It predicts a correspondence volume by evaluating dense scalar products between feature vectors extracted from pairs of locations in two images.
However, this point-to-point feature comparison is insufficient when disambiguating multiple similar regions in an image, severely affecting the performance of the end task.
We propose GOCor, a fully differentiable dense matching module, acting as a direct replacement to the feature correlation layer.
The correspondence volume generated by our module is the result of an internal optimization procedure that explicitly accounts for similar regions in the scene. Moreover, our approach is capable of effectively learning spatial matching priors to resolve further matching ambiguities.
We analyze our GOCor module in extensive ablative experiments. When integrated into state-of-the-art networks, our approach significantly outperforms the feature correlation layer for the tasks of geometric matching, optical flow, and dense semantic matching. The code and trained models will be made available at \mbox{\url{github.com/PruneTruong/GOCor}}. 

\end{abstract}

\vspace{-4mm}
\section{Introduction}

Finding pixel-wise correspondences between pairs of images is a fundamental problem in many computer vision domains, including optical flow~\cite{Dosovitskiy2015,Hui2018,Ilg2017a, Sun2018,  Sun2019,GLUNet}, geometric matching~\cite{DeToneMR16, Melekhov2019,Rocco2017a, Rocco2018a,  GLUNet}, and disparity estimation~\cite{ChenS0YH15, LiangFGLCQZZ18, Pang2017, ZhangKRVKTSIFF18}. Most recent state-of-the-art approaches rely on feature correlation layers, evaluating dense pair-wise similarities between deep representations of two images. The resulting four-dimensional \emph{correspondence volume} captures dense \emph{matching confidences} between every pair of image locations. It serves as a powerful cue in the prediction of, for instance, optical flow. This encapsulation of dense correspondences has further achieved wide success within semantic matching~\cite{ChenWLF19,Choy2016,DCCNet,Jeon,Kim2018, KimMHLS19,  Kim2019,Rocco2018b, GLUNet}, video object segmentation~\cite{blazingly, VM, STM,FEELVOS}, and few-shot segmentation~\cite{PRNet, WangLZZF19}. The feature correlation layer thus serves as a key building block when designing network architectures for a diverse range of important computer vision applications.

\begin{figure}[t]
\centering%
\newcommand{\wid}{0.193\textwidth}
\vspace{-6mm}
\subfloat[Reference image]{\includegraphics[width=\wid]{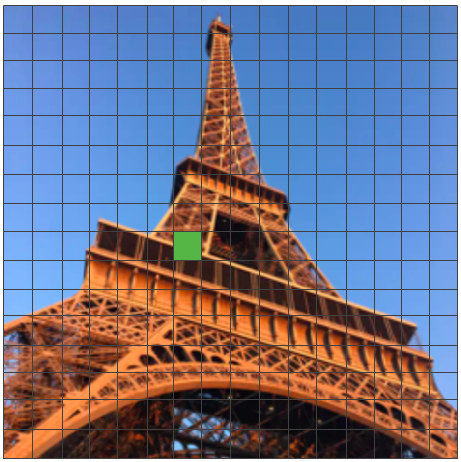}}~%
\subfloat[Query image]{\includegraphics[width=\wid]{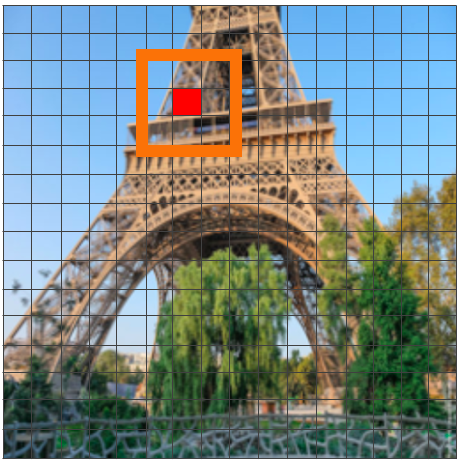}}~%
\subfloat[Ideal Correlation]{\includegraphics[width=\wid]{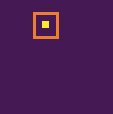}}~%
\subfloat[Feat.\ Correlation\label{fig:feat-corr}]{\includegraphics[width=\wid]{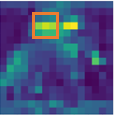}}~%
\subfloat[\dicor (Ours)\label{fig:intro-ours}]{\includegraphics[width=\wid]{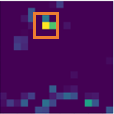}} 
\vspace{-2mm}
\caption{Visualization of the matching confidences (c)-(e) computed between the indicated location (green) in the reference image (a) and all locations of the query image (b). The feature correlation (d) generates undistinctive and inaccurate confidences due to similar regions and repetitive patterns. In contrast, our \dicor (e) predicts a distinct high-confidence value at the correct location.}
\label{fig:intro-corr}\vspace{-6mm}
\end{figure}

In the feature correlation layer, each confidence value in the correspondence volume is obtained as the scalar product between two feature vectors, extracted from specific locations in the two images, here called the \emph{reference} and the \emph{query} images.
However, the sole reliance on point-to-point feature comparisons is often insufficient in order to disambiguate multiple similar regions in an image. As illustrated in Fig.~\ref{fig:intro-corr}, in the case of repetitive patterns, the feature correlation layer generates undistinctive and inaccurate matching confidences (Fig.~\ref{fig:feat-corr}), severely affecting the performance of the end task. This remains the key limitation of feature correlation layers, since repetitive patterns, low-textured regions, and co-occurring similar objects are all pervasive in computer vision applications.

We design a new dense matching module, aiming to address the aforementioned issues by exploring information not exploited by the feature correlation layer.
We observe that a confidence value in the correspondence volume generated by the feature correlation layer only depends on the feature vectors extracted at one pair of locations in the reference and query.
However, the reference also contains the appearance information of other image locations, that are likely to occur in the query image. This includes the appearance of similar regions in the scene, opening the opportunity to actively identify and account for such similarities when estimating each matching confidence value.
Moreover, the feature correlation layer ignores prior knowledge and constraints that can be derived from the query, \eg the uniqueness and spatial smoothness of correspondences. 
Our matching module encapsulates the aforementioned information and constraints into a learnable objective function. Our enhanced correspondence volume is obtained by minimizing this objective during the forward-pass of the network. This allows us to predict \emph{globally optimised} correspondence volumes, effectively accounting for similar image regions and matching constraints, as visualized in Fig.~\ref{fig:intro-ours}.

\newcommand{\bp}[1]{\textbf{#1}}

\parsection{Contributions} We introduce \dicor, a differentiable neural network module that generates the correspondence volume between a pair of images, acting as a direct replacement to the feature correlation layer.
Our main contributions are as follows.
\bp{(i)} Our module is formulated as an internal optimization procedure that minimizes a customizable matching-objective during inference, thereby providing a general framework for effectively integrating both explicit and learnable matching constraints.
\bp{(ii)} We propose a robust objective that integrates information about similar regions in the scene, allowing our \dicor module to better disambiguate such cases.
\bp{(iii)} We introduce a learnable objective for capturing constraints and prior information about the query frame.
\bp{(iv)} We apply effective unrolled optimization, paired with accurate initialization, ensuring efficient end-to-end training and inference.
\bp{(v)} We perform extensive experiments on the geometric matching and optical flow tasks by integrating our module into state-of-the-art matching network architectures. We additionally show that our \dicor module can generalize to the dense semantic matching task. 
Our approach outperforms the feature correlation layer in terms of both accuracy and robustness. In particular, our \dicor module demonstrates better domain generalization properties. 

\section{Related work}

\parsection{Enhancing the correlation volume} 
Since the quality of the correspondence volume is of prime importance, several works focus on improving it using learned post-processing techniques \cite{Laskar2019, Shuda2020, Rocco2018b, YangR19}. Notably, Rocco~\etal~\cite{Rocco2018b} proposed a trainable neighborhood consensus network, NC-Net, applied after the correlation layer to filter out ambiguous matches. 
Instead, we propose a fundamentally different approach, operating directly on the underlying feature maps, \emph{before} the correlation operation.  
Our work is also related to~\cite{JiaBTG16, SarlinDMR20}, which generate filters dynamically conditioned on an input~\cite{JiaBTG16} or features updated with an attentional graph neural network, whose edges are defined within the same or the other image of a pair~\cite{SarlinDMR20}. Xiao~\etal~\cite{Xiao2020} also recently introduced a learnable cost volume that adapts the features to an elliptical inner product space.

\vspace{-1mm}
\parsection{Optimization-based meta-learning}
Our approach is related to optimization-based meta-learning \cite{r2d2,dimp,lwtl,metaoptnet,cfnet,cavia}. In fact, our \dicor module can be seen as an internal learner, which solves the regression problem defined by our objective. In particular, we adopt the steepest descent based optimization strategy shown effective in \cite{dimp,lwtl}. From a meta-learning viewpoint, our approach however offers a few interesting additions to the standard setting. Unlike for instance, in few-shot classification \cite{r2d2,metaoptnet,cavia} and tracking \cite{dimp,cfnet}, our learner constitutes an internal network module of a larger architecture. This implies that the output of the learner does not correspond to the final network output, and therefore does not receive direct supervision during (meta-)training. Lastly, our learner module actively utilizes the query sample through the introduced trainable objective function. 
\section{Method}
\label{sec:method}

\subsection{Feature Correlation Layers}
\label{subsec:loc-global}

The feature correlation layer has become a key building block in the design of neural network architectures for a variety of computer vision tasks, which either rely on or benefit from the estimation of dense correspondences between two images. To this end, the feature correlation layer computes a dense set of scalar products between localized deep feature vectors extracted from the two images, in the form of a four-dimensional \emph{correspondence volume}. 
We consider two deep feature maps $\ftr = \phi(\imr)$ and $\ftq = \phi(\imq)$ extracted by a deep CNN $\phi$ from the \emph{reference} image $\imr$ and the \emph{query} image $\imq$, respectively. The feature maps $\ftr, \ftq \in \reals^{H \times W \times D}$ have a spatial size of $H\times W$ and dimensionality $D$. 
We let $\ftr_{ij} \in \reals^D$ denote the feature vector at a spatial location $(i,j)$. The feature correlation layer evaluates scalar products $(\ftr_{ij})\tp \ftq_{kl}$ between the reference and query image representations. There are two common variants of the correlation layer, both relying on the same local scalar product operation, but with some important differences. We define these operations next.

The \textbf{Global correlation layer} evaluates the pairwise similarities between all locations in the reference and query feature maps. This is defined as the operation,
\begin{equation}
\label{eq:global-corr}
\gcorr(\ftr, \ftq)_{ijkl} = (\ftr_{ij})\tp \ftq_{kl} \,,\quad (i,j), (k,l) \in \{1,\ldots,H\}\times\{1,\ldots,W\} \,. 
\end{equation}
The result is thus a 4D tensor $\gcorr(\ftr, \ftq) \in \reals^{H \times W \times H \times W}$ capturing the similarities between all pairs of spatial locations.
In the \textbf{Local correlation layer}, the scalar products involving $\ftr_{ij}$ are instead only evaluated in a neighborhood of the location $(i,j)$ in the query feature map $\ftq$, 
\begin{equation}
\label{eq:local-corr}
\lcorr(\ftr, \ftq)_{ijkl} = (\ftr_{ij})\tp \ftq_{i+k,j+l} \,,\;  (i,j) \in \{1,\ldots,H\}\!\times\!\{1,\ldots,W\}, \; (k,l) \in \{-R,\ldots,R\}^2 .
\end{equation}
$(k,l)$ represents the displacement relative to the reference frame location $(i,j)$, constrained to a value within the search radius $R$. 
While the limited search region $R$ makes the local correlation practical even for feature maps of a large spatial size $H \times W$, it does not capture similarities beyond $R$. 

\subsection{Motivation}
\label{subsec:motivation}

The main purpose of feature correlation layers is to predict a dense set of matching confidences between the two images $\imr$ and $\imq$. This is performed in \eqref{eq:global-corr}-\eqref{eq:local-corr} by applying each reference frame feature vector $\ftr_{ij}$ to a region in the query $\ftq$. However, this operation ignores two important sources of valuable information when establishing dense correspondences. 

\parsection{Reference frame information } The matching confidences $\corr(\ftr, \ftq)_{ij..} \in \reals^{H \times W}$ (in \ref{eq:global-corr}-\ref{eq:local-corr})
for the reference image location $(i,j)$ does not account for the appearance at other locations of the reference image. Instead, it only depends on the feature vector $\ftr_{ij}$ at the location itself. This is particularly problematic when the reference frame contains multiple locations with similar appearance, such as repetitive patterns or homogeneous regions (see Fig.~\ref{fig:intro-corr}). These regions are also very likely to occur in the query feature map $\ftq$, since it usually depicts the same scene at a later time instance or from a different viewpoint. 
This easily results in high correlation values at multiple incorrect locations, often severely affecting the accuracy and robustness of the final network prediction. Unfortunately, patterns of similar appearance are almost ubiquitous in natural scenes.
Therefore, the estimation of matching confidences should ideally exploit the \emph{known similarities in the reference image itself}.

\parsection{Query frame information } The second source of information not exploited by the feature correlation layer is matching constraints and priors that can be derived from the query $\ftq$. One such important constraint is that each reference image location $\ftr_{ij}$ can have at most one matching location $\ftq_{kl}$ in the query image. Moreover, dense matches across the image pair generally follow spatial smoothness properties, due to the spatio-temporal continuity of the underlying 3D scene. This can serve as a powerful prior when predicting the correspondence volume between the image pair. 

Next, we set out to develop a dense matching module capable of effectively utilizing the aforementioned information when predicting the correspondence volume relating $I^r$ and $I^q$.

\subsection{General Formulation}
\label{subsec:formulation}

\begin{figure*}[t]
\centering
\includegraphics[width=0.999\textwidth]{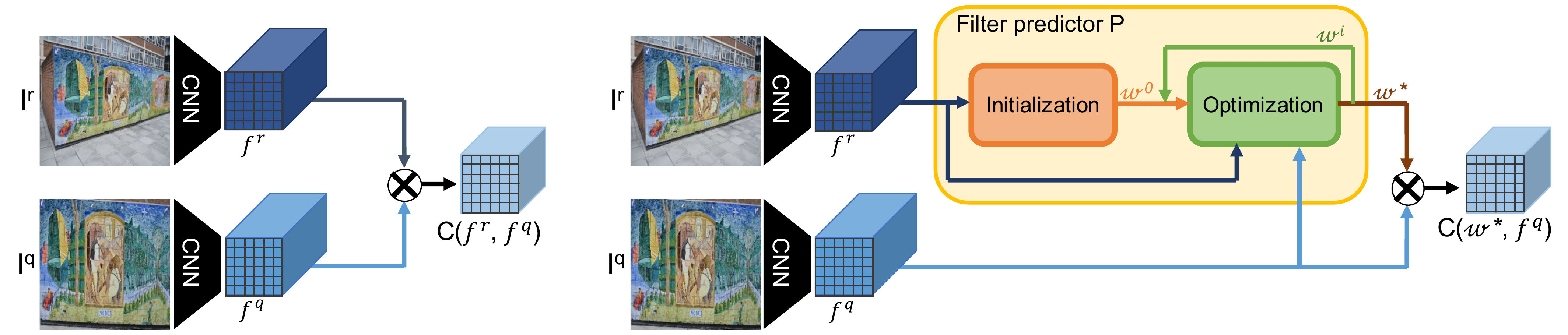}\vspace{-5mm}
\subfloat[Feature correlation layer\label{fig:arch-featcor}]{\hspace{0.4\textwidth}}%
\subfloat[\dicor feature correlation (Ours)\label{fig:arch-dicor}]{\hspace{0.6\textwidth}}\vspace{-3mm}
\caption{Schematic overview of the the feature correlation layer (a) and our \dicor module (b).}\vspace{-5mm}
\label{fig:arch}
\end{figure*}

In this section, we formulate \dicor, an end-to-end differentiable neural network module capable of generating more accurate correspondence volumes than feature correlation layers. We start by replacing the reference feature map $\ftr$ in \eqref{eq:global-corr}-\eqref{eq:local-corr} with a general tensor $\wtm$ of the same size, which we refer to as the \emph{filter map}. Instead of correlating the reference features $\ftr$ with the query $\ftq$, we aim to first predict the filter map $\wtm$, enriched with the global information about the reference $\ftr$ and query $\ftq$ described in the previous section. The filter map $\wtm$ is then applied to the query features $\ftq$ to obtain the final correspondence volume as $\corr(\wtm, \ftq)$. We use $\corr$ to denote either global \eqref{eq:global-corr} or local \eqref{eq:local-corr} correlation. We thus embrace the correlation operation \eqref{eq:global-corr}-\eqref{eq:local-corr} itself, and aim to enhance its output by enriching its input.

The remaining part of our method description is dedicated to the key question raised by the above generalization, namely how to achieve a suitable filter map $\wtm$. In general, we can consider it to be the result of a differentiable function $\wtm = \wpred_\theta(\ftr, \ftq)$, which takes the reference and query features as input and has a set of trainable parameters $\theta$. For example, simply letting $\wpred_\theta(\ftr, \ftq) = \ftr$ retrieves the original feature correlation layer $\corr(\ftr, \ftq)$. However, designing a neural network module $\wtm = \wpred_\theta(\ftr, \ftq)$ that \emph{effectively} takes advantage of the information and constraints discussed in Sec.~\ref{subsec:motivation} is challenging.
Moreover, we require our module to robustly generalize to new domains, having image content and motion patterns not seen during training. 

We tackle these challenges by formulating an objective function $L$, that explicitly encodes the constraints discussed in Sec.~\ref{subsec:motivation}.
The network module $\wpred_\theta(\ftr, \ftq)$ is then constructed to output the filter map $\wtm$ that minimizes this objective,
\begin{equation}
\label{eq:minimize-loss}
    \wtm = \wpred_\theta(\ftr, \ftq) = \argmin_\wt L(\wt; \ftr, \ftq, \theta) \,.
\end{equation}
This formulation allows us to construct the filter predictor module $\wpred_\theta$ by designing an objective $L$ along with a suitable optimization algorithm. It gives us a powerful framework to explicitly integrate the constraints discussed in Sec.~\ref{subsec:motivation}, while also benefiting from significant interpretability. In the next sections, we formulate our objective function $L$. We first integrate information about the reference features $\ftr$ into the objective \eqref{eq:minimize-loss} in Sec.~\ref{subsec:loss-ref}. In Sec.~\ref{subsec:smooth-loss}, we then extend the objective $L$ with information about the query $\ftq$. Lastly, we discuss the optimization procedure applied to our objective in Sec.~\ref{subsec:optim}. An overview of our general matching module is illustrated in Figure~\ref{fig:arch}.

\subsection{Reference Frame Objective}
\label{subsec:loss-ref}

\begin{wrapfigure}{r}{5.5 cm}
\vspace{-12mm}\includegraphics[width=5.5 cm]{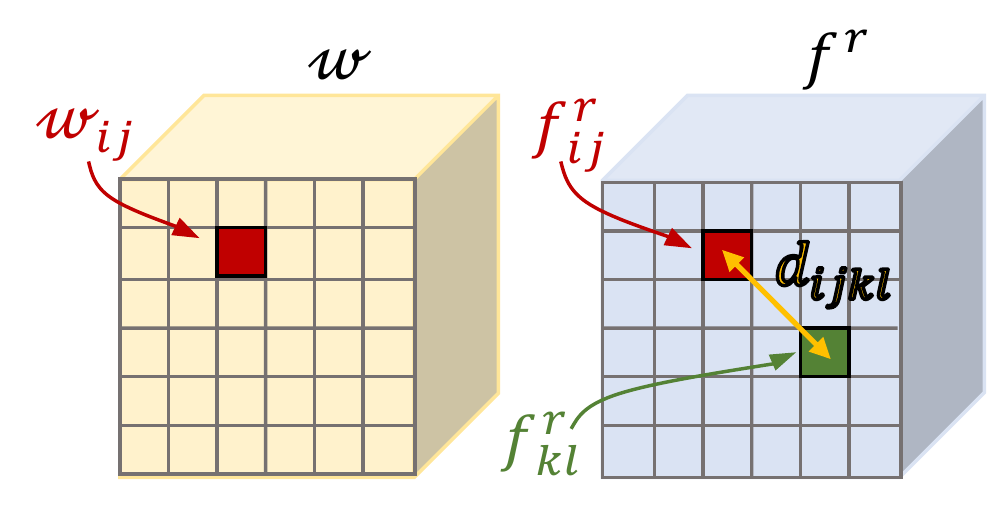}\vspace{-4mm}%
\caption{Visualization of the filter map $\wt$ and reference feature map $\ftr$.}\vspace{-4mm}
\label{fig:loss-on-ref}
\end{wrapfigure} 

Here, we introduce a flexible objective that exploits global information about the reference features $\ftr$, as discussed in Sec.~\ref{subsec:motivation}. For convenience, we follow the convention for global correlation \eqref{eq:global-corr} by letting subscripts denote absolute spatial locations. When establishing matching confidences for a reference frame location $(i,j)$, the feature correlation layer $\corr(\ftr, \ftq)$ only utilizes the encoded appearance $\ftr_{ij}$ at the location $(i,j)$. However, the reference feature map $\ftr$ also contains the encoding $\ftr_{kl}$ of other image regions $(k,l)$, which are likely to also occur in the query $\ftq$. To exploit this information, we therefore first replace the reference feature map $\ftr$ with our filter map $\wt$. The aim is then to find $\wt$ which enforces high confidences $\corr(\wt, \ftr)_{ijij} = \wt_{ij}\tp \ftr_{ij} \approx 1$ at the corresponding reference location $(i,j)$, while ensuring low matching confidences $\corr(\wt, \ftr)_{ijkl} = \wt_{ij}\tp \ftr_{kl} \approx 0$ for other locations $(k,l) \neq (i,j)$ in the reference map $\ftr$.
These constraints aim at designing $\wt_{ij}$, that explicitly suppresses the corresponding matching confidences in regions $\ftr_{kl}$ that have similar appearance as $\ftr_{ij}$, since these regions may also occur in $\ftq$. 

As a first attempt, the aforementioned reference-frame constraints could be realized by minimizing the quadratic objective $\|\corr(\wt, \ftr) - \delta\|^2$. Here, $\delta$ represents the desired correlation response, which in case of global correlation \eqref{eq:global-corr} is $\delta_{ijkl} = 1$ whenever $(i,j)=(k,l)$ and $\delta_{ijkl} = 0$ otherwise. The quadratic objective is attractive since it can be tackled with particularly effective optimization methods. On the other hand, the simple quadratic objective is known for its sensitivity to outliers. In our setting, the objective should in fact be largely indifferent to cases when a non-matching pair generates a strong negative correlation output $\wt_{ij}\tp \ftr_{kl} \ll 0$. This stems from the fact that any zero \emph{or} negative confidence is enough to indicate a non-match. However, such strong negative predictions receive a disproportionately large impact in the quadratic objective, instead compromising the quality of the correspondence volume in challenging regions with similar appearance. This issue is further amplified by the severe imbalance between examples of \emph{matches} and \emph{non-matches} in the objective.

To address these issues, we formulate a robust non-linear least squares objective. For a non-matching location pair ($\delta_{ijkl} = 0$), a positive correlation output $\corr(\wt, \ftr)_{ijkl} > 0$ corresponds to a similar appearance that should be suppressed, while negative correlation output $\corr(\wt, \ftr)_{ijkl} < 0$ is of little importance. We account for this asymmetry by introducing separate penalization weights $\ps_{ijkl}$ and $\ns_{ijkl}$ for positive and negative correlation outputs, respectively. The confidence values are thus mapped by the scalar function $\sigma$ defined as,
\begin{subequations}\label{eq:error-func}
\begin{align}
\sigma(c; \ps\!, \ns) &= 
\begin{dcases}
\ps c \,,& c\geq 0 \\ 
\ns c \,,& c < 0
\end{dcases} ,\  \\
\sigma_\eta( c; \ps\!, \ns) &= \frac{\ps\! - \ns\!}{2}\! \left(\!\sqrt{c^2 + \eta^2} - \eta\right) + \frac{\ps\! + \ns\!}{2}  c \,.
\end{align}
\end{subequations}
We have also defined a smooth approximation $\sigma_\eta$, which for $\eta > 0$ avoids the discontinuity in the derivative of $\sigma$ at $\corr(\wt, \ftr)=0$. The original function $\sigma = \sigma_0$ is retrieved by setting $\eta = 0$. 

By applying the function \eqref{eq:error-func}, the confidence values $\corr(\wt, \ftr)$ can be re-weighted using appropriate values for the weights $\ps$ and $\ns$. To address the question of how to set $\ps$ and $\ns$ in practice, recall that our objective defines a neural network module through the optimization \eqref{eq:minimize-loss}. This opens an interesting opportunity of learning $\ps$ and $\ns$ as parameters of the neural network. These can thus be trained along with all other parameters of the network for the end task. Specifically, we parametrize the weights as functions $\ps_{ijkl}=\psl(d_{ijkl})$ and $\ns_{ijkl}=\nsl(d_{ijkl})$ of the distance $d_{ijkl} = \sqrt{(i-k)^2 + (j-l)^2}$ between $w_{ij}$ and the example $\ftr_{kl}$. This strategy allows the network to learn the transition between the correct match $d_{ijij} = 0$ and the distant $d_{ijkl} \gg 0$ examples of non-matching features $\ftr_{kl}$.
Our robust and learnable objective function for integrating reference frame information is thus formulated as,
\begin{equation}
\label{eq:ref-loss}
\Lr(\wt; \ftr, \theta) = \left\| \sigma_\eta\big(\corr(\wt, \ftr); \, \ps, \ns\big) - \y \right\|^2 \,.
\end{equation}
Here we have additionally replaced the ideal correlation $\delta$ with a learnable target confidence $\y_{ijkl} = \yc(d_{ijkl})$, to add further flexibility. We parametrize $\psl$, $\nsl$, and $\yc$ using the strategy introduced in~\cite{dimp}, as piece-wise linear functions of the distance $d_{ijkl}$, further detailed in the appendix Sec.~\ref{Sec:arch-details}.

\subsection{Query Frame Objective}
\label{subsec:smooth-loss}

In the previous section, we formulated an approach that integrates the reference feature map $\ftr$ into the objective~\eqref{eq:minimize-loss}. However, as discussed in Sec.~\ref{subsec:motivation}, there is also rich information to gain from the query frame.
Firstly, correspondences between a pair of images must adhere to certain constraints, mainly that each point in the reference image can have at most a single match in the query image. 
Secondly, neighboring matches follow spatial smoothness priors, largely induced by the spatio-temporal continuity of the underlying 3D-scene. We encapsulate such constraints by defining a regularizing objective on the query frame, 
\begin{equation}
\label{eq:query-loss}
\Lq(\wt; \,\ftq, \theta) = \left\| R_\theta \ast \corr(\wt, \ftq) \right\|^2 \,.
\end{equation}
Here, $\ast$ denotes the convolution operator and $R_\theta \in \reals^{K^4\times Q}$ is a learnable 4D-kernel of spatial size $K$ and $Q$ number of output channels. A 4D-convolution operator allows us to fully utilize the structure of the 4D correspondence volume. Furthermore, its use is motivated by the translation invariance property induced by the 2D translation invariance of the two input feature maps.
$R_{\theta}$ is learnt, along with all other network parameters, by the SGD-based minimization of the final network training loss.

The use of smoothness priors has a long and successful tradition in classic variational formulations for optical flow, developed during the pre-deep learning era \cite{Baker2011,BroxM11,Horn1981,LiuYT11}.
We therefore take inspiration from these ideas. However, our approach offers several interesting conceptual differences.
First, our regularization operates directly on the matching confidences generated by the correlation operation, rather than the flow vectors. The correspondence volume provides a much richer description by encapsulating uncertainties in the correspondence assignment. Second, our objective is a function of the underlying filter map $\wt$, which is the input to the correlation layer. Third, our objective is implicitly minimized \emph{inside} a deep neural network. Finally, this further allows our regularizer $R_\theta$ to be learned in a fully end-to-end and data-driven manner. In contrast, classical methods rely on hand-crafted regularizers and priors. By integrating information from a local 4D-neighborhood, the operator $R_\theta$ in~\eqref{eq:query-loss} can enforce spatial smoothness by, for instance, learning differential operators. 
Moreover, our formulation lets the network learn the weighting of the query term~\eqref{eq:query-loss} in relation to the reference frame objective~\eqref{eq:ref-loss}, eliminating the need for such hyper-parameter tuning.

\subsection{Filter map prediction module $\wpred$}
\label{subsec:optim}

Our objective, employed in~\eqref{eq:minimize-loss}, is obtained by combining the reference~\eqref{eq:ref-loss} and query~\eqref{eq:query-loss} terms as,
\begin{equation}
\label{eq:total-loss}
L(\wt; \ftr, \ftq, \theta) = \Lr(\wt; \ftr, \theta) + \Lq(\wt; \,\ftq, \theta) + \|\lambda_\theta w\|^2 \,.
\end{equation}
The last term corresponds to a regularizing prior on $\wt$, weighted by the learnable scalar $\lambda_\theta \in \reals$. Note that while the reference frame objective $\Lr$ in~\eqref{eq:ref-loss} can be decomposed into independent terms for each location $\wt_{ij}$, the query term $\Lq$~\eqref{eq:query-loss} introduces dependencies between all elements in $\wt$. Efficiently optimizing such a high-dimensional problem during the forward pass of the network in order to implement~\eqref{eq:minimize-loss} may seem an impossibility. Next, we demonstrate that this can, in fact, be achieved by a combination of accurate initialization and a simple but powerful iterative procedure. Any neural network architecture employing feature correlation layers can thereby benefit from our module.

\parsection{Optimizer} While finding the global optima of~\eqref{eq:total-loss} within a small tolerance is costly, this is not necessary in our case. Instead, we can effectively utilize the information encoded in the objective~\eqref{eq:total-loss} by optimizing it to \emph{a sufficient degree}. We therefore derive the filter map $\wtm=\wpred_\theta(\ftr, \ftq)$ by applying an iterative optimization strategy. Specifically, we use the Steepest Descent algorithm, which was found effective in~\cite{dimp}. Given the current iterate $\wt^n$, the steepest descent method~\cite{NoceWrig06,Shewchuk} finds the step-length $\alpha^n$ that minimizes the objective in the gradient direction. This is obtained through a simple closed-form expression by first performing a Gauss-Newton approximation of~\eqref{eq:ref-loss}. The filter map is then updated by taking a gradient step with optimal length $\alpha^n$,
\begin{equation}
\label{eq:sd-iter}
\wt^{n+1}=\wt^{n}-\alpha^{n} \nabla L\left(\wt^{n}; \ftr, \ftq, \theta\right) \,,\quad \alpha^{n} = \argmin_\alpha L_\text{GN}^n\big(\wt^{n} - \alpha \nabla L(\wt^{n}; \ftr, \ftq, \theta)\big)  \,.
\end{equation}
Here, $L_\text{GN}^n$ is the Gauss-Newton approximation of \eqref{eq:total-loss} at $\wt^n$.
Both the gradient $\nabla L$ and the step length $\alpha^n$ are implemented using their closed form expressions with standard neural network modules, as detailed in the appendix Sec.~\ref{Sec:deriv}. Importantly, the operation \eqref{eq:sd-iter} is fully differentiable \wrt $\ftr$, $\ftq$, and $\theta$, allowing end-to-end training of all underlying network parameters.

\parsection{Initializer} To reduce the number of optimization iterations needed in the filter predictor network $\wpred$, we generate an initial filter map $\wt^0$ using an efficient and learnable module. We parametrize $\wt^0_{ij} = a_{ij}\ftr_{ij} + b_{ij}\bar{\ftr}$, where $\bar{\ftr} \in \reals^d$ is the spatial average reference vector, encoding contextual information. Intuitively, we wish $\wt^0$ to have a high activation $(\wt^0_{ij})\tp\ftr_{ij} = 1$ at the matching position and $(\wt^0_{ij})\tp\bar{\ftr} = 0$. The scalar coefficients $a_{ij}$ and $b_{ij}$ are then easily found by solving these equations. 
Details are given in the appendix Sec.~\ref{sec:sup-initializer}. 

\section{Experiments}
\label{sec:exp-val}

We perform comprehensive experiments for two tasks: geometric correspondences and optical flow. We additionally show that our method can be successfully applied to the task of semantic matching. Both global and local correlation-based versions of our \dicor module are analyzed by integrating them into two recent state-of-the-art networks. Further results, analysis, and visualizations along with more details regarding architectures and datasets are provided in the appendix.

\subsection{Geometric matching}
\label{subsec:geom-exp}

We first evaluate our \dicor module for dense geometric matching by integrating it into the recent GLU-Net~\cite{GLUNet}. GLU-Net is a 4-level pyramidal network, operating at two image resolutions to estimate dense flow fields. It relies on a global correlation at the coarsest level to capture long-range displacements and uses local correlations in the subsequent levels.

\parsection{Experimental setup}
We create GLU-Net-\dicor by replacing global and local feature correlation layers with our global and local \dicor modules, respectively. The global \dicor module employs the full objective \eqref{eq:total-loss}, while the local variant uses only the reference term \eqref{eq:ref-loss}. We use three steepest descent iterations during training and increase the number during inference.
We follow the same self-supervised training procedure and data as in~\cite{GLUNet}, applying synthetic homography transformations to images compiled from different sources to ensure diversity. We refer to this as the \emph{Static} dataset, since it simulates a static scene. For better compatibility with real 3D scenes and moving objects, we further introduce a \emph{Dynamic} training dataset, by augmenting the \emph{Static} data with random independently moving objects from the COCO~\cite{coco} dataset. In all experiments, we compare the results of GLU-Net and GLU-Net-\dicor trained with the \emph{same} data, and according to the \emph{same} procedure.

\parsection{Evaluation datasets and metrics} We first employ the 59 sequences of the \textbf{HPatches} dataset~\cite{Lenc}, consisting of planar scenes from different viewpoints.
We additionally utilize the multi-view \textbf{ETH3D} dataset~\cite{ETH3d}, depicting indoor and outdoor scenes captured from a moving hand-held camera.
We follow the protocol of \cite{GLUNet}, sampling image pairs at different intervals to analyze varying magnitude of geometric transformations. Finally, because of the difficulty to obtain dense annotations on real imagery with extreme viewpoint and varying imaging condition, we also evaluate our model on sparse correspondences available on the \textbf{MegaDepth}~\cite{megadepth} dataset, according to the protocol introduced in~\cite{shen2019ransacflow}. 
We use the \emph{Static} training data for the comparison on the HPatches dataset and the \emph{Dynamic} training data for the ETH3D and MegaDepth datasets. 
In line with previous works~\cite{Melekhov2019, GLUNet}, we employ the Average End-Point Error (AEPE) and Percentage of Correct Keypoints at a given pixel threshold $T$ (PCK-$T$) as the evaluation metrics. 

\begin{wraptable}{r}{6.7cm}
\vspace{-6mm}
\caption{HPatches homography dataset~\cite{Lenc}.}\resizebox{6.6cm}{!}{%
\begin{tabular}{lccc}
\toprule
             &  AEPE $\downarrow$   & PCK-1 (\%) $\uparrow$ & PCK-5 (\%) $\uparrow$  \\ \midrule
DGC-Net~\cite{Melekhov2019} &  33.26 & 12.00   & 58.06  \\
GLU-Net   &   25.05 &  39.55   &  78.54  \\ 
GLU-Net-\dicor (Ours) &  \textbf{20.16} & \textbf{41.55} & \textbf{81.43} \\
\bottomrule
\end{tabular}%
}\vspace{-4mm}
\label{tab:geo-match-HP}
\end{wraptable}

\begin{figure*}[t]
\centering
\includegraphics[width=0.31\textwidth]{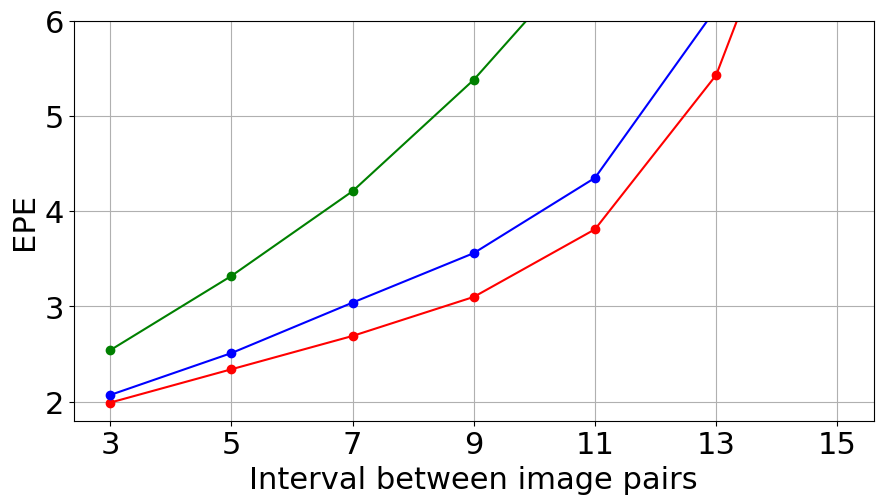} \hspace{0.2cm}
\includegraphics[width=0.31\textwidth]{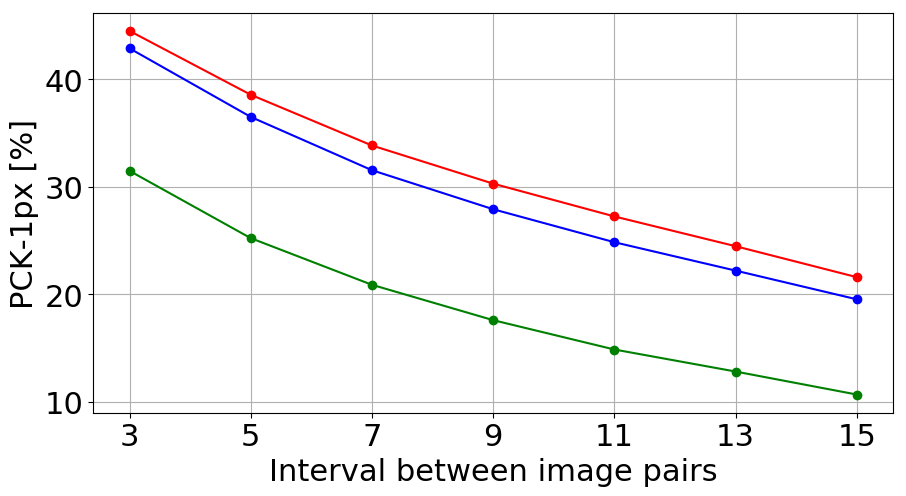} \hspace{0.2cm} 
\includegraphics[width=0.31\textwidth]{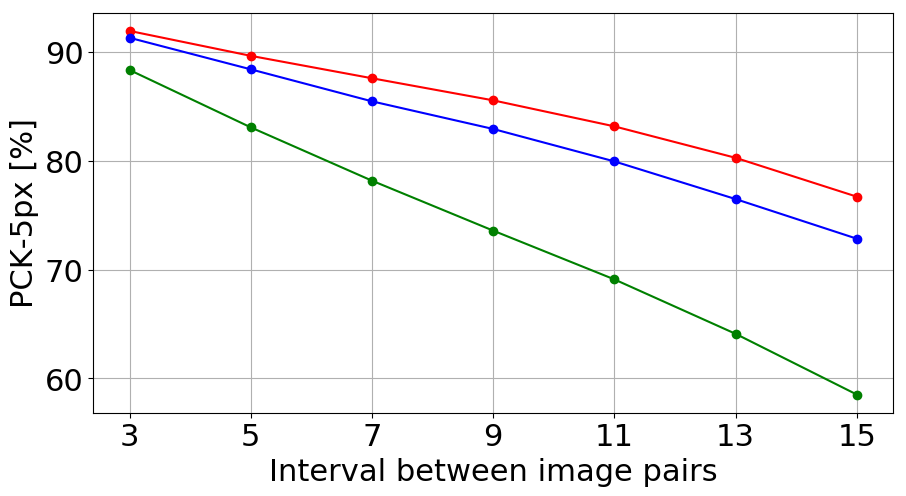}
\includegraphics[width=0.50\textwidth,trim=0 0 0 15]{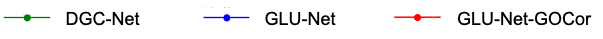}\vspace{-4mm}
\caption{Results on geometric matching dataset ETH3D~\cite{ETH3d}. AEPE (left), PCK-1 (center), and PCK-5 (right) are plotted \wrt the inter-frame interval length.}\vspace{-6mm}
\label{fig:ETH3d}
\end{figure*}

\parsection{Results} 
In Table~\ref{tab:geo-match-HP}, we present results on HPatches. We also report the results of the recent state-of-the-art DGC-Net~\cite{Melekhov2019} for reference. Our GLU-Net-\dicor outperforms original GLU-Net by a large margin, achieving both higher accuracy in terms of PCK, and better robustness to large errors as indicated by AEPE. 
In Figure~\ref{fig:ETH3d}, we plot AEPE, PCK-1 and PCK-5 obtained on the ETH3D images. 
For all intervals, our approach is consistently better than baseline GLU-Net. We note that the improvement is particularly prominent at larger intra-frame intervals, strongly indicating that our \dicor module better copes with large appearance variations due to large viewpoint changes, compared to the feature correlation layer. 

\begin{wraptable}{l}{6.5cm}
\vspace{-6mm}
\caption{Results on sparse correspondences of the MegaDepth dataset~\cite{megadepth}. }
\resizebox{6.5cm}{!}{%
\begin{tabular}{lccc} \toprule
  & PCK-1 (\%) $\uparrow$               & PCK-3 (\%) $\uparrow$     & PCK-5  (\%) $\uparrow$    \\ \midrule
GLU-Net            & 21.58           & 52.18 & 61.78 \\
GLU-Net-GOCor    (Ours)     &  \textbf{37.28} &        \textbf{61.18}         & \textbf{68.08} \\
\bottomrule
\end{tabular}%
}\vspace{-4mm}
\label{tab:megadepth}
\end{wraptable}
\vspace{-1mm} This is also confirmed by the results on MegaDepth in Table \ref{tab:megadepth}. Images depict extreme view-point changes with as little as 10\% of co-visible regions. In this case as well, GOCor brings significant improvement, particularly in pixel-accuracy (PCK-1).

\subsection{Optical flow}
\label{subsec:optical-flow}

Next, we evaluate our \dicor module for the task of optical flow estimation, by integrating it into the state-of-the-art PWC-Net~\cite{Sun2018, Sun2019} and GLU-Net~\cite{GLUNet} architectures. PWC-Net~\cite{Sun2018} is based on a 5-level pyramidal network, estimating the dense flow field at each level using a local correlation layer. 

\parsection{Experimental setup}
We replace all local correlation layers with our local \dicor module to obtain PWC-Net-\dicor. We finetune PWC-Net-\dicor on \textit{3D-Things}~\cite{Ilg2017a}, using the publicly available PWC-Net weights trained on \textit{Flying-Chairs}~\cite{Dosovitskiy2015} and \textit{3D-Things}~\cite{Ilg2017a} as initialization. For fair comparison, we also finetune the standard PWC-Net on \textit{3D-Things} with the same schedule. Finally, we also finetune PWC-Net-\dicor on the \textit{Sintel}~\cite{Butler2012} training dataset according to the schedule introduced in~\cite{Ilg2017a, Sun2018}. 
As described in Sec.~\ref{subsec:geom-exp}, we train both GLU-Net and GLU-Net-\dicor on the \emph{Dynamic} training set. For the global and local \dicor modules, we use the same settings as in \ref{subsec:geom-exp}.

\parsection{Datasets and evaluation metrics} For evaluation, we use the established \textbf{KITTI} dataset~\cite{Geiger2013}, composed of real road sequences captured by a car-mounted stereo camera rig. We also utilize the \textbf{Sintel} dataset~\cite{Butler2012}, which consists of 3D animated sequences. We use the standard evaluation metrics, namely the AEPE and F1 for KITTI. The latter represents the percentage of optical flow outliers. For Sintel, we employ AEPE together with PCK, \ie percentage of inliers. In line with \cite{Hui2018, Hui2019, Sun2018, Sun2019, GLUNet}, we show results on the training splits of these datasets. 

\newcommand{\trainres}[1]{\textcolor{gray}{#1}}
\begin{table}[t]
\centering\vspace{-3mm}
\caption{Results for the optical flow task on the training splits of KITTI~\cite{Geiger2013} and Sintel~\cite{Butler2012}.
A result in parenthesis indicates that the dataset was used for training.}\vspace{-2mm}%
\resizebox{0.99\textwidth}{!}{%
\begin{tabular}{lcc|cc|ccc|ccc}
\toprule
             & \multicolumn{2}{c}{\textbf{KITTI-2012}} & \multicolumn{2}{c}{\textbf{KITTI-2015}} & \multicolumn{3}{c}{\textbf{Sintel Clean}} & \multicolumn{3}{c}{\textbf{Sintel Final}}\\ 
  & AEPE  $\downarrow$            & F1   (\%)   $\downarrow$      & AEPE  $\downarrow$              & F1  (\%)  $\downarrow$ & AEPE  $\downarrow$  & PCK-1  (\%) $\uparrow$ & PCK-5  (\%) $\uparrow$ & AEPE $\downarrow$   & PCK-1  (\%) $\uparrow$ & PCK-5  (\%) $\uparrow$ \\ \midrule
GLU-Net & 3.14 & 19.76 & 7.49 & 33.83 & 4.25 & 62.08 & 88.40 & 5.50 & 57.85 & 85.10 \\
GLU-Net-\dicor & \textbf{2.68} & \textbf{15.43} & \textbf{6.68} & \textbf{27.57} & \textbf{3.80} & \textbf{67.12} & \textbf{90.41} & \textbf{4.90} & \textbf{63.38} & \textbf{87.69} \\ \midrule \midrule
PWC-Net (from paper) & 4.14 & 21.38 & 10.35 & 33.67 & 2.55 & - & - & 3.93 & - & - \\
PWC-Net (\textit{ft 3D-Things}) & 4.34 & 20.90 & 10.81 & 32.75 & 2.43 & 81.28 & 93.74 & 3.77 & 76.53 & 90.87 \\ 
PWC-Net-\dicor (\textit{ft 3D-Things}) & \textbf{4.12} & \textbf{19.31} & \textbf{10.33} & \textbf{30.53} & \textbf{2.38} & \textbf{82.17} & \textbf{94.13} & \textbf{3.70} &  \textbf{77.34} & \textbf{91.20} \\ \midrule
PWC-Net (\textit{ft Sintel}) & 2.94 & 12.70 & 8.15 & 24.35 & \trainres{(1.70)} & \trainres{-} & \trainres{-} & \trainres{(2.21)} & \trainres{-} & \trainres{-} \\
PWC-Net-\dicor (\textit{ft Sintel}) & \textbf{2.60} & \textbf{9.67} & \textbf{7.64} & \textbf{20.93} & \trainres{(1.74)} & \trainres{(87.93)} & \trainres{(95.54)} &  \trainres{(2.28)} &  \trainres{(84.15)} & \trainres{(93.71)} \\
\bottomrule
\end{tabular}%
}\vspace{-5mm}
\label{tab:optical-flow}
\end{table}

\parsection{Results} Results are reported in Tab.~\ref{tab:optical-flow}. First, compared to the GLU-Net baseline, our \dicor module brings significant improvements in both AEPE and F1/PCK on all optical flow datasets.
Next we compare the PWC-Net based methods trained on \emph{3D-Things} (middle section) and report the official result~\cite{Sun2018,Sun2019} along with our fine-tuned versions. While our PWC-Net-\dicor obtains a similar AEPE, it achieves substantially better accuracy, with a $3\%$ improvement in F1 metric on KITTI-2015. 
After finetuning on Sintel images, both PWC-Net and PWC-Net-\dicor achieve similar results on the Sintel training data (in parenthesis). However, the PWC-Net-\dicor version provides superior results on the two KITTI datasets. This clearly demonstrates the superior domain generalization capabilities of our \dicor module. Note that both methods in the bottom section of Tab.~\ref{tab:optical-flow} are only trained on animated datasets, while KITTI consists of natural road-scenes. Thanks to the effective objective-based adaption performed in our matching module during inference, PWC-Net-\dicor excels even with a sub-optimal feature embedding trained for animated images, and when exposed to previously unseen motion patterns. This is a particularly important property in the context of optical flow and geometric matching, where collection of labelled realistic training data is prohibitively expensive, forcing methods to resort to synthetic and animated datasets.

\subsection{Generalization to semantic matching}

\begin{wraptable}{r}{5.2cm}
\vspace{-7mm}
\caption{PCK [\%] on TSS~\cite{Taniai2016}.}
\label{tab:Robotcar-MegaDepth}
\centering
\vspace{0.5mm}
\resizebox{\linewidth}{!}{%
\begin{tabular}{@{}l@{}c@{~~}c@{~~}c@{~~}c@{}}
\toprule
              & \textbf{FGD3Car}      & \textbf{JODS}    &  \textbf{PASCAL}  &  \textbf{All}        \\ \midrule
Semantic-GLU-Net \cite{GLUNet} & 94.4 & 75.5 & \textbf{78.3} & 82.8 \\
GLU-Net & 93.2 & 73.3 & 71.1 & 79.2 \\
GLU-Net-GOCor (Ours) &  \textbf{94.6} &  \textbf{77.9} &  77.7 & \textbf{83.4}   \\ \bottomrule
\end{tabular}%
} \\
\centering
\vspace{-4mm}
\end{wraptable}
We additionally compare the performance of GOCor to the feature correlation layer on the task of semantic matching. In Table~\ref{tab:Robotcar-MegaDepth}, we evaluate our GLU-Net-\dicor trained on the \textit{Static} data without any additional finetuning, for dense semantic matching on the TSS dataset~\cite{Taniai2016}. In the semantic correspondence task, images depict different instances of the same object category (e.g. \emph{horse}). As a result, the value of additional reference frame information (Sec.~\ref{subsec:motivation} and ~\ref{subsec:loss-ref}) is not as pronounced in semantic matching compared to geometric matching or optical flow. Indeed, our reference frame objective uses its full potential when both the reference and the query images depict similar regions from the \emph{same scene}. Nevertheless, our GLU-Net-\dicor sets a new state-of-the-art on this dataset, even outperforming Semantic-GLU-Net~\cite{GLUNet}.

\subsection{Run-time}
\label{run-time}

\begin{wraptable}{r}{4.2cm}
\vspace{-7mm}
\caption{Run time [ms] averaged over the 194 image pairs of KITTI-2012.}
\label{tab:runtime-main-paper}
\centering
\vspace{0.5mm}
\resizebox{\linewidth}{!}{%
\begin{tabular}{ll}
\toprule 
& Run-time [ms] \\
               \midrule
PWC-Net & 118.05 \\
PWC-Net-\dicor & 203.02 \\
GLU-Net &  154.97 \\
GLU-Net-\dicor & 261.90 \\ \bottomrule
\end{tabular}%
} \\
\centering
\vspace{-4mm}
\end{wraptable}

In Table~\ref{tab:runtime-main-paper}, we compare the run time of our GOCor-based networks to their original versions on the KITTI-2012 dataset. The timings are obtained on the same desktop with an NVIDIA Titan X GPU.  
While our \dicor module leads to increased computation, the run-time remains within reasonable margins thanks to our dedicated optimization module, described in Sec.~\ref{subsec:optim}. We can further control the trade of between computation and performance by varying the number of steepest descent iterations in our GOCor module. In Appendix~\ref{iteration} we provide such a detailed analysis, and propose faster operating points with only minor degradation in performance.

\begin{table}[t]
\centering
\caption{Ablation study of key aspects of our approach on three different datasets.}\vspace{-2mm}%
\resizebox{0.99\textwidth}{!}{%
\begin{tabular}{clcc|cc|cc}
\toprule
          &   & \multicolumn{2}{c}{\textbf{HPatches}} & \multicolumn{2}{c}{\textbf{KITTI-2012}} & \multicolumn{2}{c}{\textbf{KITTI-2015}} \\ 
             & & AEPE  $\downarrow$   & PCK-5  (\%) $\uparrow$  & AEPE  $\downarrow$             & F1   (\%)  $\downarrow$        & AEPE    $\downarrow$            & F1  (\%)     $\downarrow$         \\ \midrule
\textbf{(I)} & BaseNet  &  30.94  & 69.22  & 4.03 & 30.49 & 8.93 & 48.66 \\
\textbf{(II)} & BaseNet + NC-Net~\cite{Rocco2018b}  & 39.15 & 63.52 & 4.41 & 34.78 & 9.86 & 52.78 \\
\textbf{(III)} & BaseNet + Global-\dicor Linear Regression & 27.02 & 68.12 & 4.31 & 35.30 & 8.93 & 52.64 \\
\textbf{(IV)} & BaseNet + Global-\dicor $\Lr$  & 26.27 & 71.29 & 3.91 & 29.77 & 8.50 & 46.24  \\
\textbf{(V)} & BaseNet + Global-\dicor $\Lr + \Lq$  & 25.30 & 71.21 & 3.74 & 26.82 & 7.87 & 43.08  \\
\textbf{(VI)} & BaseNet + Global-\dicor $\Lr + \Lq$ + Local-\dicor  & \textbf{23.57} & \textbf{78.30} & \textbf{3.45} & \textbf{25.42} & \textbf{7.10} & \textbf{39.57} \\
\bottomrule
\end{tabular}%
}\vspace{-5mm}
\label{tab:ablation}
\end{table}

\begin{figure}[b]
\centering
\vspace{-5mm}
\includegraphics[width=0.999\textwidth]{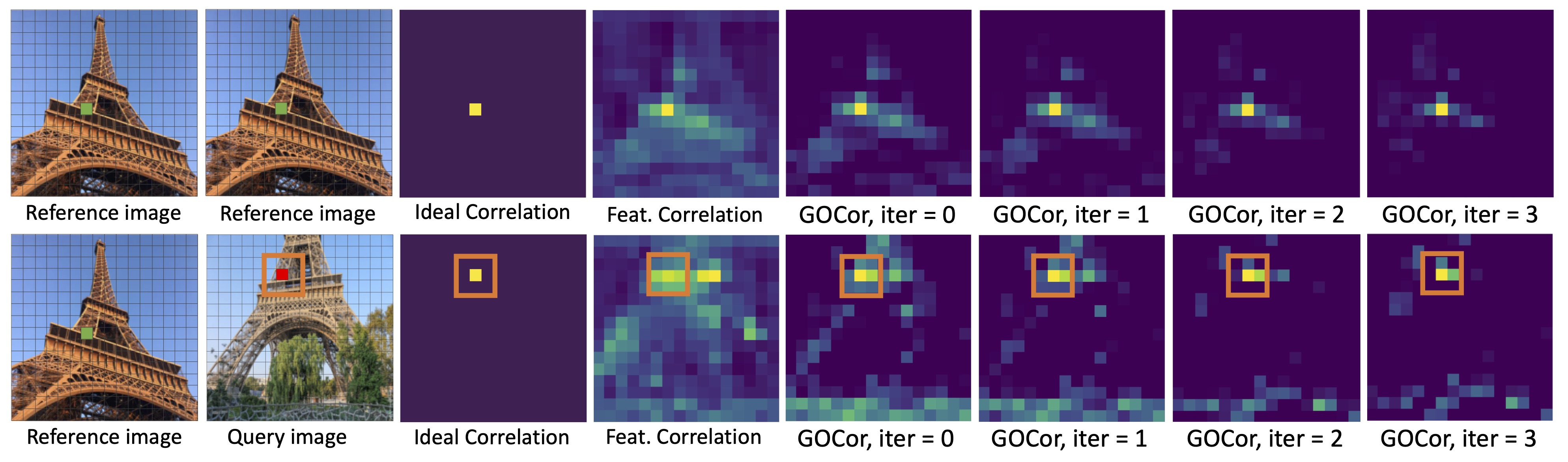}
\vspace{-8mm}
\caption{Visualization of the matching confidences computed between the indicated location (green) in the reference image and all locations of either the reference image itself or the query image. 
}
\label{fig:corr-diff-iter}
\end{figure}

\subsection{Ablation study}
\label{subsec:ablation}

Finally, we analyze key components of our approach. We first design a powerful baseline architecture estimating dense flow fields, called BaseNet. 
It consists of a three-level pyramidal CNN-network, inspired by~\cite{GLUNet}, employing a global correlation layer followed by two local layers. All methods are trained with the \emph{Dynamic} data, described in Sec.~\ref{subsec:geom-exp}.
Results on HPatches, KITTI-2012 and KITTI-2015 are reported in Tab.~\ref{tab:ablation}. We first analyse the effect of replacing the feature correlation layer with \dicor at the global correlation level. The version denoted \textbf{(IV)} employs our global \dicor using solely the reference-based objective $\Lr$ (Sec.~\ref{subsec:loss-ref}). It leads to significantly better results on all datasets compared to standard BaseNet \textbf{(I)}. 
Instead of our robust reference loss $\Lr$, the version \textbf{(III)} employs a standard linear regression objective $\|\corr(\wt, \ftr) - \delta\|^2$, leading to substantially worse results. 
We also compare with adding the post-processing strategy proposed in~\cite{Rocco2018b} \textbf{(II)}, employing 4D-convolutions and enforcing cyclic consistency. This generally leads to a degradation in performance, likely caused by the inability to cope with the domain gap between training and test data.
From \textbf{(IV)} to \textbf{(V)} we integrate our query frame objective $\Lq$ (Sec.~\ref{subsec:smooth-loss}), which results in major gains, particularly on the more challenging KITTI datasets.
Finally, we replace the local correlation layers with our local \dicor module in \textbf{(VI)}. This leads to large improvements on all datasets and metrics. 

In Figure~\ref{fig:corr-diff-iter}, we visualize the relevance of our reference loss (Sec.~\ref{subsec:loss-ref}) qualitatively by plotting the correspondence volume outputted by our global \dicor module, when correlating a particular point (i,j) of the reference image with all locations of either \emph{the reference itself or the query image}. The predicted correspondence volume gets increasingly distinctive after each iteration in the GOCor layer. Specifically, it is clearly visible that final matching confidences with the query image benefits from optimizing the correlation scores with the reference image itself, using Eq. \eqref{eq:ref-loss}.

\section{Conclusion}
\label{sec:conclusion}

We propose a neural network module for predicting globally optimized matching confidences between two deep feature maps. It acts as a direct alternative to feature correlation layers. We integrate unexploited information about the reference and query frames by formulating an objective function, which is minimized during inference through an iterative optimization strategy. Our approach thereby explicitly accounts for, \eg, similar image regions. Our resulting \dicor module is thoroughly analysed and evaluated on the tasks of geometric correspondences and optical flow, with an extension to dense semantic matching. When integrated into state-of-the-art networks, it significantly outperforms the feature correlation layer.

\section*{Broader Impact}
Our feature correspondence matching module can be beneficial in a wide range of applications relying on explicit or implicit matching between images, such as visual localization~\cite{johannes2017, Taira2018}, 3D-reconstruction~\cite{AgarwalFSSCSS11}, structure-from-motion~\cite{SchonbergerF16}, action recognition~\cite{SimonyanZ14} and autonomous driving~\cite{JanaiGBG17}.
On the other hand, any image matching algorithm runs the risk of being used for malevolent tasks, such as malicious image manipulation or image surveillance system. However, our module is only one building block to be integrated in a larger pipeline. On its own, it therefore has little chances of being wrongfully used. 

\begin{ack}
This work was partly supported by the ETH Z\"urich Fund (OK), a Huawei Technologies Oy (Finland) project, an Amazon AWS grant, and an Nvidia hardware grant.
\end{ack}

\medskip
{\small
\bibliographystyle{plain}
\bibliography{biblio}

\begin{thebibliography}{10}

\bibitem{AgarwalFSSCSS11}
Sameer Agarwal, Yasutaka Furukawa, Noah Snavely, Ian Simon, Brian Curless,
  Steven~M. Seitz, and Richard Szeliski.
\newblock Building rome in a day.
\newblock {\em Commun. {ACM}}, 54(10):105--112, 2011.

\bibitem{Baker2011}
Simon Baker, Daniel Scharstein, J.~P. Lewis, Stefan Roth, Michael~J. Black, and
  Richard Szeliski.
\newblock A database and evaluation methodology for optical flow.
\newblock {\em International Journal of Computer Vision}, 92(1):1--31, 2011.

\bibitem{Lenc}
Vassileios Balntas, Karel Lenc, Andrea Vedaldi, and Krystian Mikolajczyk.
\newblock Hpatches: {A} benchmark and evaluation of handcrafted and learned
  local descriptors.
\newblock In {\em 2017 {IEEE} Conference on Computer Vision and Pattern
  Recognition, {CVPR} 2017, Honolulu, HI, USA, July 21-26, 2017}, pages
  3852--3861, 2017.

\bibitem{r2d2}
Luca Bertinetto, Jo{\~{a}}o~F. Henriques, Philip H.~S. Torr, and Andrea
  Vedaldi.
\newblock Meta-learning with differentiable closed-form solvers.
\newblock In {\em 7th International Conference on Learning Representations,
  {ICLR} 2019, New Orleans, LA, USA, May 6-9, 2019}, 2019.

\bibitem{dimp}
Goutam Bhat, Martin Danelljan, Luc~Van Gool, and Radu Timofte.
\newblock Learning discriminative model prediction for tracking.
\newblock In {\em 2019 {IEEE/CVF} International Conference on Computer Vision,
  {ICCV} 2019, Seoul, Korea (South), October 27 - November 2, 2019}, pages
  6181--6190. {IEEE}, 2019.

\bibitem{lwtl}
Goutam Bhat, Felix~J{\"{a}}remo Lawin, Martin Danelljan, Andreas Robinson,
  Michael Felsberg, Luc~Van Gool, and Radu Timofte.
\newblock Learning what to learn for video object segmentation.
\newblock In {\em European Conference on Computer Vision {ECCV}}, 2020.

\bibitem{BroxM11}
Thomas Brox and Jitendra Malik.
\newblock Large displacement optical flow: Descriptor matching in variational
  motion estimation.
\newblock {\em {IEEE} Trans. Pattern Anal. Mach. Intell.}, 33(3):500--513,
  2011.

\bibitem{Butler2012}
Daniel~J. Butler, Jonas Wulff, Garrett~B. Stanley, and Michael~J. Black.
\newblock A naturalistic open source movie for optical flow evaluation.
\newblock In {\em Computer Vision - {ECCV} 2012 - 12th European Conference on
  Computer Vision, Florence, Italy, October 7-13, 2012, Proceedings, Part
  {VI}}, pages 611--625, 2012.

\bibitem{Chatfield14}
K.~Chatfield, K.~Simonyan, A.~Vedaldi, and A.~Zisserman.
\newblock Return of the devil in the details: Delving deep into convolutional
  nets.
\newblock In {\em BMVC}, 2014.

\bibitem{ChenWLF19}
Jianchun Chen, Lingjing Wang, Xiang Li, and Yi~Fang.
\newblock Arbicon-net: Arbitrary continuous geometric transformation networks
  for image registration.
\newblock In Hanna~M. Wallach, Hugo Larochelle, Alina Beygelzimer, Florence
  d'Alch{\'{e}}{-}Buc, Emily~B. Fox, and Roman Garnett, editors, {\em Advances
  in Neural Information Processing Systems 32: Annual Conference on Neural
  Information Processing Systems 2019, NeurIPS 2019, 8-14 December 2019,
  Vancouver, BC, Canada}, pages 3410--3420, 2019.

\bibitem{blazingly}
Yuhua Chen, Jordi Pont-Tuset, Alberto Montes, and Luc Van~Gool.
\newblock Blazingly fast video object segmentation with pixel-wise metric
  learning.
\newblock In {\em Proceedings of the IEEE Conference on Computer Vision and
  Pattern Recognition}, pages 1189--1198, 2018.

\bibitem{ChenS0YH15}
Zhuoyuan Chen, Xun Sun, Liang Wang, Yinan Yu, and Chang Huang.
\newblock A deep visual correspondence embedding model for stereo matching
  costs.
\newblock In {\em 2015 {IEEE} International Conference on Computer Vision,
  {ICCV} 2015, Santiago, Chile, December 7-13, 2015}, pages 972--980, 2015.

\bibitem{Choy2016}
Christopher~Bongsoo Choy, JunYoung Gwak, Silvio Savarese, and Manmohan~Krishna
  Chandraker.
\newblock Universal correspondence network.
\newblock In {\em Advances in Neural Information Processing Systems 29: Annual
  Conference on Neural Information Processing Systems 2016, December 5-10,
  2016, Barcelona, Spain}, pages 2406--2414, 2016.

\bibitem{Cordts2016}
Marius Cordts, Mohamed Omran, Sebastian Ramos, Timo Rehfeld, Markus Enzweiler,
  Rodrigo Benenson, Uwe Franke, Stefan Roth, and Bernt Schiele.
\newblock The cityscapes dataset for semantic urban scene understanding.
\newblock In {\em Proc. of the IEEE Conference on Computer Vision and Pattern
  Recognition (CVPR)}, 2016.

\bibitem{DeToneMR16}
Daniel DeTone, Tomasz Malisiewicz, and Andrew Rabinovich.
\newblock Deep image homography estimation.
\newblock {\em CoRR}, abs/1606.03798, 2016.

\bibitem{Dosovitskiy2015}
Alexey Dosovitskiy, Philipp Fischer, Eddy Ilg, Philip H{\"{a}}usser, Caner
  Hazirbas, Vladimir Golkov, Patrick van~der Smagt, Daniel Cremers, and Thomas
  Brox.
\newblock Flownet: Learning optical flow with convolutional networks.
\newblock In {\em 2015 {IEEE} International Conference on Computer Vision,
  {ICCV} 2015, Santiago, Chile, December 7-13, 2015}, pages 2758--2766, 2015.

\bibitem{Geiger2013}
Andreas Geiger, Philip Lenz, Christoph Stiller, and Raquel Urtasun.
\newblock Vision meets robotics: The kitti dataset.
\newblock {\em I. J. Robotic Res.}, 32(11):1231--1237, 2013.

\bibitem{Horn1981}
Berthold K.~P. Horn and Brian~G. Schunck.
\newblock "determining optical flow": {A} retrospective.
\newblock {\em Artif. Intell.}, 59(1-2):81--87, 1993.

\bibitem{VM}
Yuan-Ting Hu, Jia-Bin Huang, and Alexander~G Schwing.
\newblock Videomatch: Matching based video object segmentation.
\newblock In {\em European Conference on Computer Vision}, pages 56--73.
  Springer, 2018.

\bibitem{DCCNet}
Shuaiyi Huang, Qiuyue Wang, Songyang Zhang, Shipeng Yan, and Xuming He.
\newblock Dynamic context correspondence network for semantic alignment.
\newblock In {\em 2019 {IEEE/CVF} International Conference on Computer Vision,
  {ICCV} 2019, Seoul, Korea (South), October 27 - November 2, 2019}, pages
  2010--2019. {IEEE}, 2019.

\bibitem{Hui2018}
Tak{-}Wai Hui, Xiaoou Tang, and Chen~Change Loy.
\newblock Liteflownet: {A} lightweight convolutional neural network for optical
  flow estimation.
\newblock In {\em 2018 {IEEE} Conference on Computer Vision and Pattern
  Recognition, {CVPR} 2018, Salt Lake City, UT, USA, June 18-22, 2018}, pages
  8981--8989, 2018.

\bibitem{Hui2019}
Tak-Wai Hui, Xiaoou Tang, and Chen~Change Loy.
\newblock A {L}ightweight {O}ptical {F}low {CNN} - {R}evisiting {D}ata
  {F}idelity and {R}egularization.
\newblock 2020.

\bibitem{Ignatov2017}
Andrey Ignatov, Nikolay Kobyshev, Radu Timofte, Kenneth Vanhoey, and Luc~Van
  Gool.
\newblock Dslr-quality photos on mobile devices with deep convolutional
  networks.
\newblock In {\em {IEEE} International Conference on Computer Vision, {ICCV}
  2017, Venice, Italy, October 22-29, 2017}, pages 3297--3305, 2017.

\bibitem{Ilg2017a}
Eddy Ilg, Nikolaus Mayer, Tonmoy Saikia, Margret Keuper, Alexey Dosovitskiy,
  and Thomas Brox.
\newblock Flownet 2.0: Evolution of optical flow estimation with deep networks.
\newblock In {\em 2017 {IEEE} Conference on Computer Vision and Pattern
  Recognition, {CVPR} 2017, Honolulu, HI, USA, July 21-26, 2017}, pages
  1647--1655. {IEEE} Computer Society, 2017.

\bibitem{JanaiGBG17}
Joel Janai, Fatma G{\"{u}}ney, Aseem Behl, and Andreas Geiger.
\newblock Computer vision for autonomous vehicles: Problems, datasets and
  state-of-the-art.
\newblock {\em CoRR}, abs/1704.05519, 2017.

\bibitem{Jeon}
Sangryul Jeon, Seungryong Kim, Dongbo Min, and Kwanghoon Sohn.
\newblock {PARN:} pyramidal affine regression networks for dense semantic
  correspondence.
\newblock In {\em Computer Vision - {ECCV} 2018 - 15th European Conference,
  Munich, Germany, September 8-14, 2018, Proceedings, Part {VI}}, pages
  355--371, 2018.

\bibitem{JiaBTG16}
Xu~Jia, Bert~De Brabandere, Tinne Tuytelaars, and Luc~Van Gool.
\newblock Dynamic filter networks.
\newblock In {\em Advances in Neural Information Processing Systems 29: Annual
  Conference on Neural Information Processing Systems 2016, December 5-10,
  2016, Barcelona, Spain}, pages 667--675, 2016.

\bibitem{Kim2018}
Seungryong Kim, Stephen Lin, Sangryul Jeon, Dongbo Min, and Kwanghoon Sohn.
\newblock Recurrent transformer networks for semantic correspondence.
\newblock In {\em Advances in Neural Information Processing Systems 31: Annual
  Conference on Neural Information Processing Systems 2018, NeurIPS 2018, 3-8
  December 2018, Montr{\'{e}}al, Canada.}, pages 6129--6139, 2018.

\bibitem{KimMHLS19}
Seungryong Kim, Dongbo Min, Bumsub Ham, Stephen Lin, and Kwanghoon Sohn.
\newblock {FCSS:} fully convolutional self-similarity for dense semantic
  correspondence.
\newblock {\em {IEEE} Trans. Pattern Anal. Mach. Intell.}, 41(3):581--595,
  2019.

\bibitem{Kim2019}
Seungryong Kim, Dongbo Min, Somi Jeong, Sunok Kim, Sangryul Jeon, and Kwanghoon
  Sohn.
\newblock Semantic attribute matching networks.
\newblock In {\em {IEEE} Conference on Computer Vision and Pattern Recognition,
  {CVPR} 2019, Long Beach, CA, USA, June 16-20, 2019}, pages 12339--12348,
  2019.

\bibitem{Laskar2019}
Zakaria Laskar, Iaroslav Melekhov, Hamed~R. Tavakoli, Juha Ylioinas, and Juho
  Kannala.
\newblock Geometric image correspondence verification by dense pixel matching.
\newblock {\em CoRR}, abs/1904.06882, 2019.

\bibitem{metaoptnet}
Kwonjoon Lee, Subhransu Maji, Avinash Ravichandran, and Stefano Soatto.
\newblock Meta-learning with differentiable convex optimization.
\newblock In {\em {IEEE} Conference on Computer Vision and Pattern Recognition,
  {CVPR} 2019, Long Beach, CA, USA, June 16-20, 2019}, pages 10657--10665,
  2019.

\bibitem{Shuda2020}
Shuda Li, Kai Han, Theo~W. Costain, Henry Howard{-}Jenkins, and Victor
  Prisacariu.
\newblock Correspondence networks with adaptive neighbourhood consensus.
\newblock {\em CoRR}, abs/2003.12059, 2020.

\bibitem{megadepth}
Zhengqi Li and Noah Snavely.
\newblock Megadepth: Learning single-view depth prediction from internet
  photos.
\newblock In {\em 2018 {IEEE} Conference on Computer Vision and Pattern
  Recognition, {CVPR} 2018, Salt Lake City, UT, USA, June 18-22, 2018}, pages
  2041--2050, 2018.

\bibitem{LiangFGLCQZZ18}
Zhengfa Liang, Yiliu Feng, Yulan Guo, Hengzhu Liu, Wei Chen, Linbo Qiao,
  Li~Zhou, and Jianfeng Zhang.
\newblock Learning for disparity estimation through feature constancy.
\newblock In {\em 2018 {IEEE} Conference on Computer Vision and Pattern
  Recognition, {CVPR} 2018, Salt Lake City, UT, USA, June 18-22, 2018}, pages
  2811--2820, 2018.

\bibitem{coco}
Tsung{-}Yi Lin, Michael Maire, Serge~J. Belongie, Lubomir~D. Bourdev, Ross~B.
  Girshick, James Hays, Pietro Perona, Deva Ramanan, Piotr Doll{\'{a}}r, and
  C.~Lawrence Zitnick.
\newblock Microsoft {COCO:} common objects in context.
\newblock {\em CoRR}, abs/1405.0312, 2014.

\bibitem{LiuYT11}
Ce~Liu, Jenny Yuen, and Antonio Torralba.
\newblock {SIFT} flow: Dense correspondence across scenes and its applications.
\newblock {\em {IEEE} Trans. Pattern Anal. Mach. Intell.}, 33(5):978--994,
  2011.

\bibitem{PRNet}
Jinlu Liu and Yongqiang Qin.
\newblock Prototype refinement network for few-shot segmentation.
\newblock {\em CoRR}, abs/2002.03579, 2020.

\bibitem{Melekhov2019}
Iaroslav Melekhov, Aleksei Tiulpin, Torsten Sattler, Marc Pollefeys, Esa Rahtu,
  and Juho Kannala.
\newblock {DGC-Net}: Dense geometric correspondence network.
\newblock In {\em Proceedings of the IEEE Winter Conference on Applications of
  Computer Vision (WACV)}, 2019.

\bibitem{relu}
Vinod Nair and Geoffrey~E. Hinton.
\newblock Rectified linear units improve restricted boltzmann machines.
\newblock In {\em Proceedings of the 27th International Conference on Machine
  Learning (ICML-10), June 21-24, 2010, Haifa, Israel}, pages 807--814, 2010.

\bibitem{NoceWrig06}
Jorge Nocedal and Stephen~J. Wright.
\newblock {\em Numerical Optimization}.
\newblock Springer, New York, NY, USA, second edition, 2006.

\bibitem{STM}
Seoung~Wug Oh, Joon-Young Lee, Ning Xu, and Seon~Joo Kim.
\newblock Video object segmentation using space-time memory networks.
\newblock {\em Proceedings of the IEEE International Conference on Computer
  Vision}, 2019.

\bibitem{Pang2017}
Jiahao Pang, Wenxiu Sun, Jimmy S.~J. Ren, Chengxi Yang, and Qiong Yan.
\newblock Cascade residual learning: {A} two-stage convolutional neural network
  for stereo matching.
\newblock {\em CoRR}, abs/1708.09204, 2017.

\bibitem{Ranjan2017}
Anurag Ranjan and Michael~J. Black.
\newblock Optical flow estimation using a spatial pyramid network.
\newblock In {\em 2017 {IEEE} Conference on Computer Vision and Pattern
  Recognition, {CVPR} 2017, Honolulu, HI, USA, July 21-26, 2017}, pages
  2720--2729, 2017.

\bibitem{Rocco2017a}
Ignacio Rocco, Relja Arandjelovic, and Josef Sivic.
\newblock Convolutional neural network architecture for geometric matching.
\newblock In {\em 2017 {IEEE} Conference on Computer Vision and Pattern
  Recognition, {CVPR} 2017, Honolulu, HI, USA, July 21-26, 2017}, pages 39--48,
  2017.

\bibitem{Rocco2018a}
Ignacio Rocco, Relja Arandjelovic, and Josef Sivic.
\newblock End-to-end weakly-supervised semantic alignment.
\newblock In {\em 2018 {IEEE} Conference on Computer Vision and Pattern
  Recognition, {CVPR} 2018, Salt Lake City, UT, USA, June 18-22, 2018}, pages
  6917--6925, 2018.

\bibitem{Rocco2018b}
Ignacio Rocco, Mircea Cimpoi, Relja Arandjelovic, Akihiko Torii, Tom{\'{a}}s
  Pajdla, and Josef Sivic.
\newblock Neighbourhood consensus networks.
\newblock In {\em Advances in Neural Information Processing Systems 31: Annual
  Conference on Neural Information Processing Systems 2018, NeurIPS 2018, 3-8
  December 2018, Montr{\'{e}}al, Canada.}, pages 1658--1669, 2018.

\bibitem{SarlinDMR20}
Paul{-}Edouard Sarlin, Daniel DeTone, Tomasz Malisiewicz, and Andrew
  Rabinovich.
\newblock Superglue: Learning feature matching with graph neural networks.
\newblock In {\em 2020 {IEEE/CVF} Conference on Computer Vision and Pattern
  Recognition, {CVPR} 2020, Seattle, WA, USA, June 13-19, 2020}, pages
  4937--4946, 2020.

\bibitem{SchonbergerF16}
Johannes~L. Sch{\"{o}}nberger and Jan{-}Michael Frahm.
\newblock Structure-from-motion revisited.
\newblock In {\em 2016 {IEEE} Conference on Computer Vision and Pattern
  Recognition, {CVPR} 2016, Las Vegas, NV, USA, June 27-30, 2016}, pages
  4104--4113, 2016.

\bibitem{johannes2017}
Johannes~L. Sch{\"{o}}nberger, Marc Pollefeys, Andreas Geiger, and Torsten
  Sattler.
\newblock Semantic visual localization.
\newblock In {\em 2018 {IEEE} Conference on Computer Vision and Pattern
  Recognition, {CVPR} 2018, Salt Lake City, UT, USA, June 18-22, 2018}, pages
  6896--6906. {IEEE} Computer Society, 2018.

\bibitem{ETH3d}
Thomas Sch{\"{o}}ps, Johannes~L. Sch{\"{o}}nberger, Silvano Galliani, Torsten
  Sattler, Konrad Schindler, Marc Pollefeys, and Andreas Geiger.
\newblock A multi-view stereo benchmark with high-resolution images and
  multi-camera videos.
\newblock In {\em 2017 {IEEE} Conference on Computer Vision and Pattern
  Recognition, {CVPR} 2017, Honolulu, HI, USA, July 21-26, 2017}, pages
  2538--2547, 2017.

\bibitem{shen2019ransacflow}
Xi~Shen, Fran{\c{c}}ois Darmon, Alexei~A Efros, and Mathieu Aubry.
\newblock Ransac-flow: generic two-stage image alignment.
\newblock In {\em 16th European Conference on Computer Vision}, 2020.

\bibitem{Shewchuk}
Jonathan~R Shewchuk.
\newblock An introduction to the conjugate gradient method without the
  agonizing pain.
\newblock Technical report, USA, 1994.

\bibitem{SimonyanZ14}
Karen Simonyan and Andrew Zisserman.
\newblock Two-stream convolutional networks for action recognition in videos.
\newblock {\em CoRR}, abs/1406.2199, 2014.

\bibitem{Sun2018}
Deqing Sun, Xiaodong Yang, Ming{-}Yu Liu, and Jan Kautz.
\newblock Pwc-net: Cnns for optical flow using pyramid, warping, and cost
  volume.
\newblock In {\em 2018 {IEEE} Conference on Computer Vision and Pattern
  Recognition, {CVPR} 2018, Salt Lake City, UT, USA, June 18-22, 2018}, pages
  8934--8943, 2018.

\bibitem{Sun2019}
Deqing Sun, Xiaodong Yang, Ming-Yu Liu, and Jan Kautz.
\newblock Models matter, so does training: An empirical study of cnns for
  optical flow estimation.
\newblock {\em IEEE transactions on pattern analysis and machine intelligence},
  2019.

\bibitem{Taira2018}
Hajime Taira, Masatoshi Okutomi, Torsten Sattler, Mircea Cimpoi, Marc
  Pollefeys, Josef Sivic, Tom{\'{a}}s Pajdla, and Akihiko Torii.
\newblock Inloc: Indoor visual localization with dense matching and view
  synthesis.
\newblock In {\em 2018 {IEEE} Conference on Computer Vision and Pattern
  Recognition, {CVPR} 2018, Salt Lake City, UT, USA, June 18-22, 2018}, pages
  7199--7209. {IEEE} Computer Society, 2018.

\bibitem{Taniai2016}
Tatsunori Taniai, Sudipta~N. Sinha, and Yoichi Sato.
\newblock Joint recovery of dense correspondence and cosegmentation in two
  images.
\newblock In {\em 2016 {IEEE} Conference on Computer Vision and Pattern
  Recognition, {CVPR} 2016, Las Vegas, NV, USA, June 27-30, 2016}, pages
  4246--4255, 2016.

\bibitem{GLUNet}
Prune Truong, Martin Danelljan, and Radu Timofte.
\newblock {GLU-Net}: Global-local universal network for dense flow and
  correspondences.
\newblock In {\em 2020 {IEEE} Conference on Computer Vision and Pattern
  Recognition, {CVPR} 2020}, 2020.

\bibitem{cfnet}
Jack Valmadre, Luca Bertinetto, Jo{\~{a}}o~F. Henriques, Andrea Vedaldi, and
  Philip H.~S. Torr.
\newblock End-to-end representation learning for correlation filter based
  tracking.
\newblock In {\em 2017 {IEEE} Conference on Computer Vision and Pattern
  Recognition, {CVPR} 2017, Honolulu, HI, USA, July 21-26, 2017}, pages
  5000--5008, 2017.

\bibitem{FEELVOS}
Paul Voigtlaender and Bastian Leibe.
\newblock Feelvos: Fast end-to-end embedding learning for video object
  segmentation.
\newblock In {\em IEEE Conference on Computer Vision and Pattern Recognition
  (CVPR)}, 2019.

\bibitem{WangLZZF19}
Kaixin Wang, Jun~Hao Liew, Yingtian Zou, Daquan Zhou, and Jiashi Feng.
\newblock Panet: Few-shot image semantic segmentation with prototype alignment.
\newblock In {\em 2019 {IEEE/CVF} International Conference on Computer Vision,
  {ICCV} 2019, Seoul, Korea (South), October 27 - November 2, 2019}, pages
  9196--9205, 2019.

\bibitem{Xiao2020}
Taihong Xiao, Jinwei Yuan, Deqing Sun, Qifei Wang, Xin{-}Yu Zhang, Kehan Xu,
  and Ming{-}Hsuan Yang.
\newblock Learnable cost volume using the cayley representation.
\newblock {\em CoRR}, abs/2007.11431, 2020.

\bibitem{YangR19}
Gengshan Yang and Deva Ramanan.
\newblock Volumetric correspondence networks for optical flow.
\newblock In Hanna~M. Wallach, Hugo Larochelle, Alina Beygelzimer, Florence
  d'Alch{\'{e}}{-}Buc, Emily~B. Fox, and Roman Garnett, editors, {\em Advances
  in Neural Information Processing Systems 32: Annual Conference on Neural
  Information Processing Systems 2019, NeurIPS 2019, 8-14 December 2019,
  Vancouver, BC, Canada}, pages 793--803, 2019.

\bibitem{ZhangKRVKTSIFF18}
Yinda Zhang, Sameh Khamis, Christoph Rhemann, Julien P.~C. Valentin, Adarsh
  Kowdle, Vladimir Tankovich, Michael Schoenberg, Shahram Izadi, Thomas~A.
  Funkhouser, and Sean~Ryan Fanello.
\newblock Activestereonet: End-to-end self-supervised learning for active
  stereo systems.
\newblock In {\em Computer Vision - {ECCV} 2018 - 15th European Conference,
  Munich, Germany, September 8-14, 2018, Proceedings, Part {VIII}}, pages
  802--819, 2018.

\bibitem{Zhou2019}
Bolei Zhou, Hang Zhao, Xavier Puig, Tete Xiao, Sanja Fidler, Adela Barriuso,
  and Antonio Torralba.
\newblock Semantic understanding of scenes through the {ADE20K} dataset.
\newblock {\em Int. J. Comput. Vis.}, 127(3):302--321, 2019.

\bibitem{cavia}
Luisa~M. Zintgraf, Kyriacos Shiarlis, Vitaly Kurin, Katja Hofmann, and Shimon
  Whiteson.
\newblock Fast context adaptation via meta-learning.
\newblock In {\em Proceedings of the 36th International Conference on Machine
  Learning, {ICML} 2019, 9-15 June 2019, Long Beach, California, {USA}}, pages
  7693--7702, 2019.

\end{thebibliography}
}

\newpage
\appendix
\begin{center}
	\textbf{\Large Appendix}
\end{center}

In this appendix, we first provide the details of the derivation of the filter map $\wtm$ within the filter map predictor module $\wpred$ in Section \ref{Sec:deriv}. We then give the expression for the initial estimate $\wt^0$ in Section \ref{sec:sup-initializer}. In Section \ref{Sec:arch-details}, we provide further insights on the architecture of our \dicor module as well as the implementation details.
In Section \ref{Sec:details-evaluation}, we give more details about the evaluation datasets, metrics and networks utilized. 
We then provide additional quantitative and qualitative results of our approach \dicor compared to the feature correlation layer in Section \ref{Sec:sup-results}. Finally, we further analyse our approach in an extended ablation study in Section \ref{Sec:sup-ablation}.

\section{Derivation of filter map prediction module $\wpred$}
\label{Sec:deriv}

Here, we derive the iterative updates employed in our module $\wpred_\theta$, which aims to solve $\wtm = \wpred_\theta(\ftr, \ftq) = \argmin_\wt L(\wt; \ftr, \ftq, \theta)$ (Eq.~\ref{eq:minimize-loss}). Our final objective (Eqs.~\ref{eq:ref-loss}-\ref{eq:total-loss}) is given by,
\begin{subequations}
\begin{align}
L(\wt; \ftr, \ftq, \theta) &= \Lr(\wt; \ftr, \theta) + \Lq(\wt; \ftq, \theta) + \|\lambda_\theta w\|^2  \label{eq:sup-total-loss}\\
\Lr(\wt; \ftr, \theta) &= \left\| \sigma_\eta\big(\corr(\wt, \ftr); \, \ps, \ns\big) - \y \right\|^2 \label{eq:sup-ref-loss}\\
\Lq(\wt; \ftq, \theta) &= \left\| R_\theta \ast \corr(\wt, \ftq) \right\|^2 \,. \label{eq:sup-query-loss}
\end{align}
\end{subequations}
As discussed in Sec.~\ref{subsec:optim}, we do not need to attain a global optimum. The goal is to significantly minimize the loss $L$, using only a few iterations for efficiency. To this end we employ the Steepest Descent methodology~\cite{NoceWrig06,Shewchuk}. In the steepest descent algorithm, we update the parameters by taking steps $\wt^{n+1}=\wt^{n}-\alpha^{n} \nabla L\left(\wt^{n}\right)$ in the gradient direction $\nabla L\left(\wt^{n}\right)$ with step length $\alpha^n$. The aim is to find the step length $\alpha^n$ that leads to a  maximal decrease in the objective. This is performed by first approximating the loss with a quadratic function at the current estimate $\wt^n$,
\begin{equation}
\label{eq:sup-taylor}
L(\wt) \approx L^n_\text{GN}(\wt)= \frac{1}{2}\left(\wt-\wt^{n}\right)^{\mathrm{T}} Q^{n}\left(\wt-\wt^{n}\right)+
\left(\wt-\wt^{n}\right)^{\mathrm{T}} \nabla L\left(\wt^{n}\right)+L\left(\wt^{n}\right)
\end{equation}
Here, we see $\wt^n$ as a vector. We set the Hermitian positive definite matrix $Q^n$ according to the Gauss-Newton method~\cite{NoceWrig06}  $Q^n=(J^n)\tp J^n$, where $J^n$ is the Jacobian of the residual at $\wt^n$. 
To avoid clutter, the dependence on $\ftr, \ftq, \theta$ is made implicit. In the rest of the section, unless otherwise stated, matrices multiplications are element-wise. 

The steepest descent method~\cite{NoceWrig06, Shewchuk} finds the step-length $\alpha^n$ that minimizes the loss \eqref{eq:sup-taylor} in the gradient direction. Due to the convexity of \eqref{eq:sup-taylor}, this is obtained by solving $\frac{\mathrm{d}}{\mathrm{d} \alpha} L_\text{GN}\left(\wt^n-\alpha \nabla L\left(\wt^n\right)\right)=0$, which leads to the expression,
\begin{equation}
\label{eq:sup-alpha}
\alpha^n=\frac{\nabla L\left(\wt^{n}\right)^{\mathrm{T}} \nabla L\left(\wt^{n}\right)}{\nabla L\left(\wt^{n}\right)^{\mathrm{T}} Q^n \nabla L\left(\wt^{n}\right)}=\frac{\left\|\nabla L\left(\wt^{n}\right)\right\|^2}{\left\| J^n \nabla L\left(\wt^{n}\right)\right\|^2}
\end{equation}

In the next subsections, we derive the expression for $\nabla L$ and subsequently for step-length $\alpha$.

\subsection{Closed form expression of $\nabla L$}
Here, we derive a closed-form expression for the gradient of the loss \eqref{eq:sup-total-loss}.
The gradient $\nabla L (\wt)$ of the loss \eqref{eq:sup-total-loss} with respect to the filters $\wt$ is then computed as, 
\begin{equation}
\nabla L(\wt) = \nabla \Lr(\wt) + \nabla \Lq(\wt) + 2\lambda_\theta^2 \wt\,.
\end{equation}

\parsection{Expression of $\nabla \Lr(\wt)$}
$\Lr$ is defined according to \eqref{eq:sup-ref-loss} and equation \eqref{eq:ref-loss} of the main paper, such as $L_r = \left\| r_r(\wt, \ftr) \right\|^2$. Here, $r_{r}$ designates the residual function, which is formulated as (also Eq. \ref{eq:error-func}),
\begin{align}
\label{eq:sup-ref-res}
    r_{r}(\wt, f^r) = & \,\sigma_\eta\big(\corr(\wt, \ftr); \, \ps, \ns\big) - \y \\
    \sigma_\eta\big(\corr(\wt, \ftr); \, \ps, \ns\big) = & \,\frac{\ps - \ns}{2}\! \left(\!\sqrt{\corr(\wt, \ftr)^2 + \eta^2} - \eta\right) + \frac{\ps + \ns}{2}  \corr(\wt, \ftr) \,. \label{eq:sup-error-func}
\end{align}

The gradient of $\nabla \Lr(\wt)$ of the loss \eqref{eq:sup-ref-loss} w.r.t $\wt$ is given by:
\begin{equation}
\label{eq:deriv-ref-loss}
   \nabla \Lr(\wt) = 2\left[\frac{\partial r_r(\wt, f^r)}{\partial \wt} \right]\tp r_r(\wt, f^r)
\end{equation}
where $J_r = \frac{\partial r_r(\wt, f^r)}{\partial \wt}$ corresponds to the Jacobian of the residual function \eqref{eq:sup-ref-res} with respect to filters \wt. Using the chain rule we obtain, 
\begin{equation}
\label{eq:deriv_residual}
    J_r = \frac{\partial r_r(\wt, f^r)}{\partial \wt}  = \frac{\partial \sigma_\eta}{ \partial \corr(\wt, \ftr)} \frac{\partial \corr(\wt, \ftr)}{ \partial \wt} \,,
\end{equation}
Using \eqref{eq:sup-error-func}, the derivative of the error function is obtained as
\begin{equation}
\label{eq:error-func-deriv}
    \frac{\partial \sigma_\eta}{ \partial \corr(\wt, \ftr)} = \left[\frac{\ps - \ns}{2}  \left( \frac{\corr(\wt, \ftr)}{\sqrt{\corr(\wt, \ftr)^2 + \eta^2}} \right) + \frac{\ps + \ns}{2} \right]\,.
\end{equation}

Integrating \eqref{eq:deriv_residual} into \eqref{eq:deriv-ref-loss} leads to the final formulation of $\nabla \Lr(\wt)$ as
\begin{equation}
\label{eq:final-deriv-ref-loss}
   \nabla \Lr(\wt) = 2\left[\frac{\partial \corr(\wt, \ftr)}{\partial \wt}\right]\tp 
   \left[ \left( \frac{\ps - \ns}{2}    \left( \frac{\corr(\wt, \ftr) - \y }{\sqrt{(\corr(\wt, \ftr) - \y)^2 + \eta^2}} \right) + \frac{\ps + \ns}{2}\right) r_r(\wt, \ftr)  \right] \,.
\end{equation}
The multiplication with the transposed Jacobian $\left[\frac{\partial \corr(\wt, \ftr)}{\partial \wt}\right]\tp$ corresponds to back-propagation through the correlation layer $\corr$. This can be efficiently implemented with standard operations. 

\parsection{Expression of $\nabla \Lq$}
The loss on the query frame $\Lq$ is formulated in \eqref{eq:sup-query-loss} and in eq. \ref{eq:query-loss} of the main paper, as $\Lq(\wt) = \left\| r_q(\wt, f^q) \right\|^2$, where the residual $r_q$ is defined below:
\begin{equation}
\label{L_q_residual}
    r_q(\wt, f^q) = R_\theta \ast \corr(\wt, \ftq)
\end{equation}

Following similar steps than for $\Lr$, the gradient $\nabla \Lq$ of the loss \eqref{eq:sup-query-loss} \wrt the filters $\wt$ is then computed as,
\begin{equation}
    \nabla \Lq = 2 \left[\frac{\partial  r_q(\wt, f^q)}{\partial \wt} \right]\tp r_q(\wt, f^q)
\end{equation}
where $J_q = \frac{\partial r_q(\wt, f^q)}{\partial \wt}$ corresponds to the Jacobian of the residual function \eqref{L_q_residual} with respect to the filter map $\wt$. 
\begin{equation}
\label{J_q}
J_q = R_\theta \ast \frac{\partial \corr(\wt, \ftq)}{\partial \wt} \,.
\end{equation}

This leads to the final formulation of the gradient as,
\begin{equation}
\label{eq:sup-gradient-Lq}
    \nabla \Lq = 2 \left[\frac{\partial \corr(\wt, \ftq)}{\partial \wt}\right]\tp \left[R_\theta \ast\right]\tp  r_q(\wt, \ftq) \,.
\end{equation}
Here, $\left[R_\theta \ast\right]\tp$ denotes the transposed convolution with the kernel $R_\theta$.

\subsection{Calculation of step-length $\alpha^n$}

In this section, we show the calculation of the denominator of $\alpha^n = \frac{\alpha^n_\text{num}}{\alpha^n_\text{den}}$. The denominator in equation \eqref{eq:sup-alpha} is given by,
\begin{equation}\begin{aligned}
    \alpha^n_\text{den} = & \left \| J^n \nabla L\left(\wt^{n}\right) \right \|^2 \\
    = & \left\|\left. J_r(\wt)\right|_{\wt^{n}} \nabla L\left(\wt^{n}\right)\right\|^{2} +
    \left\|  \left. J_q(\wt)\right|_{\wt^{n}}  \nabla L\left(\wt^{n}\right)\right\|^{2}
    +\left\|\lambda_\theta \nabla L\left(\wt^{n}\right)\right\|^{2}
\end{aligned}\end{equation}

Using equations \eqref{eq:deriv_residual} and \eqref{J_q}, we finally obtain:
\begin{align}
\label{eq:alpha-den}
\alpha^n_\text{den} = &\,\left\|\left.\     \frac{\partial \sigma_\eta}{ \partial \corr(\wt, \ftr)}   \frac{\partial \corr(\wt, \ftr)}{ \partial \wt}   \right|_{\wt^{n}} \nabla L\left(\wt^{n}\right)\right\|^{2} +
\left\|\left.  R_\theta \ast \frac{\partial \corr(\wt, \ftq)}{\partial \wt}  \right|_{\wt^{n}} \nabla L\left(\wt^{n}\right)\right\|^{2}
+\left\|\lambda_\theta \nabla L\left(\wt^{n}\right)\right\|^{2} \nonumber \\
= & \, \left\| \frac{\partial \sigma_\eta}{ \partial \corr(\wt, \ftr)} \corr(\nabla L\left(\wt^{n}\right), \ftr)\right\|^{2} +
\left\| R_\theta \ast \corr(\nabla L\left(\wt^{n}\right), \ftq)\right\|^{2}
+\left\|\lambda_\theta \nabla L\left(\wt^{n}\right)\right\|^{2}
\end{align}
The relation $\left.\frac{\partial \corr(\wt, \ftr)}{ \partial \wt} \right|_{\wt^{n}} \nabla L\left(\wt^{n}\right) = \corr(\nabla L\left(\wt^{n}\right), \ftr)$ stems from the linearity of $\corr$ in the first argument.
All operations in \eqref{eq:alpha-den} can thus easily be implemented using standard neural network operations.
We summarize the different steps taking place within the filter map predictor  module $\wpred$ in algorithm~\ref{alg:model-predictor}. 

\newcommand{\init}{\mathtt{ModelInit}}
\newcommand{\filtergrad}{\mathtt{FiltGrad}}
\newcommand{\assign}{\leftarrow}
\newcommand{\algcomment}[2]{\hspace{#2mm}{\footnotesize\# #1}}
\begin{algorithm}[t]
	\caption{Filter Predictor module $P$.}
	\begin{algorithmic}[1]
		\Require Reference and Query feature maps $\ftr, \ftq$, iterations $N_\text{iter}$
		\State $\wt^{0} \assign \init(\ftr)$ \algcomment{Initialize filter map (sec. \ref{subsec:optim} of main paper)}{8}
		\For{$i = 0, \ldots, N_\text{iter} - 1$} \algcomment{Optimizer module loop}{8}
		\State $\nabla L(\wt^{n}) \assign \filtergrad(\wt^{n}, \ftr, \ftq)$ \algcomment{Using \eqref{eq:final-deriv-ref-loss} - \eqref{eq:sup-gradient-Lq}}{3}

		\State $\alpha^n_\text{num} \assign \left \|\nabla L\left(\wt^{n}\right) \right \|^2$
		
		\State $\alpha^n_\text{den} \assign \left \| J^n \nabla L\left(\wt^{n}\right) \right \|^2$ \algcomment{Apply Jacobian \eqref{eq:deriv_residual} and \eqref{J_q}}{14}
		
		\State $\alpha^n \assign \alpha^n_\text{num} / \alpha^n_\text{den} $
		\algcomment{Compute step length \eqref{eq:sup-alpha}}{5.5}
		\State $\wt^{n+1} \assign \wt^{n} - \alpha^n \nabla L(\wt^{n})$ \algcomment{Update filter map}{5}
		\EndFor
	\end{algorithmic}
	\label{alg:model-predictor}
\end{algorithm}

\section{Initial estimate of $w^0$}
\label{sec:sup-initializer}

As explained in Section \ref{subsec:optim} of the main paper, to reduce the number of optimization iterations needed in the filter predictor network $\wpred$, we generate an initial filter map $\wt^0$, which is then processed by the optimizer module to provide the final discriminative filter $\wtm=\wpred(\ftr, \ftq)$. 

We wish that $\wt^0$ integrates information about the entire reference feature map $\ftr$.  We thus formulate $w^0$ at location $(i,j)$ as a linear combination of $f^r_{ij}$ and $\bar{f^r}$,  where $\bar{\ftr} \in \reals^D$ is the spatial average reference vector, encoding contextual information. We obtain $\wt^0$ by solving for the scalar factors $a_{ij},b_{ij}$ that adhere to the following constraints,
\begin{subequations}
\label{eq:initializer}
\begin{align}
&\wt^0_{ij}  = a_{ij}\ftr_{ij} + b_{ij}\bar{\ftr} \\
&(\wt^0_{ij})\tp\ftr_{ij}  = \beta  \\
&(\wt^0_{ij})\tp\bar{\ftr}  = \gamma 
\end{align}
\end{subequations}
In the simplest setting, $\beta$ can be set to one and $\gamma$ to zero. However, we let these values be learnt from data. 
The scalar coefficients $a_{ij}$ and $b_{ij}$ are then easily found by solving these equations, resulting in the following formulation for $\wt^0$, 
\begin{equation}
w^0_{ij} = \frac{ \left[\beta \left \| \bar{f^r} \right \|^2 - \gamma (f^r_{ij}) \tp \bar{f^r} \right]f^r_{ij} - \left[\beta (f^r_{ij})\tp  \bar{f^r} - \gamma \left \| f^r_{ij} \right \|^2 \right] \bar{f^r}}   {\left \| \bar{f^r} \right \|^2\left \| f^r_{ij} \right \|^2 - \left( (f^r_{ij})\tp \bar{f^r} \right)^2}
\end{equation}

As already mentioned, $\beta$ and $\gamma$ are learnable weights. In the simplest case, both are just scalars. We call this version \textbf{ContextAwareInitializer}. To add further flexibility, they can alternatively be vectors of the same dimension $D$ than the reference feature map $\ftr \in \reals^{H \times W \times D}$, such that $\beta, \gamma \in \reals^D$. We refer to this variant of the initializer as \textbf{Flexible-ContextAwareInitializer}. 
Both versions explicitly integrate context information about the entire reference feature map. 

We additionally define a simpler alternative for $\wt^0$, that we call \textbf{SimpleInitializer} for which it is assumed that $(f^r_{ij})\tp \bar{f^r} = 0$. As a result, $\wt_{ij}$ only depends on the reference feature $f^r_{ij}$ at this location, $w^0$ can thus be formulated as:
\begin{equation}
w^0_{ij} = \beta \frac{f^r_{ij}} {\left \| f^r_{ij} \right \|} 
\end{equation}
Here, $\beta$ can also be either a scalar (\textbf{SimpleInitializer}) or a vector of dimension $D$ (\textbf{Flexible-SimpleInitializer}). 

In our Global-\dicor module, we use the variant Flexible-ContextAwareInitializer for our initializer module. We defend this choice in our supplementary ablation study Section \ref{Sec:sup-ablation}. For our Local-\dicor module, we instead use the SimpleInitializer variant of the initializer.

\section{Architecture details}
\label{Sec:arch-details}

\parsection{Expression for $\y, \ns, \ps$} Here we discuss the parametrization of $y, \ns, \ps$, introduced in the reference loss formulation in Sec. \ref{subsec:loss-ref}.

For the implementation, we define $y = \ps y'$ and $\ns = \ps m$, with element-wise multiplication. 
We parametrize $y', \ps, m$ as functions of the distance $d_{ijkl} = \sqrt{(i-k)^2 + (j-l)^2}$ between $w_{ij}$ and the example $\ftr_{kl}$, such that $y'_{ijkl}=y'_{\theta}(d_{ijkl}), \, \ps_{ijkl}=\psl(d_{ijkl}),\, m_{ijkl}=m_{\theta}(d_{ijkl})$.

All three are expressed with triangular basis function, as in \cite{dimp}. For example, the function $\ps$ at position $(i,j,k,l)$ is given by:
\begin{equation}
\label{eq:triangular-func}
    \ps_{ijkl}=\sum_{k=0}^{N-1} (\ps_\theta)^{k} \rho_{k}(d_{ijkl})
\end{equation}
with triangular basis functions $\rho_{k}$, expressed as
\begin{equation}
    \rho_{k}(d)=\left\{\begin{array}{ll}
    \max \left(0,1-\frac{|d-k \Delta|}{\Delta}\right), & k<N-1 \\
    \max \left(0, \min \left(1,1+\frac{d-k \Delta}{\Delta}\right)\right), & k=N-1
    \end{array}\right.
\end{equation}
We use $N = 10$ basis functions and set the knot displacement to $\Delta = 0.5$ in the resolution of the deep feature space. The final case $k=N-1$ represents all locations $(k,l)$ that are far away from $(i,j)$ and thus can be treated identically.

The coefficients $y'_{\theta}, \psl, m_{\theta}$ are learnt from data, as part of the filter predictor module $\wpred$. For $m$, we constrain the values in the interval $\left [ 0, 1 \right]$ by passing the output of \eqref{eq:triangular-func} through a Sigmoid function. 
We initialize the target confidence $y'_{\theta}$ to a Gaussian, with mean equal to 0 and standard deviation equal to 1. 
The positive weight function $\psl$ is initialized to a constant so that $\ps_{ijkl} = 1$ while we initialize the function $m_{\theta}$ with a scaled tanh function. 

\begin{figure*}[t]
\centering
\includegraphics[width=0.80\textwidth]{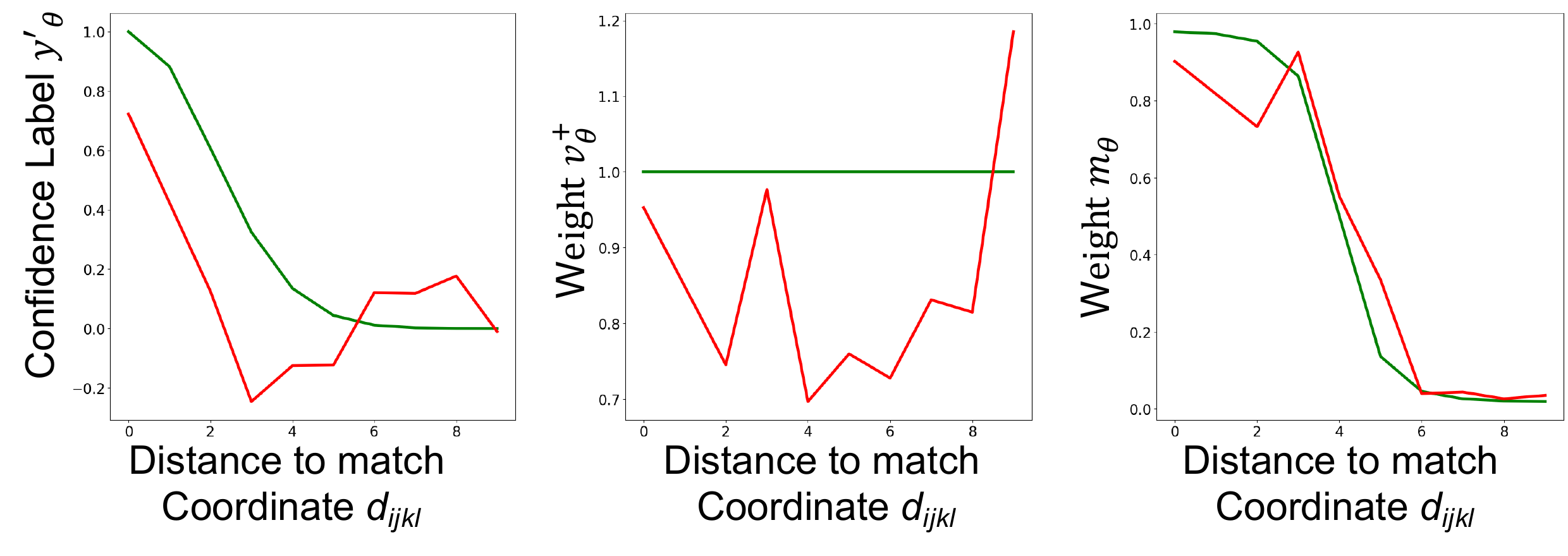}
\vspace{-3mm}
\caption{Plot of the learnt target confidence $y'_{\theta}$ and weights $\psl, m_{\theta}$.  The learnt values are shown in red while the initialization of each function is presented in green.}
\label{fig:network-weights}
\end{figure*}

\begin{figure}[t]
\centering%
\newcommand{\wid}{0.25\textwidth}
\subfloat[Initial Label $\y_{ij..}$]{\includegraphics[width=\wid]{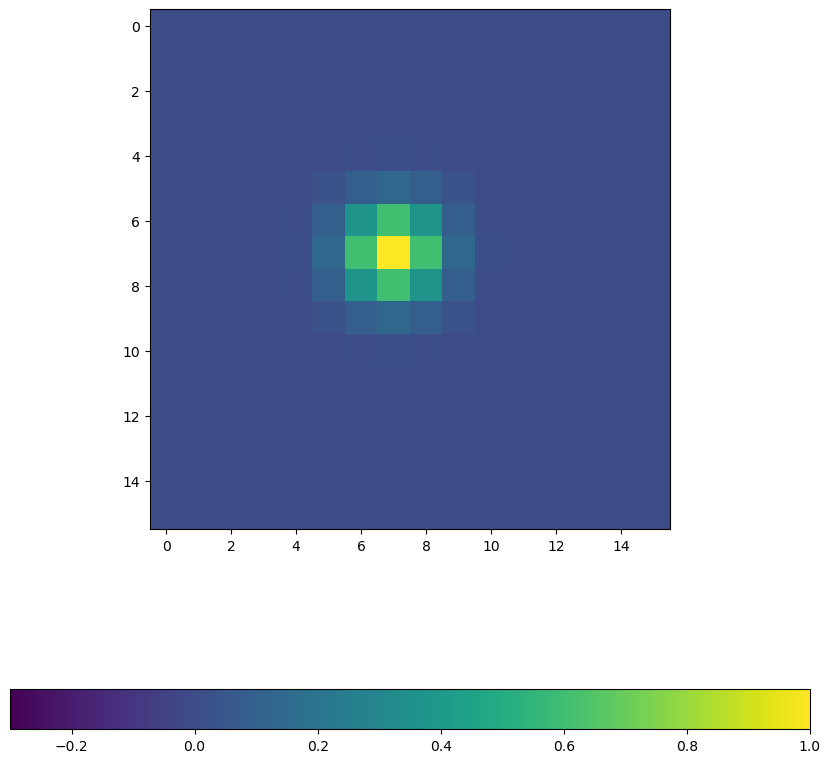}}~%
\subfloat[Initial Weight $\ps_{ij..}$]{\includegraphics[width=\wid]{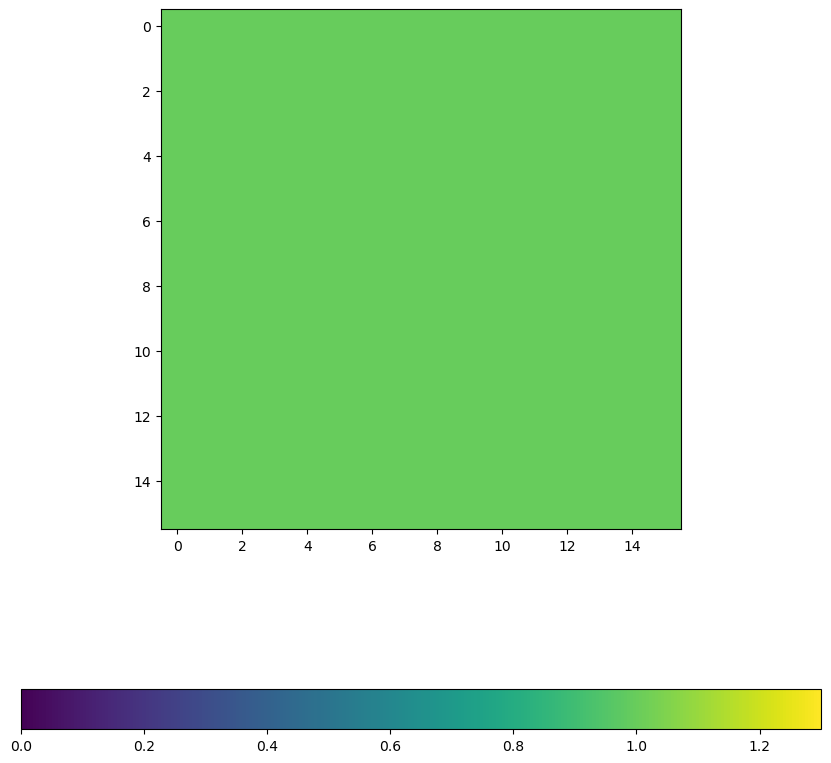}}~%
\subfloat[Initial Weight $\ns_{ij..}$]{\includegraphics[width=\wid]{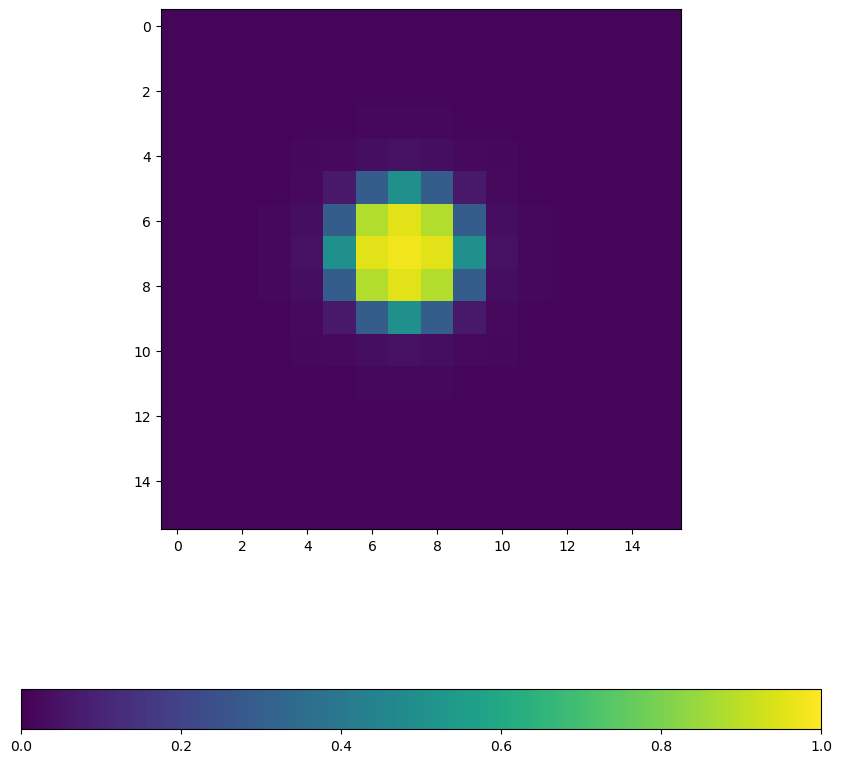}}~%
\\

\subfloat[Learnt Label $\y_{ij..}$]{\includegraphics[width=\wid]{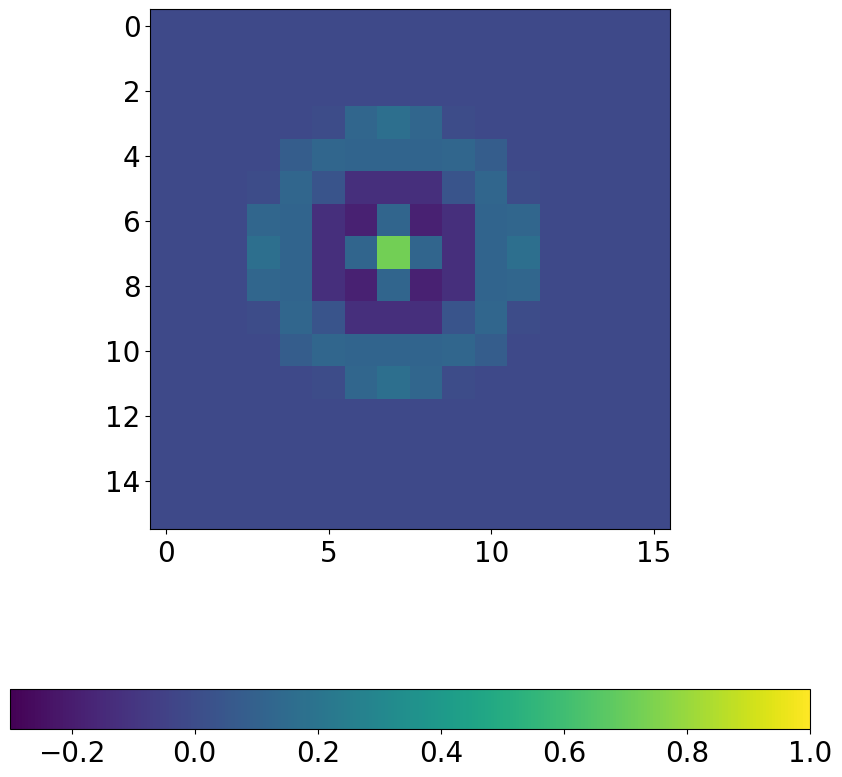}}~%
\subfloat[Learnt Weight $\ps_{ij..}$]{\includegraphics[width=\wid]{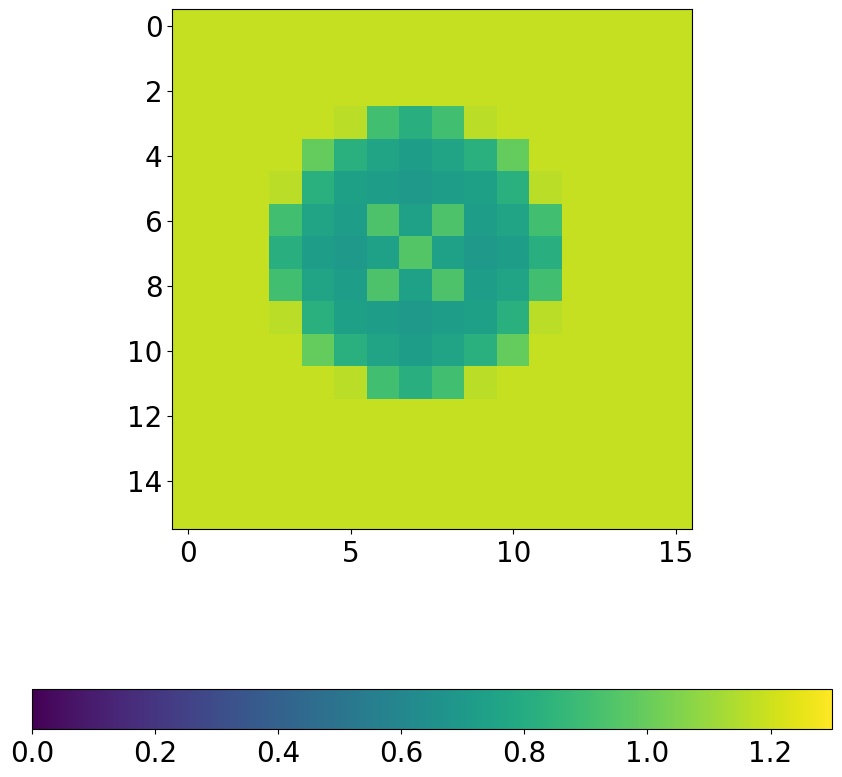}}~%
\subfloat[Learnt Weight $\ns_{ij..}$]{\includegraphics[width=\wid]{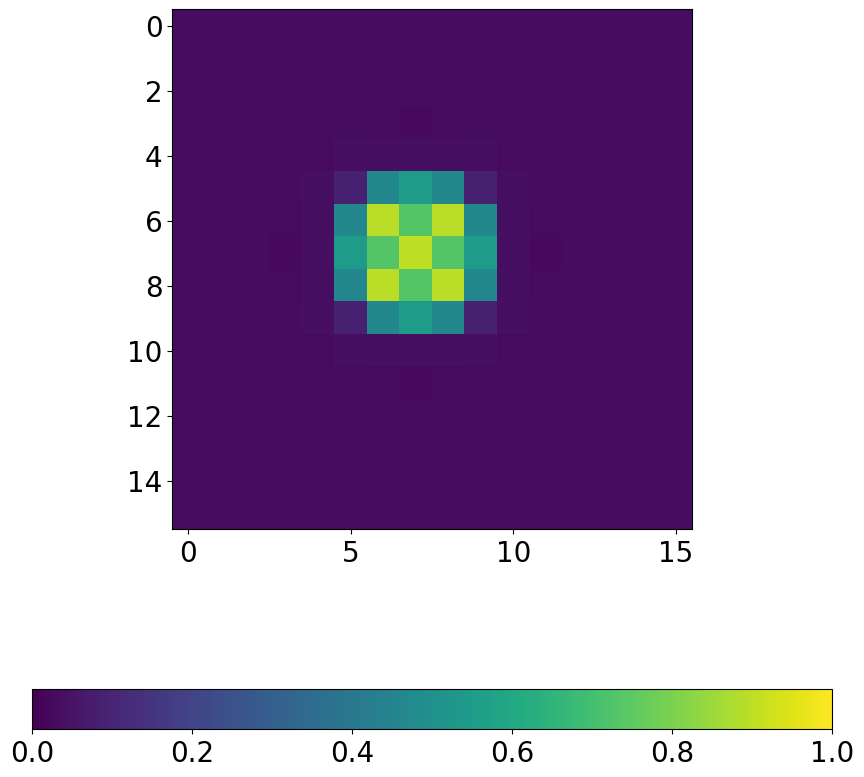}}~%
\vspace{-2mm}
\caption{Visualization of the heat maps corresponding to the learnt target confidence $y$ and weights $\ns, \ps$ for a particular location $(i,j)=(7,7)$. (a), (b) and (c) show the initialization of each function while (d), (e) and (f) depict the learnt values. } 
\label{fig:net-weights}\vspace{-5mm}
\end{figure}

The initial and learnt values for $y'_{\theta}, \psl, m_{\theta}$ of our Global-\dicor module are visualized in Figure~\ref{fig:network-weights}. They result from the training of GLU-Net-\dicor on the synthetic \textit{Dynamic} training dataset. 
We additionally provide the visualization of $y_{ij..}, \ns_{ij..}, \ps_{ij..} \in \reals^{H \times W}$ as heat-maps for a particular location $(i,j)$ in Figure \ref{fig:net-weights}.

\parsection{Smoothness operator $\mathbf{R_\theta}$}  We now focus on the operator $R_\theta$, introduced in the loss formulation on the query image in Sec. \ref{subsec:smooth-loss}. $R_\theta \in \reals^{K^4\times Q}$ is a learnable 4D-kernel of spatial size $K$ and $Q$ number of output channels. We set $K=3$ and $Q = 16$ output channels. For implementation purposes, the 4-D convolution is factorized as two consecutive 2-D convolutional layers, operating over the two first and two latter dimensions respectively. The output dimension of the first 2D-convolution is also set to 16. 
Note that the kernel $R_{\theta}$ is learnt, along with all other network parameters, by the SGD-based minimization of the same final network training loss used in the GLU-Net and PWC-Net baselines. This is contrary to the filter map $\wt$ that is optimized using Eq.~\ref{eq:ref-loss} and \ref{eq:query-loss} at each forward pass of the network.

\section{Experimental setup and datasets}
\label{Sec:details-evaluation}

In this section, we first provide details about the evaluation metrics and datasets. We then explain the procedure used to create the \textit{Dynamic} dataset, utilized for training state-of-the-art GLU-Net. 
Finally, we detail the architecture of our baseline network, used for our ablation study, namely BaseNet. 

\subsection{Evaluation metrics}

\parsection{AEPE} AEPE is defined as the Euclidean distance between estimated and ground truth flow fields, averaged over all valid pixels of the reference image. 

\parsection{PCK} The Percentage of Correct Keypoints (PCK) is computed as the percentage per image, of correspondences $\mathbf{\tilde{x}}_{j}$ with an Euclidean distance error $\left \| \mathbf{\tilde{x}}_{j} - \mathbf{x}_{j}\right \| \leq  T$, w.r.t.\ to the ground truth $\mathbf{x}_{j}$, that is smaller than a threshold $T$. 
Note that all PCK percentages presented, except for the ones reported on MegaDepth,   are computed per image and then averaged over all image pairs of the dataset. On MegaDepth however, following \cite{shen2019ransacflow}, the PCK are calculated over all valid pixels of the dataset. 

\parsection{F1} F1 designates the percentage of outliers averaged over all valid pixels of the dataset \cite{Geiger2013}. They are defined as follows, where $F_{gt}$ indicates the ground-truth flow field and $F$ the estimated flow by the network.
\begin{equation}
    F1 = \frac{    \left \| F-F_{gt} \right \| > 3  \textrm{ and }   \frac{\left \| F-F_{gt} \right \|}{\left \|F_{gt} \right \|} > 0.05 } {\textrm{\#valid pixels}}
\end{equation}

\subsection{Evaluation datasets}
\label{details-eval-data}

\parsection{HP} The HPatches dataset~\cite{Lenc} is a benchmark for geometric matching correspondence estimation. It depicts planar scenes, with transformations restricted to homographies.  We only employ the 59 sequences labelled with \verb|v_X|, which have viewpoint changes, thus excluding the ones labelled \verb|i_X|, which only have illumination changes. Each image sequence contains a query image and 5 reference images taken under increasingly larger viewpoints changes, with sizes ranging from $450 \times 600$ to $1613 \times 1210$. 

\parsection{ETH3D}  To validate our approach for real 3D scenes, where image transformations are not constrained to simple homographies, we also employ the Multi-view dataset ETH3D~\cite{ETH3d}. It contains 10 image sequences at $480 \times 752$ or $514 \times 955$ resolution, depicting indoor and outdoor scenes and resulting from the movement of a camera completely unconstrained, used for benchmarking 3D reconstruction. 
The authors additionally provide a set of sparse geometrically consistent image correspondences (generated by~\cite{SchonbergerF16}) that have been optimized over the entire image sequence using the reprojection error. 
We sample image pairs from each sequence at different intervals to analyze varying magnitude of geometric transformations, and use the provided points as sparse ground truth correspondences. This results in about 500 image pairs in total for each selected interval.
Note that the PCK and AEPE values computed per image are then averaged over all image pairs of each sequence. The final metrics presented are the averages over all sequences. 

\parsection{MegaDepth} To validate our approach on real scenes depicting extreme viewpoint changes, we use images of the MegaDepth dataset. No real ground-truth correspondences are available, so we use the result of SfM reconstructions to obtain sparse ground-truth correspondences. We follow the same procedure and test images than \cite{shen2019ransacflow}. More precisely, we use 3D points and project them onto pairs of matching images to obtain correspondences and we randomly sample 1600 pairs of images that shared more than 30 points. It results in approximately 367K correspondences. Note that for evaluation, following \cite{shen2019ransacflow}, the PCK metrics are calculated over all valid pixels of the dataset. 

\parsection{KITTI} The KITTI dataset~\cite{Geiger2013} is composed of real road sequences captured by a car-mounted stereo camera rig. The KITTI benchmark is targeted for autonomous driving applications and its semi-dense ground truth is collected using LIDAR. The 2012 set only consists of static scenes while the 2015 set is extended to dynamic scenes via human annotations.  The later contains  large motion, severe illumination changes, and occlusions. 

\parsection{Sintel} The Sintel benchmark~\cite{Butler2012} is created using the open source graphics movie “Sintel” with two passes, clean and final. The final pass contains strong atmospheric effects, motion blur, and camera noise.

\parsection{TSS} The TSS dataset~\cite{Taniai2016} contains 400 image pairs, divided into three groups: FG3DCAR, JODS, and PASCAL, according to the origins of the images. The dense flow fields annotations for the foreground object in each pair is provided along with a segmentation mask. Evaluation is done on 800 pairs, by also exchanging query and reference images. 

\subsection{Training dataset for GLU-Net based networks}
\label{details-dynamic-data}

In~\cite{GLUNet}, GLUNet is trained on \textit{DPED-ADE-CityScapes}, created by applying synthetic affine, TPS and homography transformations to real images of the DPED~\cite{Ignatov2017}, CityScapes~\cite{Cordts2016} and ADE-20K~\cite{Zhou2019} datasets. Here, we refer to this dataset as the \emph{Static} training dataset, since it simulates a static scene.  
While GLU-Net trained on the \textit{Static} dataset obtains state-of-the-art results on geometric matching and optical flow datasets (see Table \ref{tab:dataset}), the \textit{Static} dataset does not capture independently moving objects, present in optical flow data. For this reason, we introduce a \emph{Dynamic} training dataset, created from the original \textit{Static} dataset with additional random independently moving objects. To do so, these objects are sampled from the COCO dataset~\cite{coco}, and inserted on top of the images of the \textit{Static} data using their segmentation masks. To generate motion, we randomly sample affine transformation parameters for the foreground objects, which are independent of the background transformations. This can be interpreted as both the camera and the objects moving independently of each other. The \textit{Dynamic} dataset allows the network to learn the presence of independently moving objects and motion boundaries.

In Table \ref{tab:dataset}, we compare evaluation results of original GLU-Net trained on either the \textit{Static} or the \textit{Dynamic} datasets. While training on the \textit{Dynamic} data leads to worse results on HPatches, it leads to improved performances on all optical flow data, particularly significant on Sintel and KITTI-2015. Only the F1 metric on KITTI-2012 is slightly worse when training on the \textit{Dynamic} dataset. This is consistent with the fact that the \textit{Static} training dataset is in line with HPatches, both restricted to homography transformations, while the \textit{Dynamic} one is better suited for  optical flow data, that depict independently moving objects. The \textit{Dynamic} dataset is especially suitable for KITTI-2015 and Sintel, since both represent dynamic scenes, while KITTI-2012 only experiences static 3D scenes. 

Results on HPatches and TSS are computed with GLU-Net and GLU-Net-\dicor trained on the \textit{Static} dataset while results on MegaDepth, ETH3D and optical flow data KITTI and Sintel use the networks trained on the \textit{Dynamic} data instead. Here, we emphasize that both GLU-Net and GLU-Net-GOCor are trained with exactly the same procedure, introduced in \cite{GLUNet}.

\begin{table}[H]
\centering
\caption{Evaluation results of GLU-Net when trained on the \textit{Static} or the \textit{Dynamic} datasets.}
\resizebox{0.99\textwidth}{!}{%
\begin{tabular}{lcc|cc|cc|cc|cc}
\toprule
             & \multicolumn{2}{c}{\textbf{HP}} & \multicolumn{2}{c}{\textbf{KITTI-2012}} & \multicolumn{2}{c}{\textbf{KITTI-2015}} & 
             \multicolumn{2}{c}{\textbf{Sintel-Cleam}} & 
             \multicolumn{2}{c}{\textbf{Sintel-Final}}\\ 
             & AEPE  $\downarrow$  &    PCK-5  [\%] $\uparrow$   & AEPE  $\downarrow$            & F1   [\%] $\downarrow$        & AEPE  $\downarrow$              & F1  [\%]   $\downarrow$      & AEPE $\downarrow$    & PCK-5  [\%] $\uparrow$ & AEPE $\downarrow$    & PCK-5  [\%] $\uparrow$     \\ \midrule
GLU-Net (\textit{Static})  &  \textbf{25.05}   &  \textbf{78.54}  & 3.34    &      \textbf{18.93}     &      9.79      &   37.52  & 6.03 & 84.21 & 7.01 & 81.92   \\   
GLU-Net (\textit{Dynamic})   & 27.01  & 78.37 & \textbf{3.14} & 19.76 & \textbf{7.49} & \textbf{33.83}  & \textbf{4.25} & \textbf{88.40}  & \textbf{5.50} & \textbf{85.10} \\
\bottomrule
\end{tabular}%
}\vspace{1mm}
\label{tab:dataset}
\end{table} 

\begin{figure}[t]
\centering%
\newcommand{\wid}{0.45\textwidth}
\includegraphics[width=0.99\textwidth]{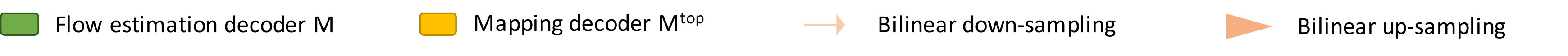}\vspace{-4mm}
\subfloat[BaseNet]{\includegraphics[width=\wid]{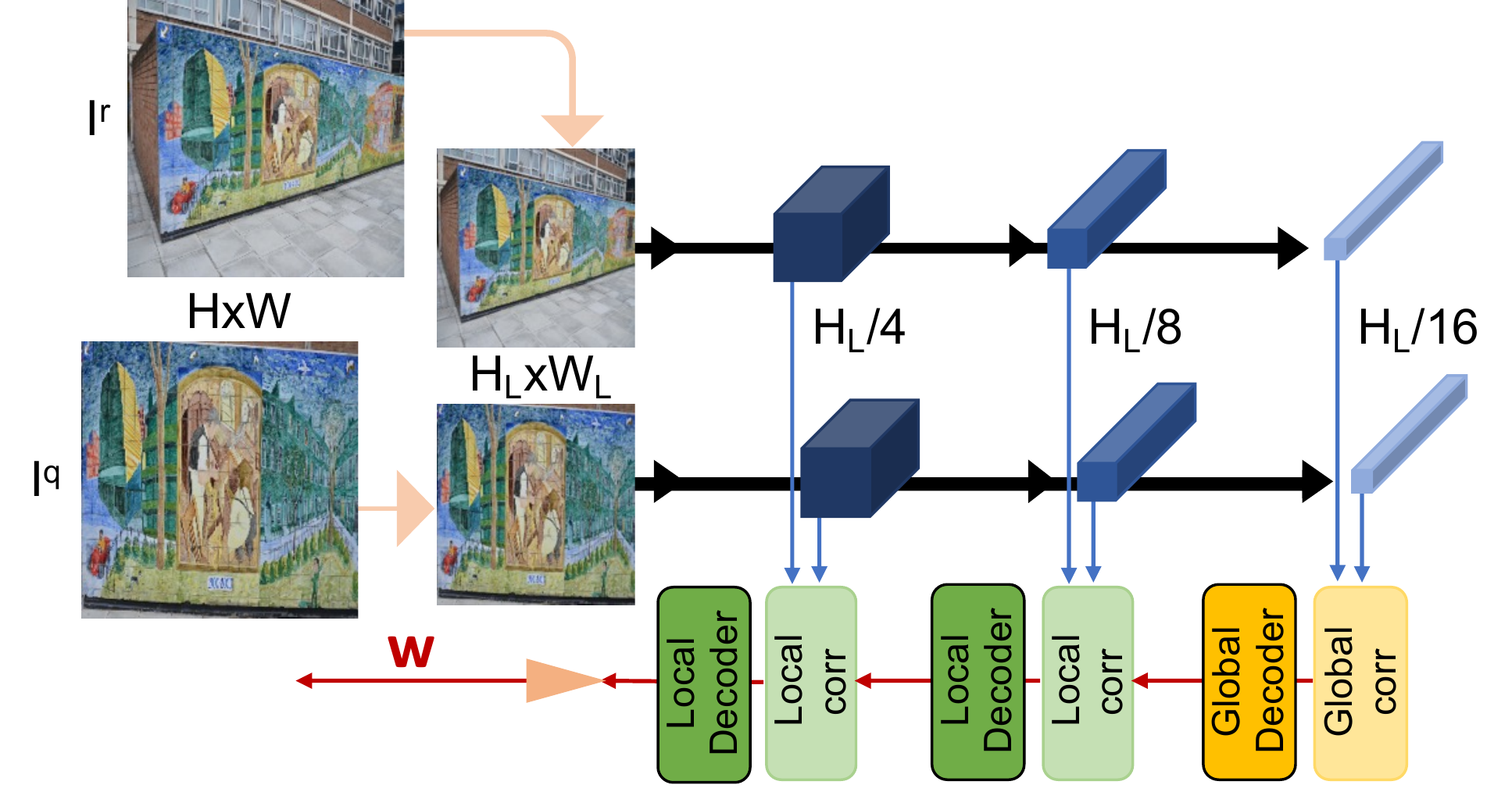}}~%
\subfloat[BaseNet-\dicor]{\includegraphics[width=\wid]{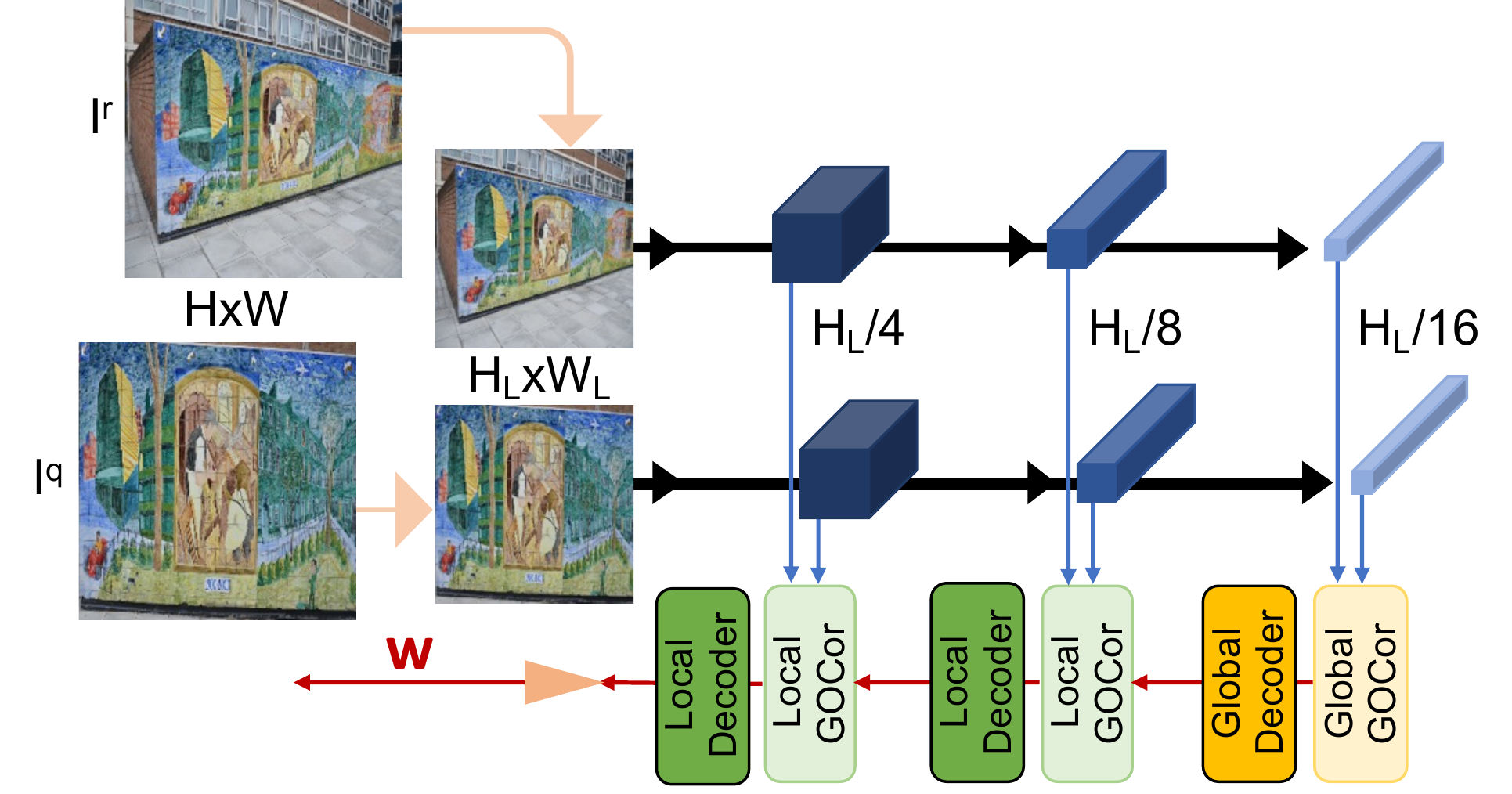}}~%
\vspace{-2mm}
\caption{Schematic representation of BaseNet and BaseNet-\dicor, estimating dense flow field w from a pair of reference and query images. } 
\label{fig:sup-basenet}\vspace{-5mm}
\end{figure}

\subsection{Architecture and training of BaseNet}
\label{arch-basenet}

We introduce BaseNet, a simpler version of GLU-Net \cite{GLUNet}, estimating the dense flow fields relating a pair of images. The network is composed of three pyramid levels and it uses VGG-16~\cite{Chatfield14} as feature extractor backbone. The coarsest level is based on a global correlation layer, followed by a mapping decoder estimating the correspondence map at this resolution. The two next pyramid levels instead rely on local correlation layers. The dense flow field is then estimated with flow decoders, taking as input the correspondence volumes resulting from the local feature correlation layers. 
Besides, BaseNet is restricted to a pre-determined input resolution $H_L \times W_L = 256 \times 256$ due to its global correlation at the coarsest pyramid level. It estimates a final flow-field at a quarter of the input resolution $H_L \times W_L$, which needs to be upsampled to original image resolution $H \times W$. The mapping and flow decoders have the same architecture as those used for GLU-Net \cite{GLUNet}. 

To create BaseNet-\dicor, we simply replace the global and local correlation layers by respectively our global and local \dicor modules. In the standard BaseNet, the correspondence volume generated by the global correlation layer is passed through a ReLU non linearity \cite{relu} and further L2-normalized in the channel dimension, to enhance high correlation values and to down-weight noise values. While beneficial for the standard feature correlation layer, we found the L2-normalization to be slightly harmful for the performance when using our \dicor module. Indeed, our \dicor module inherently already suppresses correlation values at ambiguous matches while enhancing the correct match. We therefore only pass the correspondence volume through a Leaky-ReLU. 
The rest of the architecture remains unchanged. Schematic representations of BaseNet and BaseNet-\dicor are presented in Figure \ref{fig:sup-basenet}.

Both networks are trained end-to-end, following the same procedure introduced in \cite{GLUNet}. We set the batch size to 40 and the initial learning rate of $10^{-3}$, which is further reduced during training.

\section{Additional results}
\label{Sec:sup-results}

Here, we first look at the impact of the number of Steepest Descent iterations used within the filter predictor module during inference in section \ref{iteration}. In section \ref{sup-geo-results}, we then give more detailed results for the task of geometric matching. We subsequently provide an extended table of results on optical flow datasets in section \ref{sup-oF-results} as well as additional results on the ETH3D dataset.  We then illustrate the superiority of our approach through multiple qualitative examples in section \ref{qualitative-results}. Finally, we compare results for different loss parametrization in section \ref{dualbent-resuls}.

Note that for GLU-Net and GLU-Net-\dicor, results on HPatches and TSS are computed with the networks trained on the \textit{Static} dataset while results on MegaDepth, ETH3D and optical flow data KITTI and Sintel use the networks trained on the \textit{Dynamic} data instead. All presented results of GLU-Net-\dicor and PWC-Net-\dicor use 3 global optimization iterations (when applicable) and 7 local optimization iterations. Only on the TSS dataset, we compute metrics for GLU-Net-\dicor with only 3 local optimization iterations. This is because on semantic data, depicting different instances of the same object, the value of the reference objective is less pronounced compared to geometric and optical flow data, for which image pairs represent the same scene. Therefore, on semantic data, using too many local optimization iterations can be harmful.

\begin{figure*}[t]
\centering
\begin{tabular}{c}
(a) HPatches \\
\includegraphics[width=0.24\textwidth]{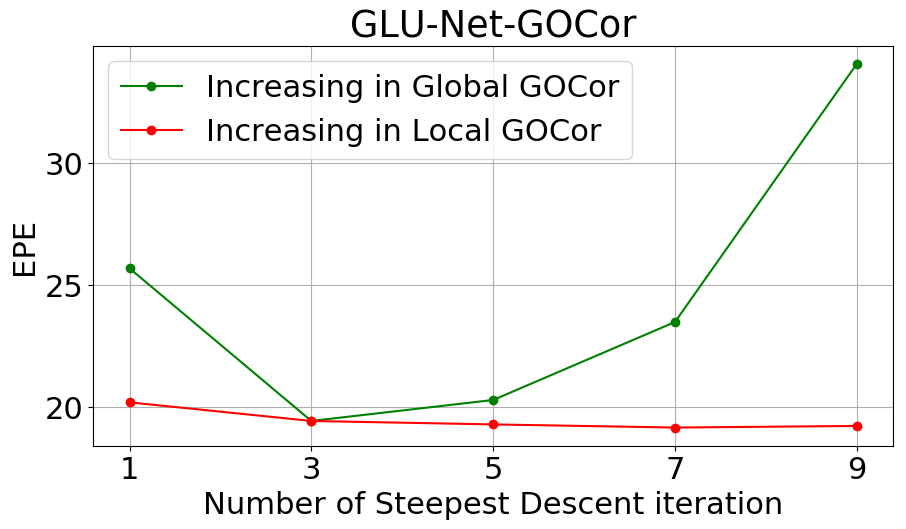} 
\includegraphics[width=0.24\textwidth]{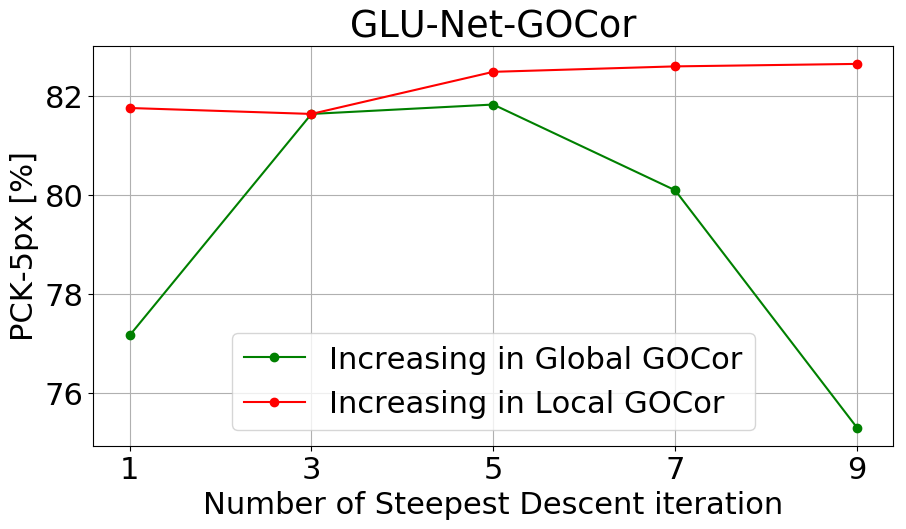} \\ \\

(b) KITTI-2012 \\
\includegraphics[width=0.24\textwidth]{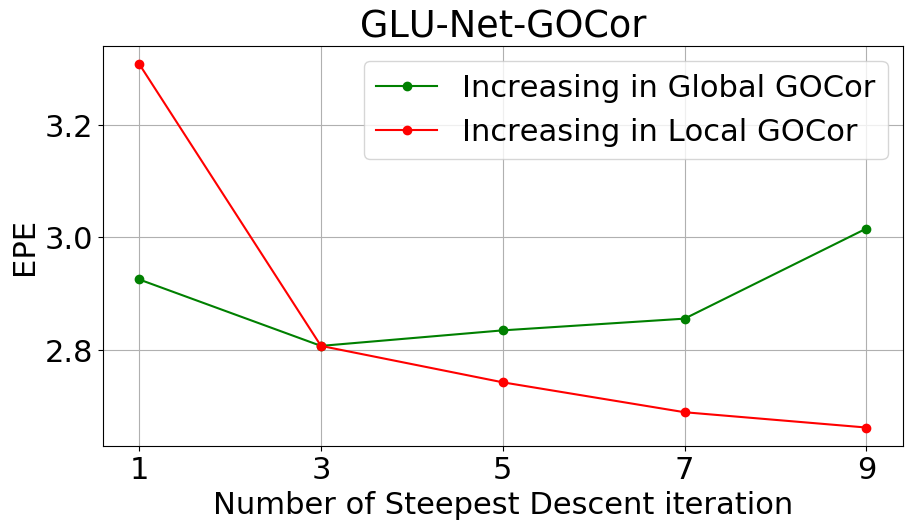}
\includegraphics[width=0.24\textwidth]{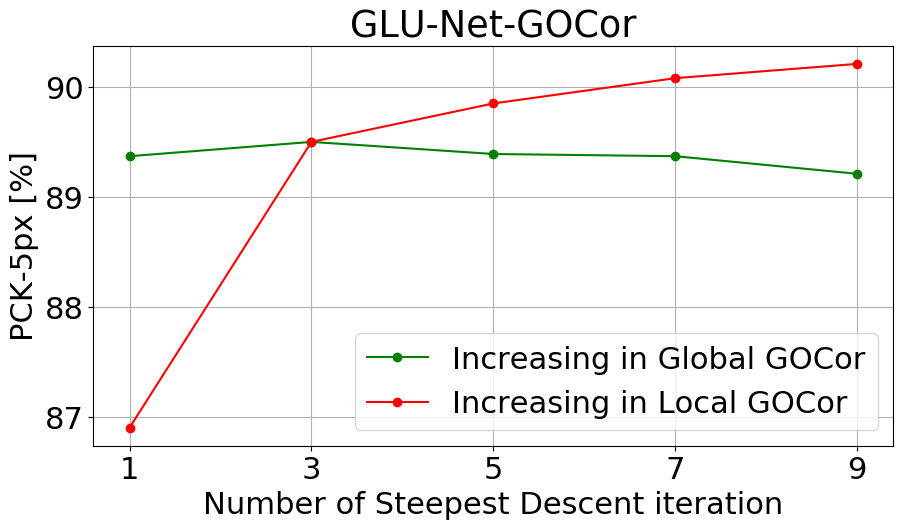} 
\includegraphics[width=0.24\textwidth]{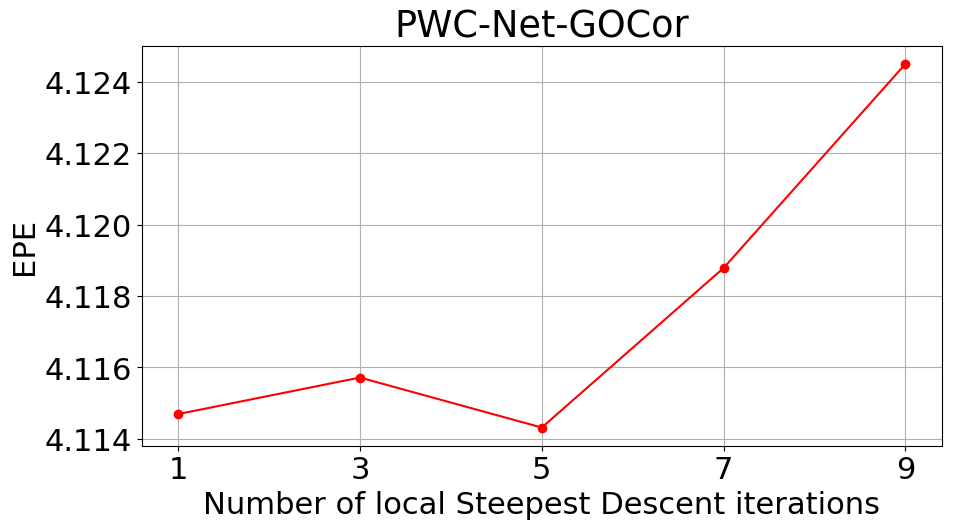}
\includegraphics[width=0.24\textwidth]{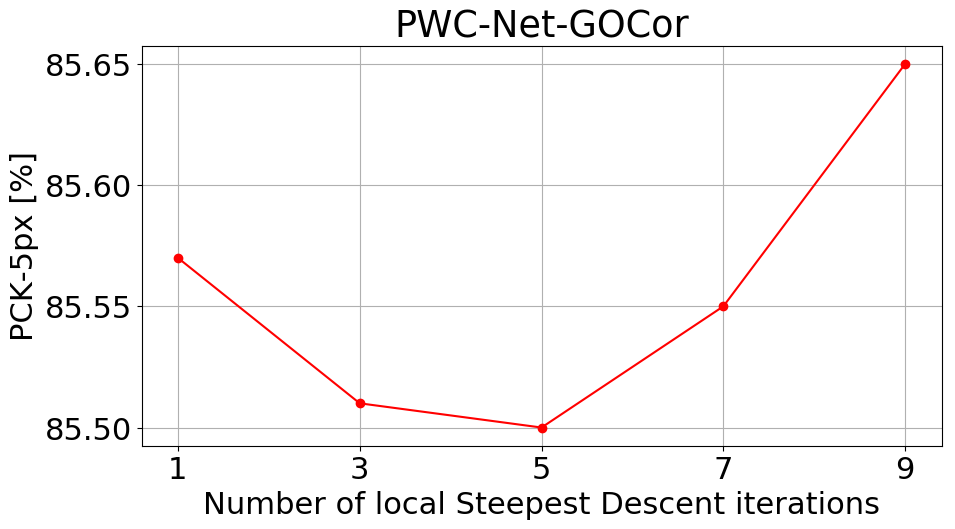}\\ \\

(b) Sintel-clean \\
\includegraphics[width=0.24\textwidth]{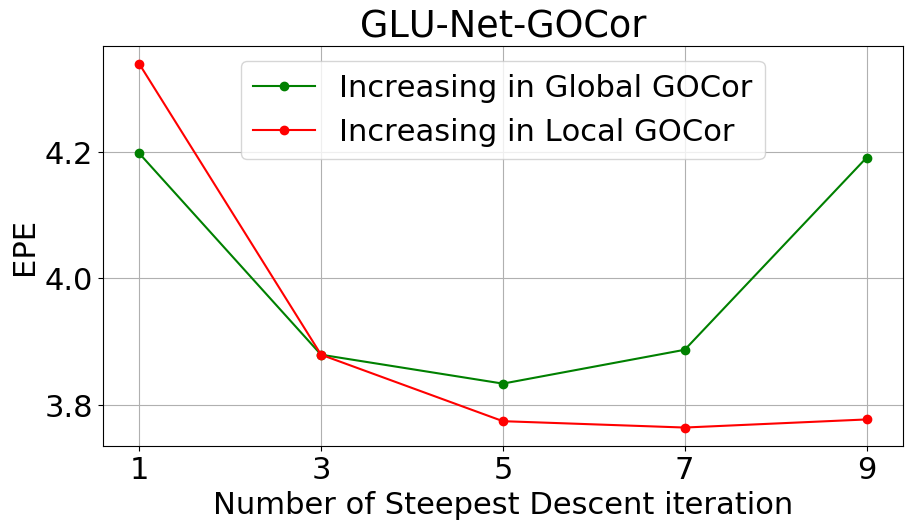}
\includegraphics[width=0.24\textwidth]{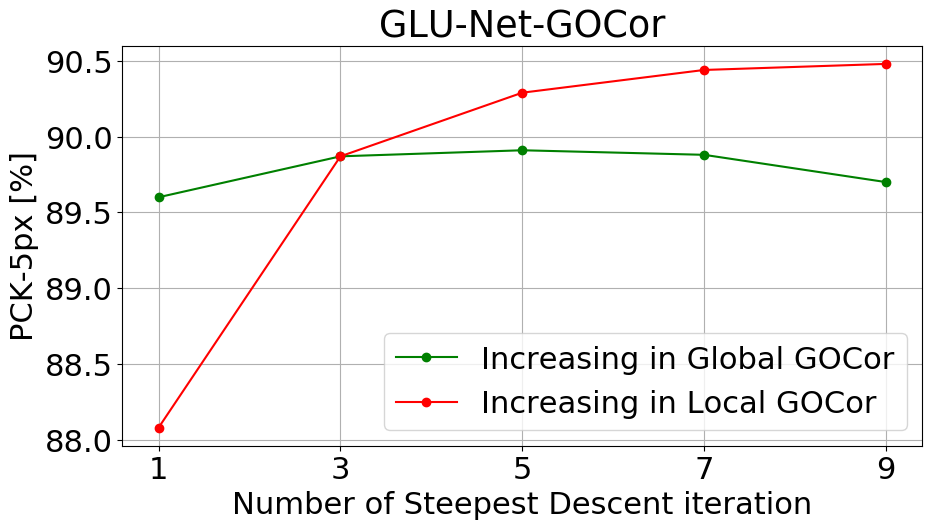}
\includegraphics[width=0.24\textwidth]{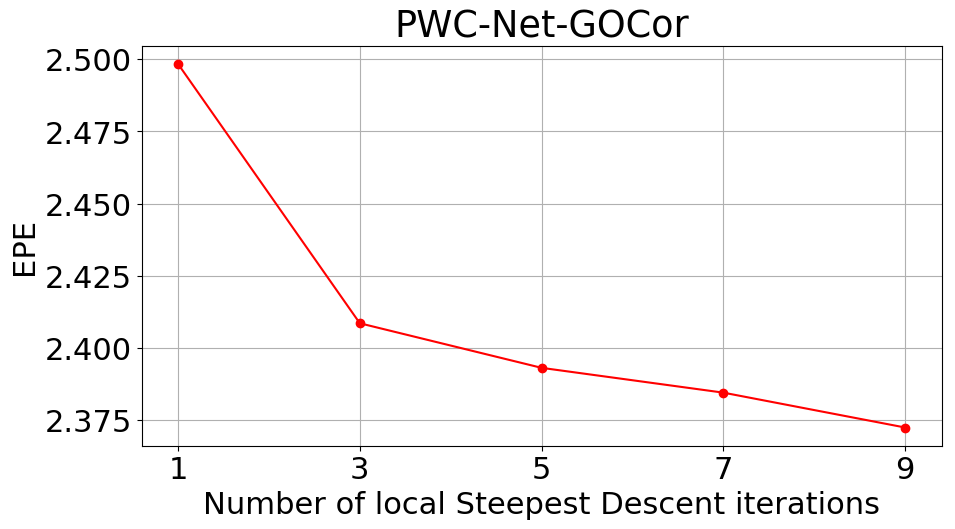}
\includegraphics[width=0.24\textwidth]{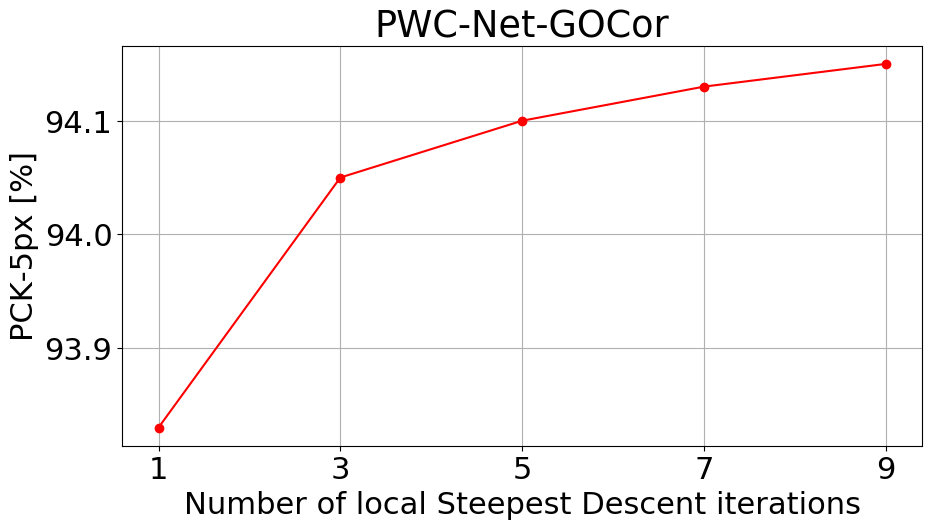}
\end{tabular}
\vspace{-3mm}\caption{Evaluation results of GLU-Net-\dicor and PWC-Net-\dicor when increasing the number of steepest descent iterations in either the global or the local \dicor modules.
While increasing the number of global iterations, the number of local iterations is fixed to three, and similarly. Note that GLU-Net-\dicor was trained on the \textit{Dynamic} dataset and PWC-Net-\dicor was trained on \textit{3D-Things}. Both networks were trained with three steepest descent iterations for both the local and global \dicor modules, if applicable.}
\label{fig:eval-it}
\vspace{-5mm}\end{figure*}

\subsection{Impact of number of inference Steepest Descent iterations}
\label{iteration}

\parsection{Impact on performance} Here, we analyse the impact of the number of Steepest Descent iterations in the global and local \dicor modules used during inference, on the performance of the corresponding network. We first focus on the global \dicor module. 
In Figure \ref{fig:eval-it}, we plot the AEPE and PCK-5px obtained by GLU-Net-\dicor when increasing the number of global steepest descent iterations during inference. Note that the network was trained with three global iterations. On both HPatches and KITTI-2012, GLU-Net-\dicor  with three global iterations, i.e.\ with the same number of iterations than used during training, leads to the best performance. Increasing or decreasing the number of global iterations leads to a significant drop in performance. This is primarily due to our query frame objective (Sec.~\ref{subsec:smooth-loss} of the main paper), which learns the optimal regularizer weights for the number of steepest descent iterations used during training.  

We next look at the impact of the number of steepest descent iterations used in the \emph{local} \dicor module. In Figure \ref{fig:eval-it}, we thus plot the AEPE and PCK-5px of GLU-Net-\dicor and PWC-Net-\dicor for different inference number of local optimization iterations. Both networks were trained with three such iterations. Increasing the number of local steepest descent iterations during inference improves the network performances on all datasets. On KITTI-2012 only, the trend is slightly different, however the difference in performance for different number of iterations is insignificant, in the order of 0.01. 
It is important to note that in the local \dicor module, we only use our robust loss in the reference frame (Sec. \ref{subsec:loss-ref}), ignoring the loss on the query frame (Sec. \ref{subsec:smooth-loss}). Therefore, increasing the number of iterations during inference will in that case make the predicted filter map $\wt_{ij}$ at location $(i,j)$ more and more discriminative to reference feature $\ftr_{ij}$. The final correspondence volume obtained from applying optimized $\wtm$ to the query feature map will thus be more accurate. 

Taking into consideration solely the performance gain, in the global-\dicor, the best alternative during inference is to use the same number of steepest descent iterations than during training (i.e. three iterations here). For the local-\dicor on the other hand, increasing the number of inference steepest descent iterations leads to better resulting network metrics. However, one must take into account that while increasing the number of inference iterations in the local \dicor module leads to improved performances, it also results in increased inference run-time. 

\begin{table}[b]
\centering
\vspace{-3mm}\caption{Run time of our method compared to original versions of PWC-Net and GLU-Net, averaged over the 194 image pairs of KITTI-2012. The number of optimization iterations is indicated for the local-\dicor modules. For GLU-Net-\dicor, we use three steepest descent iterations in the global-\dicor. }
\vspace{-1mm}
\resizebox{0.99\textwidth}{!}{%
\begin{tabular}{@{}llll|lll@{}}
\toprule
                & PWC-Net  & PWC-Net-\dicor & PWC-Net-\dicor  & GLU-Net & GLU-Net-\dicor & GLU-Net-\dicor \\
                &  & optim-iter = 3&  optim-iter = 7 & & optim-iter = 3 & optim-iter = 7 \\

                \midrule
Run-time [ms]  & 118.05 & 166.00 & 203.02  &  154.97 & 211.02 & 261.90 \\ \bottomrule
\end{tabular}%
}\vspace{1mm}
\label{tab:runtime}
\end{table}

\parsection{Impact on run-time} We thus compare the run-time of our \dicor-networks for different number of optimization iterations in the local \dicor module. For reference, we additionally compare them to their corresponding original networks GLU-Net and PWC-Net. The run-times computed on all images of the KITTI-2012 images are presented in Table~\ref{tab:runtime}. The timings have been obtained on the same desktop with an NVIDIA Titan X GPU. All networks output a flow at a quarter resolution of the input images. We up-scale to the image resolution with bilinear interpolation. This up-scaling operation is included in the estimated time. 

Therefore, we found that setting seven steepest descent iterations during inference in the local \dicor was a good compromise between excellent performance and reasonable run-time. All results in the main paper are indicated with this setting. Nevertheless, for time-demanding applications, only using three local optimization iterations (i.e. the same number than during training) results in faster \dicor-networks with still a significant performance gain compared to their original feature correlation layer-based networks.

\begin{figure}[t]
\centering
\includegraphics[width=0.99\textwidth]{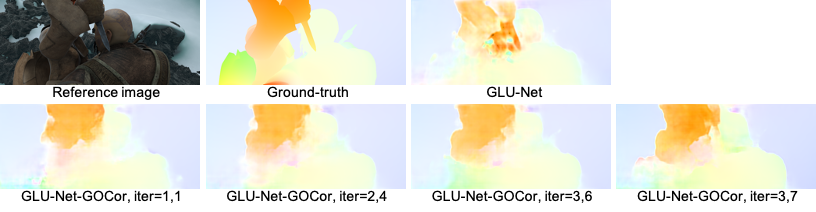}
\vspace{-3mm}\caption{Visualization of the flow field estimated by GLU-Net and GLU-Net-\dicor for different steepest descent iterations during inference. The image pair is extracted from the clean pass of Sintel. The first number indicates the number of optimization iterations in the global \dicor module while the second refer to the number of steepest descent iterations in the local \dicor module. 
Both GLU-Net and GLU-Net-\dicor were trained on the \textit{Dynamic} dataset. GLU-Net-\dicor was trained with three steepest descent iterations for both the local and global \dicor modules. }
\vspace{-4mm}\label{fig:glunet-sintel-it}
\end{figure}

In Figure \ref{fig:glunet-sintel-it}, we visualize the estimated flow field for a pair of Sintel images, by GLU-Net and GLU-Net-\dicor for different optimization iterations. It is very clear that increasing the number of optimization iterations leads to a more accurate estimated flow field. In particular, the estimated flow field becomes more detailed, with well-defined motion boundaries.

\subsection{Additional geometric matching results}
\label{sup-geo-results}

Detailed results obtained by GLU-Net and GLU-Net-\dicor on the various view-points of the HP dataset are presented in Table~\ref{tab:details-HP}. It extends Table \ref{tab:geo-match-HP} of the main paper, that only provides the average over all viewpoint IDs. In addition to the Average End-Point Error (AEPE), we also provide the standard deviation over the End-Point Error per image. It represents the distribution of EPE per image, averaged over all images of each view-point. 
Note that increasing view-point IDs lead to increasing geometric transformations due to larger changes in viewpoint. 

Our approach GLU-Net-\dicor outperforms original GLU-Net for each viewpoint ID. Particularly, GLU-Net-\dicor is significantly more robust for large view-point changes, such as those experienced in Viewpoint V, with an AEPE of 38.41 against 51.47 for original GLU-Net. Besides, GLU-Net-\dicor always obtains a narrower distribution of errors. This implies that our approach enables the network to have a more steady performance over the whole dataset.

\begin{table*}[t]
\centering
\caption{Details of AEPE and PCK evaluated over each view-point ID of the HPatches dataset. All methods are trained on the \textit{Static} dataset. } \vspace{1mm}
\resizebox{\textwidth}{!}{%
\begin{tabular}{ll|cccccc}
\toprule
 & & I  & II  & III  & IV  & V  & all \\ \midrule
        & AEPE $\downarrow$ & 1.55 $\pm$ 1.80 & 12.66 $\pm$ 10.43 & 27.54 $\pm$ 16.05 & 32.04 $\pm$ 20.01 & 51.47 $\pm$ 94.77 & 25.05 $\pm$ 16.67 \\
GLU-Net & PCK-1px [\%]  $\uparrow$ & 61.72 & 42.43 & 40.57 & 29.47 & 23.55 & 39.55 \\
        & PCK-5px [\%] $\uparrow$ & 96.15 & 84.35 & 79.46 & 73.80 & 58.92 & 78.54 \\\midrule
                & AEPE $\downarrow$ & \textbf{1.29 $\pm$ 1.31} &\textbf{ 10.07 $\pm$ 7.44} & \textbf{23.86 $\pm$ 14.01} & \textbf{27.17 $\pm$ 16.84} & \textbf{38.41 $\pm$ 28.52} & \textbf{20.16 $\pm$ 13.63} \\
\textbf{GLU-Net-\dicor}  & PCK-1px [\%] $\uparrow$ & \textbf{64.93} & \textbf{43.86} & \textbf{42.52} & \textbf{30.68} & \textbf{25.78} & \textbf{41.55} \\
                & PCK-5px [\%] $\uparrow$ & \textbf{96.95} & \textbf{86.41} & \textbf{82.47} & \textbf{76.17} & \textbf{65.15} & \textbf{81.43} \\ \bottomrule
 \end{tabular}%
}\vspace{-4mm}
\label{tab:details-HP}
\end{table*}

\subsection{Additional optical flow results}
\label{sup-oF-results}

\parsection{Extended optical flow results} Table \ref{tab:sup-optical-flow} here extends Table~\ref{tab:optical-flow} in the paper with more results on optical flow datasets. Specifically, we compare our approaches with other state-of-the-art networks applied to the train splits of the KITTI and Sintel datasets. Similarly to PWC-Net, these other methods, such as LiteFlowNet \cite{Hui2018}, rely on local correlation layers at multiple levels to infer the final flow field relating a pair of images. Our local \dicor module could therefore easily be integrated into any of these networks in place of the local correlation layers. 

In Table \ref{tab:sup-optical-flow}, PWC-Net* refers to the results presented in the original PWC-Net publication \cite{Sun2018}. In the middle section, we show the evaluation results of PWC-Net*, as well as PWC-Net-\dicor and PWC-Net both further finetuned on \textit{3D-Things} according to the same schedule. 
In the last section of the table, we focus on the PWC-Net variants finetuned on the \textit{Sintel} training dataset. For a fair comparison, we here also provide the official PWC-Net* ft Sintel results, as well as the PWC-Net-\dicor and PWC-Net versions that we finetuned on \textit{Sintel}. 
Our PWC-Net-\dicor outperforms both PWC-Net* and PWC-Net on the KITTI data by a large margin, while obtaining similar results on the training set of Sintel. As already mentioned in Section \ref{subsec:optical-flow} of the main paper, this highlights the generalization capabilities of our \dicor module.

\begin{table}[b]
\centering
\caption{Results for the optical flow task on the training splits of KITTI~\cite{Geiger2013} and Sintel~\cite{Butler2012}. 
A result in parenthesis indicates that the dataset was used for training. PWC-Net* indicates the results stated in the original PWC-Net paper \cite{Sun2018}. For all methods, the training dataset is indicated in parenthesis next to the method. When not indicated, the method was trained on \textit{Flying-Chairs}~\cite{Dosovitskiy2015} followed by \textit{3D-Things}~\cite{Ilg2017a}.}\vspace{-2mm}%
\resizebox{0.99\textwidth}{!}{%
\begin{tabular}{lcc|cc|ccc|ccc}
\toprule
             & \multicolumn{2}{c}{\textbf{KITTI-2012}} & \multicolumn{2}{c}{\textbf{KITTI-2015}} & \multicolumn{3}{c}{\textbf{Sintel Clean}} & \multicolumn{3}{c}{\textbf{Sintel Final}}\\ 
  & AEPE  $\downarrow$            & F1   (\%)   $\downarrow$      & AEPE  $\downarrow$              & F1  (\%)  $\downarrow$ & AEPE  $\downarrow$  & PCK-1  (\%) $\uparrow$ & PCK-5  (\%) $\uparrow$ & AEPE $\downarrow$   & PCK-1  (\%) $\uparrow$ & PCK-5  (\%) $\uparrow$ \\ \midrule

GLU-Net (\textit{Dynamic}) & 3.14 & 19.76 & 7.49 & 33.83 & 4.25 & 62.08 & 88.40 & 5.50 & 57.85 & 85.10 \\
\textbf{GLU-Net-\dicor} (\textit{Dynamic}) (Ours) & \textbf{2.68} & \textbf{15.43} & \textbf{6.68} & \textbf{27.57} & \textbf{3.80} & \textbf{67.12} & \textbf{90.41} & \textbf{4.90} & \textbf{63.38} & \textbf{87.69} \\ \midrule \midrule

FlowNet2.0~\cite{Ilg2017a} & 4.09 & - & 10.06 & 30.37 & 2.02 & - & - & 3.14 & - & - \\      
SpyNet~\cite{Ranjan2017} & 9.12 & - & - & - & 4.12 & - & - & 6.69 & - & - \\

LiteFlowNet~\cite{Hui2018}   &    4.00                 &           -         &       10.39                &        28.50          & 2.48 & - & - & 4.04 & - & - \\
LiteFlowNet2~\cite{Hui2019} & 3.42 & - & 8.97 & 25.88 & 2.24 & - & - & 3.78 & - & - \\ \midrule

PWC-Net* & 4.14 & 21.38 & 10.35 & 33.67 & 2.55 & - & - & 3.93 & - & - \\
PWC-Net (\textit{ft 3D-Things}) & 4.34 & 20.90 & 10.81 & 32.75 & 2.43 & 81.28 & 93.74 & 3.77 & 76.53 & 90.87 \\ 
\textbf{PWC-Net-\dicor} (\textit{ft 3D-Things}) (Ours) & \textbf{4.12} & \textbf{19.31} & \textbf{10.33} & \textbf{30.53} & \textbf{2.38} & \textbf{82.17} & \textbf{94.13} & \textbf{3.70} &  \textbf{77.34} & \textbf{91.20} \\ \midrule

FlowNet2 (\textit{ft Sintel}) & 3.54 & - & 9.94 & 28.02 & \trainres{(1.45)} &  \trainres{-} & \trainres{-} & \trainres{(2.19)} &  \trainres{-} & \trainres{-} \\
PWC-Net* (\textit{ft Sintel}) & 2.94 & 12.70 & 8.15 & 24.35 & \trainres{(1.70)} & \trainres{-} & \trainres{-} & \trainres{(2.21)} & \trainres{-} & \trainres{-} \\

PWC-Net (\textit{ft Sintel}) & 2.87 & 11.97 & 8.68 & 23.82 & \trainres{(1.76)} & \trainres{(87.24)} & \trainres{(95.37)} & \trainres{(2.23)} & \trainres{(83.61)} & \trainres{(93.61)}\\

\textbf{PWC-Net-\dicor} (\textit{ft Sintel}) (Ours) & \textbf{2.60} & \textbf{9.67} & \textbf{7.64} & \textbf{20.93} & \trainres{(1.74)} & \trainres{(87.93)} & \trainres{(95.54)} &  \trainres{(2.28)} &  \trainres{(84.15)} & \trainres{(93.71)} \\
\bottomrule
\end{tabular}%
}
\label{tab:sup-optical-flow}
\end{table}

Here, we also present the evaluation results on the test set of the Sintel dataset in Table~\ref{tab:sintel-test}.  We compare PWC-Net and PWC-Net-\dicor, both finetuned on the training split of \textit{Sintel}. For reference and as previously, we also present the official results from the PWC-Net publication \cite{Sun2018}, denoted as PWC-Net*. 
PWC-Net-\dicor outperforms both PWC-Net and PWC-Net* on the clean pass. On the final pass, PWC-Net-\dicor obtains better performance than PWC-Net, but slightly worse results than PWC-Net*. The authors of PWC-Net employ special data augmentation strategies and training procedures for fine-tuning, which are not shared as PyTorch code by the authors. The finetuning procedure that we employed for our PWC-Net-\dicor and standard PWC-Net is therefore different from the one used for in the official PWC-Net results (PWC-Net*). However, we finetuned both PWC-Net and PWC-Net-\dicor with the same setting and procedure, enabling fair comparison between the two. 
PWC-Net-\dicor performs particularly better in regions with large motions and close to the motion boundaries. This is in line with the behavior of the \dicor module observed previously, according to which the \dicor module particularly improves performance on large displacements. 

\begin{table}[H]
\centering
\caption{Detailed results on the test set of Sintel benchmark for different regions, velocities (s), and distances from motion boundaries (d). All methods are trained on \textit{Flying-Chairs}~\cite{Dosovitskiy2015} followed by \textit{3D-Things}~\cite{Ilg2017a}, and further finetuned on the training split of \textit{Sintel}.}
\resizebox{0.99\textwidth}{!}{%
\begin{tabular}{lccccccccc}
\toprule
& \multicolumn{9}{c}{\textbf{Sintel-Clean}} \\
  & EPE-all & EPE matched & EPE unmatched & d0-10 & d10-60 & d60-140 & s0-10 & s10-40 & s40+ \\ \midrule
PWC-Net* (\textit{ft-Sintel}) & 4.386 & 1.719 & 26.166 & 4.282 & 1.657 & \textbf{0.657} & \textbf{0.606} & 0.2070 & 28.783 \\ \midrule
PWC-Net  (\textit{ft-Sintel}) & 4.637 & 1.951 & 26.571 & 4.018 & 1.626 & 1.040 & 0.649 & 2.070 & 30.671 \\
\textbf{PWC-Net-\dicor} (\textit{ft-Sintel})  & \textbf{4.195} & \textbf{1.660} & \textbf{24.909} & \textbf{3.843} & \textbf{1.448} & \textbf{0.778} & \textbf{0.609} & \textbf{1.914} & \textbf{27.552} \\
\midrule \midrule
    & \multicolumn{9}{c}{\textbf{Sintel-Final}} \\
  & EPE-all & EPE matched & EPE unmatched & d0-10 & d10-60 & d60-140 & s0-10 & s10-40 & s40+ \\ \midrule
PWC-Net* (\textit{ft-Sintel})  & \textbf{5.042} & \textbf{2.445} & \textbf{26.221} & 4.636 & 2.087 & \textbf{1.475} & \textbf{0.799} & 2.986 & \textbf{31.070} \\  \midrule
PWC-Net  (\textit{ft-Sintel}) & 5.300 & 2.576 & 27.528 & 4.717 & 2.204 & \textbf{1.580} & 0.929 & 2.994 & 32.584 \\
\textbf{PWC-Net-\dicor} (\textit{ft-Sintel})  &  \textbf{5.133} & \textbf{2.458} & \textbf{26.945} & \textbf{4.504} & \textbf{2.063} & 1.603 & \textbf{0.834} & \textbf{2.906} & \textbf{31.858} \\ 
\bottomrule
\end{tabular}%
}
\label{tab:sintel-test}
\end{table}

\begin{wraptable}{r}{6.0cm}
\vspace{-6mm}
\caption{AEPE/F1 [\%] on KITTI-2015.}
\centering
\resizebox{\linewidth}{!}{%
\begin{tabular}{@{}l@{~~}c@{~~}c@{~~}c@{}}
\toprule
                    & \textbf{Not occluded}          & \textbf{Occluded} &   \textbf{All}           \\ \midrule
GLU-Net       & 4.67 / 27.83 &  21.95 / 67.44   & 7.49 / 33.83  \\
\textbf{GLU-Net-GOCor}  & \textbf{4.22 / 22.03} &  \textbf{19.07 / 58.61}   & \textbf{6.68 / 27.57}  \\
PWC-Net      & 5.40 / 25.16 &   34.39 / 78.58  & 10.81 / 32.75 \\
\textbf{PWC-Net-GOCor} &  \textbf{5.02 / 23.53} &   \textbf{34.06 / 77.84}  & \textbf{10.33 / 30.53} \\ 
\bottomrule
\end{tabular}%
}
\label{tab:occ}
\vspace{-3mm}
\end{wraptable}

\parsection{Performance on occlusion data}
Here, we focus specifically on the performance of GOCor in occluded regions. As shown in Tab.~\ref{tab:sintel-test}, in occluded regions (``EPE unmatched'') of the Sintel test set, GOCor provides relative improvements of 6.25\% and 2.16\% on the clean and final pass respectively. In Tab.~\ref{tab:occ}, we present the details of the metrics on occluded and non-occluded regions of KITTI-2015. GOCor improves the performance of PWC-Net and GLU-Net in occluded regions of a substantial amount as compared to the feature correlation layer. 

\begin{figure*}[b]
\centering
\begin{tabular}{c}
\includegraphics[width=0.31\textwidth]{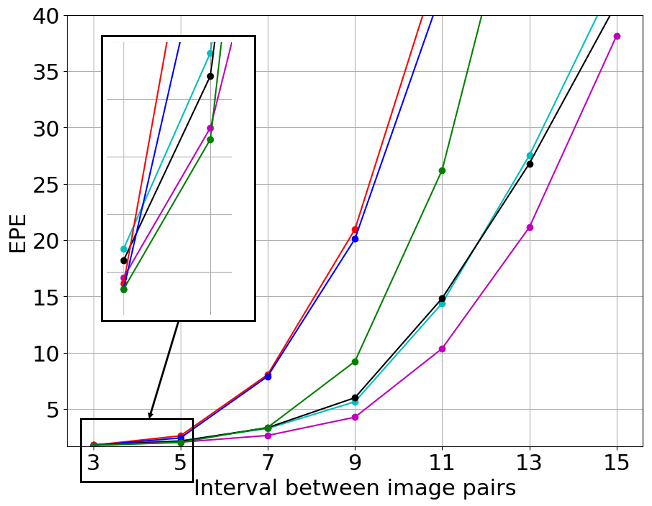} \hspace{0.2cm}
\includegraphics[width=0.31\textwidth]{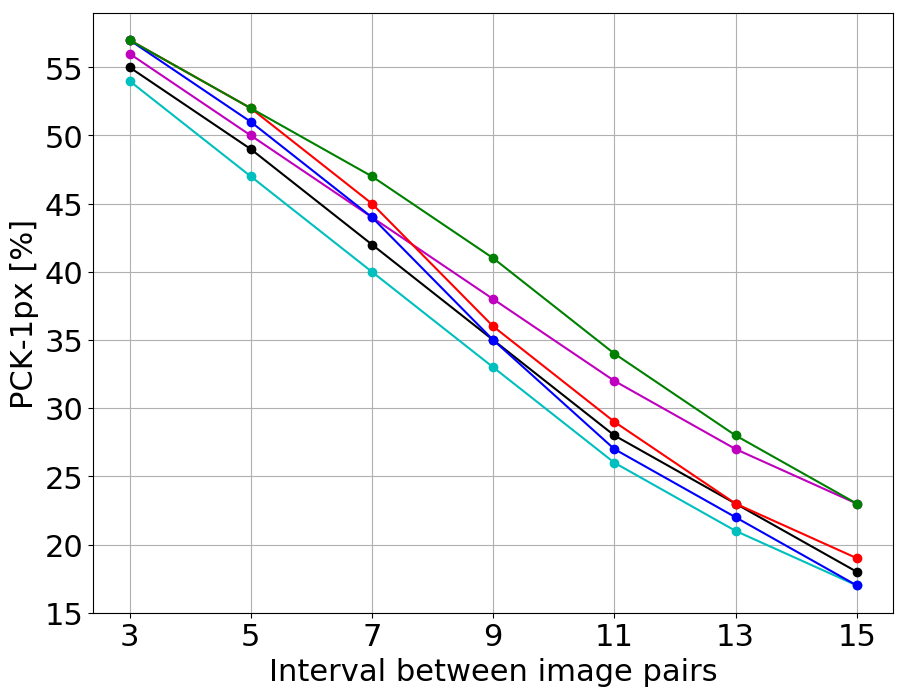}
\includegraphics[width=0.31\textwidth]{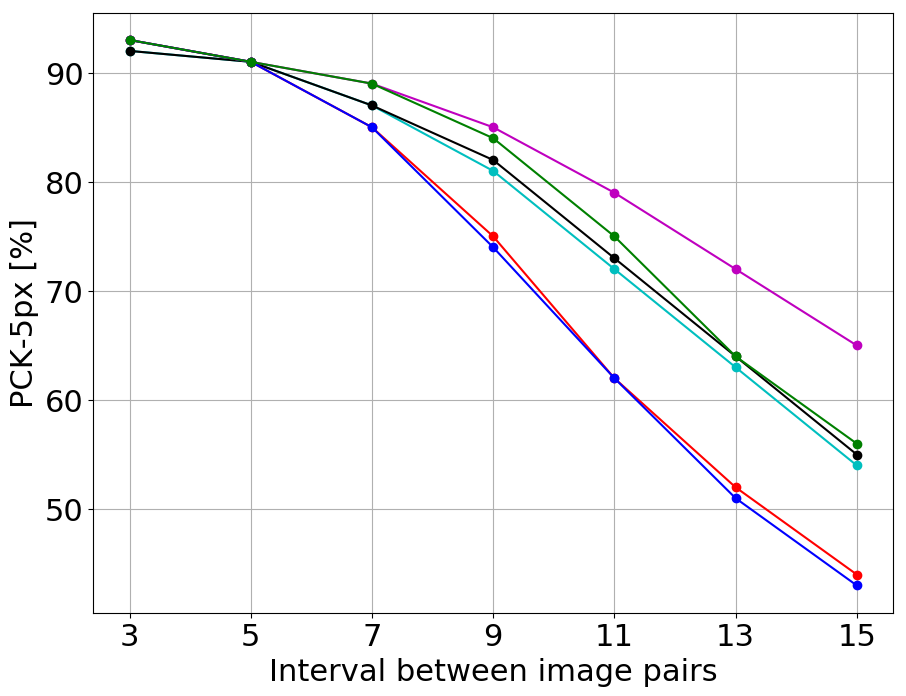}
\end{tabular}
\includegraphics[width=0.80\textwidth,trim=0 0 0 15]{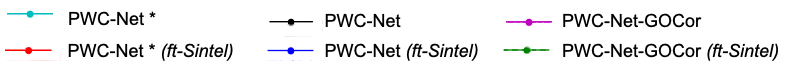}\vspace{-3mm}
\caption{Quantitative results on ETH3D~\cite{ETH3d} images. AEPE, PCK-1 and PCK-5 are computed on pairs of images sampled from consecutive images of ETH3D at different intervals. PWC-Net* refers to the official pre-trained weights provided by the authors.}
\label{fig:sup-ETH3d}
\vspace{-4mm}
\end{figure*}

\subsection{Additional results on ETH3D}

In the main paper, Figure~\ref{fig:ETH3d}, we quantitatively evaluated our approach over pairs of ETH3D images sampled from consecutive frames at different intervals. 
For completeness, in Figure~\ref{fig:sup-ETH3d}, we also present the evaluation results of PWC-Net and PWC-Net-\dicor applied to the ETH3D images, sampled at increasingly high intervals. Indeed, for small intervals, finding correspondences strongly resembles optical flow task while increasing the interval leads to larger displacements. Optical flow network PWC-Net can thus be very suitable, particularly when estimating the flow at small intervals. 
In Figure~\ref{fig:sup-ETH3d}, both PWC-Net and PWC-Net-\dicor are finetuned on \textit{3D-Things} according to the same schedule. For reference, we additionally provide the results of PWC-Net*, which refers to the official pre-trained weights provided by the authors, after training on \textit{Flying-chairs} and \textit{3D-Things}. We also provide the results of PWC-Net*, PWC-Net and PWC-Net-\dicor further finetuned on \textit{Sintel}. 

For all intervals, and independently of the dataset it was trained on, our approach PWC-Net-\dicor obtains better metrics than both PWC-Net and PWC-Net*. Particularly, the gap in performance broadens with the intervals between the frames, implying that our module is especially better at handling large view-point changes. This is especially obvious for the networks finetuned on Sintel. Indeed, as further evidenced in Tab.~\ref{tab:ETH3d-details}, PWC-Net and PWC-Net* finetuned on Sintel experience a very large drop in performance on all metrics from a frame interval of 7, as compared to PWC-Net-\dicor finetuned on the same data.
This particular robustness to large displacements was similarly observed when our \dicor modules were integrated into GLU-Net (Sec. \ref{subsec:geom-exp}). 
It highlights that the improved performances brought by our \dicor approach compared to the feature correlation layer \emph{generalize to different networks}.  

In Table~\ref{tab:ETH3d-details}, we give the corresponding detailed evaluation metrics (AEPE and PCK) obtained by geometric matching networks DGC-Net, GLU-Net, GLU-Net-\dicor and optical flow models LiteFlowNet, PWC-Net,  PWC-Net* and PWC-Net-\dicor. 

\begin{table}[t]
\centering
\caption{Metrics evaluated over scenes of ETH3D with different intervals between consecutive pairs of images (taken by the same camera). Note that those results are the average over the different sequences of the ETH3D dataset. Small AEPE and high PCK are better.}\label{tab:ETH3d-details}
(a) Geometric matching methods \\
\resizebox{0.99\textwidth}{!}{%
\begin{tabular}{llcccc}
\toprule
&&   DGC-Net & GLU-Net & GLU-Net  & \textbf{GLU-Net-\dicor} \\ 
&& \cite{Melekhov2019} & \textit{Static} = \textit{DPED-CityScape-ADE} in \cite{GLUNet} & \textit{Dynamic}  & \textit{Dynamic} \\\midrule
 
& AEPE  & 2.49 & 1.98 & 2.01 & \textbf{1.93} \\
\textbf{interval = 3} &PCK-1px [\%] & 34.19 &  \textbf{50.55} & 46.27 & 47.97 \\
&PCK-5px [\%] & 88.50 & 91.22 & 91.45 & \textbf{92.08} \\ \midrule

&AEPE  & 3.28 & 2.54 &2.46 &  \textbf{2.28} \\
\textbf{interval = 5} &PCK-1px [\%]  & 27.22 &  \textbf{43.08} & 39.28 & 41.79 \\
&PCK-5px [\%] & 83.25  & 87.91 & 88.57 & \textbf{89.87} \\ \midrule

&AEPE  & 4.18 &  3.48 & 2.98 & \textbf{2.64} \\
\textbf{interval = 7} &PCK-1px [\%] & 22.45 &  \textbf{36.98} & 34.05 & 36.81 \\
&PCK-5px [\%] & 78.32  &  84.23 & 85.64 & \textbf{87.77} \\ \midrule

&AEPE  & 5.35 & 4.23 & 3.51 & \textbf{3.01}\\
\textbf{interval = 9} &PCK-1px [\%]  & 18.82 &  32.45 & 30.11 &  \textbf{33.03} \\
&PCK-5px [\%] &  73.74 &   80.74 & 83.10 & \textbf{85.88} \\ \midrule

&AEPE  & 6.78 & 5.59 & 4.30 & \textbf{3.62} \\
\textbf{interval = 11} &PCK-1px [\%]  & 15.82 & 28.45 & 26.69 & \textbf{29.80} \\
&PCK-5px [\%]& 69.23  &  \textbf{76.84} & 80.12 & \textbf{83.69} \\ \midrule

&AEPE &         9.02 & 7.54 & 6.11 & \textbf{4.79} \\
\textbf{interval = 13} &PCK-1px [\%]  & 13.49 &  25.06 & 23.73 & \textbf{26.93} \\
&PCK-5px [\%] & 64.28  & 72.35 & 76.66 & \textbf{81.12} \\ \midrule

&AEPE &         12.25 &  10.75 & 9.08 & \textbf{7.80} \\
\textbf{interval = 15} &PCK-1px [\%]  & 11.25 &  21.89 & 20.85 & \textbf{23.99} \\
&PCK-5px [\%]  & 58.66  & 67.77 &73.02 &  \textbf{77.90} \\ \midrule
\end{tabular}%
}
\\
(b) Optical flow methods \\
\resizebox{0.99\textwidth}{!}{%
\begin{tabular}{llcccc|cc}
\toprule
&&  LiteFlowNet & PWC-Net* & PWC-Net & \textbf{PWC-Net-\dicor} & PWC-Net & \textbf{PWC-Net-\dicor}  \\ 
&&  \textit{Chairs-Things} & \textit{Chairs-Things} & \textit{Chairs-Things} & \textit{Chairs-Things} & \textit{ft Sintel} & \textit{ft Sintel}  \\\midrule
 
& AEPE &  \textbf{1.67} & 1.75 & 1.73 & 1.70 & 1.68 & \textbf{1.67} \\
\textbf{interval = 3} &PCK-1px [\%] & \textbf{61.63} & 57.54 & 58.50 & 58.93 & 59.93 &  \textbf{60.46} \\
&PCK-5px [\%] & 92.79 & 92.62 & 92.64 & \textbf{92.81} & 92.89 & \textbf{92.92}\\ \midrule

&AEPE &  2.58 & 2.10 & 2.06 & \textbf{1.98} &  2.35 & \textbf{1.95}\\
\textbf{interval = 5} &PCK-1px [\%] & \textbf{56.55} & 50.41 & 52.02 &  53.10 & 54.41 & \textbf{55.54} \\
&PCK-5px [\%] & 90.70 & 90.71 & 90.82 & \textbf{91.45} & 91.10 & \textbf{91.59}\\ \midrule

&AEPE &  6.05 & 3.21 & 3.28 &  \textbf{2.58} & 7.83 & \textbf{3.29} \\
\textbf{interval = 7} &PCK-1px [\%] & \textbf{49.83} & 42.95&  44.86 & 46.91 &  46.76 & \textbf{50.37}  \\
&PCK-5px [\%] & 86.29 & 87.04 & 87.32 & \textbf{88.96} & 85.21 & \textbf{89.51}  \\ \midrule

&AEPE &         12.95 & 5.59 & 5.95 & \textbf{4.22} &  20.05 & \textbf{9.16} \\
\textbf{interval = 9} &PCK-1px [\%] & \textbf{42.00} & 35.23 & 37.41 & 40.93 &  37.24 & \textbf{44.33} \\
&PCK-5px [\%] & 78.50 & 81.17 & 81.80 & \textbf{85.53} & 73.90 & \textbf{84.65}  \\ \midrule

&AEPE &         29.67 & 14.35 & 14.79 & \textbf{10.32} & 42.17 & \textbf{26.16} \\
\textbf{interval = 11} &PCK-1px [\%] & 33.14 & 28.14  & 30.36 & \textbf{34.58} & 29.24 & \textbf{36.91} \\
&PCK-5px [\%] & 66.07 & 71.91 & 72.95 & \textbf{79.44}  & 61.64 & \textbf{75.49}\\ \midrule

&AEPE &         52.41 & 27.49 & 26.76 & \textbf{21.10} &  67.79 & \textbf{56.47}\\
\textbf{interval = 13} &PCK-1px [\%] & 26.46 & 22.91 & 24.75 & \textbf{29.25} &  23.40 & \textbf{30.27} \\
&PCK-5px [\%] & 55.05 & 63.19 & 64.07 & \textbf{72.06} & 51.31 & \textbf{63.79} \\ \midrule

&AEPE &         74.96 & 43.41 & 40.99 & \textbf{38.12} & 94.59 & \textbf{83.18} \\
\textbf{interval = 15} &PCK-1px [\%] & 21.22 & 18.34 & 19.89 & \textbf{24.59} &  18.72 &  \textbf{25.17} \\
&PCK-5px [\%] & 46.29 & 54.39 & 55.47 & \textbf{64.92} & 43.22 &  \textbf{55.84} \\ \midrule
\end{tabular}%
}\vspace{1mm}

\end{table}

\subsection{Qualitative examples}
\label{qualitative-results}

\begin{figure*}[t]
\centering
\includegraphics[width=0.95\textwidth]{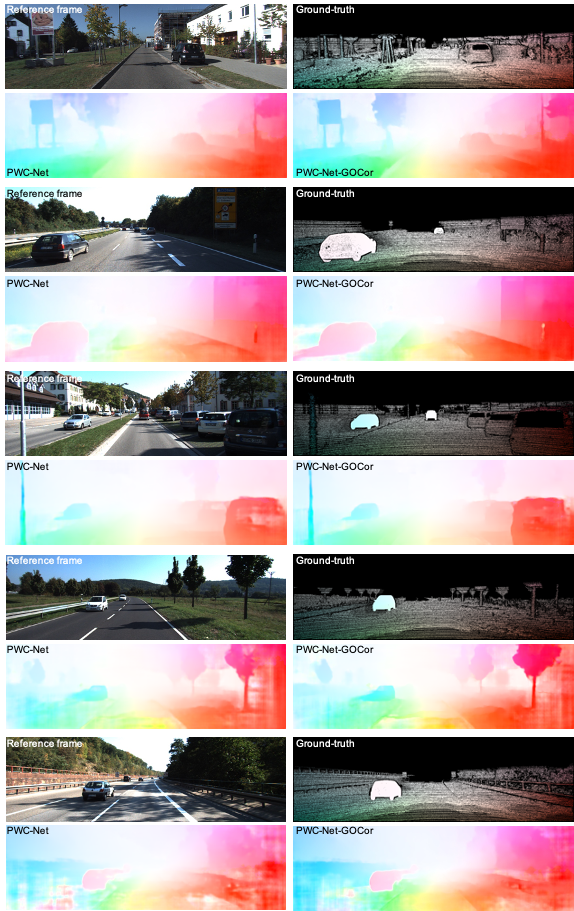}
\caption{Qualitative examples of PWC-Net and PWC-Net-\dicor applied to images of KITTI-2012 and KITTI-2015. Both models are trained on \textit{Flying-Chairs} followed by \textit{3D-Things}.}
\label{fig:kitti-pwcnet}
\end{figure*}

\begin{figure*}[t]
\centering
\includegraphics[width=0.90\textwidth]{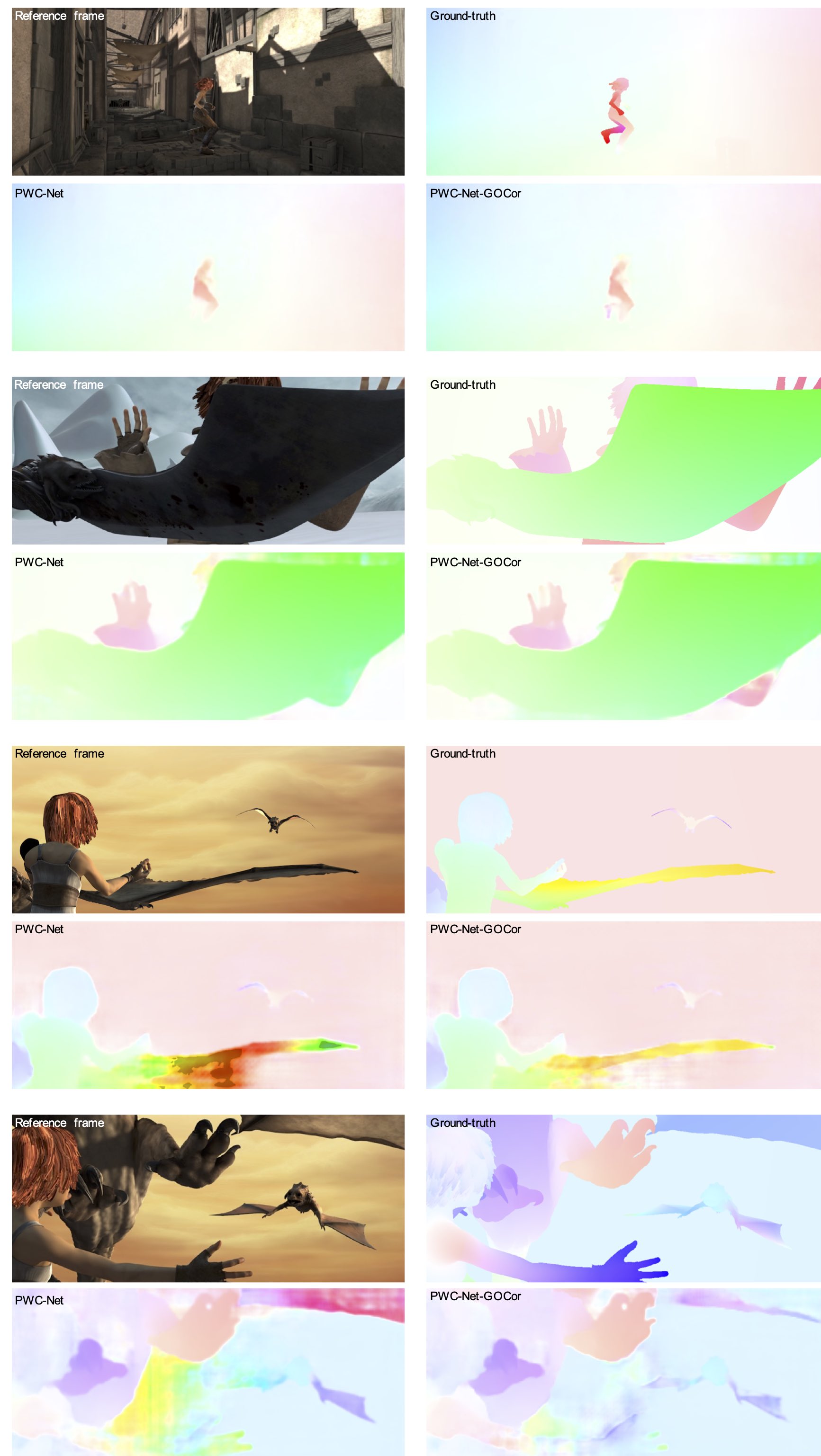}
\caption{Qualitative examples of PWC-Net and PWC-Net-\dicor applied to images of Sintel-clean. Both models are trained on \textit{Flying-Chairs} followed by \textit{3D-Things}.}
\label{fig:sintel-pwcnet}
\end{figure*}

Here, we first present qualitative comparisons of PWC-Net and PWC-Net-\dicor. In Figure \ref{fig:kitti-pwcnet}, we show examples of PWC-Net and PWC-Net-\dicor applied to images of optical flow datasets KITTI-2012 and KITTI-2015. Both networks are trained on \textit{3D-Things}. PWC-Net-\dicor shows more defined motion boundaries and generally more accurate estimated flow fields.  
Similarly, we present examples on the clean pass of the Sintel training set in Figure \ref{fig:sintel-pwcnet}. PWC-Net-\dicor captures more detailed flow fields. This is for instance illustrated in the first example of Figure \ref{fig:sintel-pwcnet}, where PWC-Net-\dicor correctly identified the foot contrary to PWC-Net, which failed to capture it. However, both PWC-Net and PWC-Net-\dicor may fail on small and rapidly moving objects, such as the arm in the last example of Figure \ref{fig:sintel-pwcnet}. 

\begin{figure}
\centering
\includegraphics[width=0.90\textwidth]{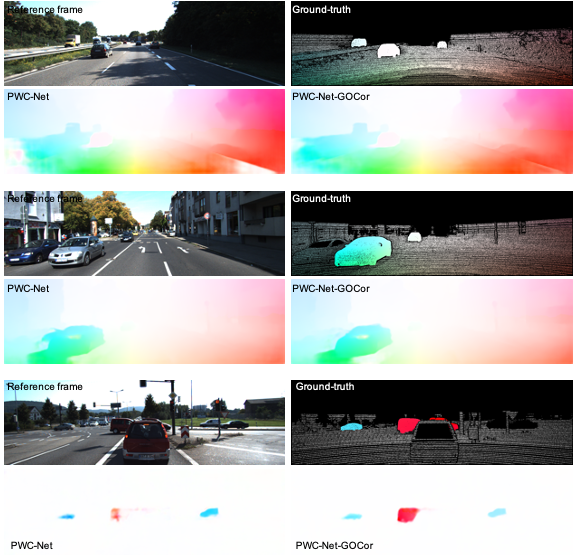}
\caption{Qualitative examples of PWC-Net and PWC-Net-\dicor applied to images of KITTI2012 and 2015. Both models are finetuned on \textit{Sintel}.}
\label{fig:kitti-pwcnet-ft-sintel}
\end{figure}

In Figure \ref{fig:kitti-pwcnet-ft-sintel}, we additionally present qualitative results on the KITTI images when the networks are finetuned on Sintel. Indeed, in Table \ref{tab:sup-optical-flow}, we showed that while both PWC-Net and PWC-Net-\dicor obtain very similar results on training data \textit{Sintel}, PWC-Net-\dicor performs largely better on the KITTI datasets compared to original PWC-Net, especially in terms of F1 metric. PWC-Net-\dicor also obtains visually more accurate flow fields on the KITTI images.

\begin{figure*}[t]
\centering
(a) ETH3D images \\
\includegraphics[width=0.99\textwidth]{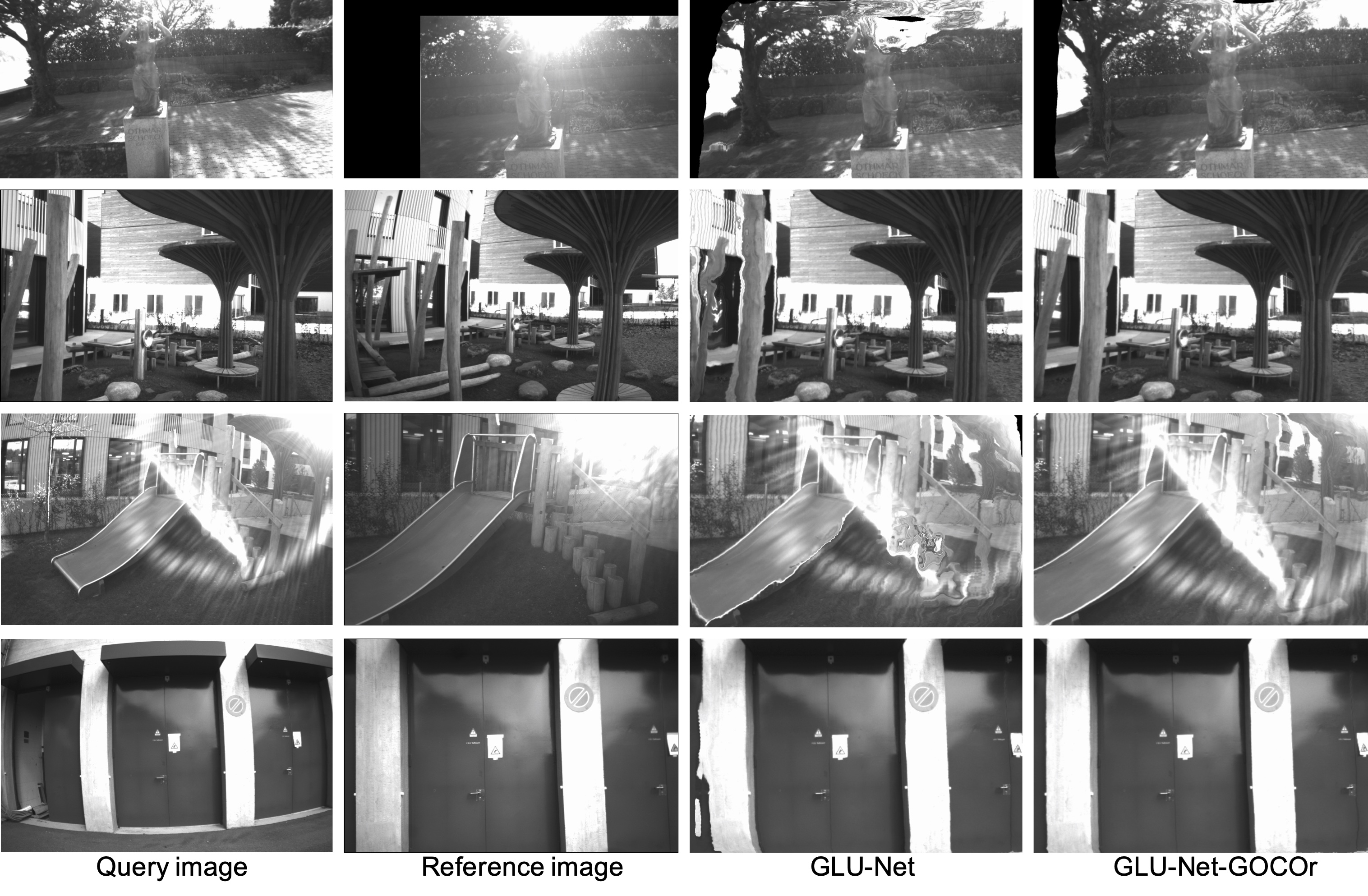} \\
(b) Sintel-clean images \\
\includegraphics[width=0.90\textwidth]{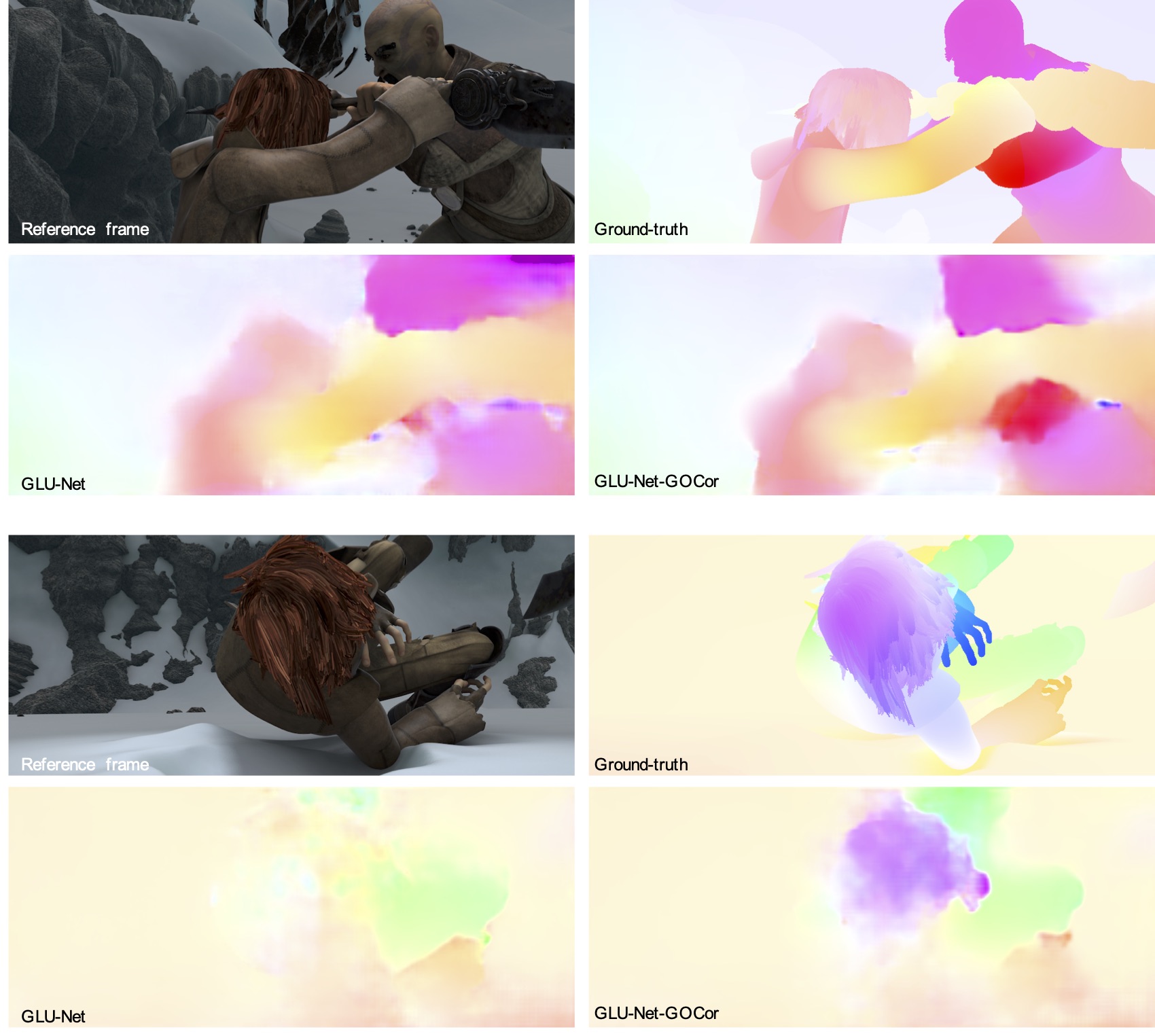}
\caption{Qualitative examples of GLU-Net and GLU-Net-\dicor applied to images of (a) the ETH3D dataset and (b) the Sintel dataset, Clean pass. Both models are trained on the \textit{Dynamic} dataset. In the case of the ETH3D images, we visualize the query images warped according to the flow fields estimated by the network. The warped query images should resemble the reference images. In the case of Sintel images, we plot directly the estimated flow field for each image pair.}
\label{fig:sintel-glunet}
\end{figure*}

Besides, we show the advantage of our approach as compared to the feature correlation layer when integrated in GLU-Net. In Figure \ref{fig:sintel-glunet}, we visually compare GLU-Net and GLU-Net-\dicor when applied to images of the clean pass of the Sintel benchmark and to images of the ETH3D dataset. In the case of the ETH3D images, the pairs of images are taken by two different cameras simultaneously. The camera of the first images has a field-of-view of 54 degrees while the other camera has a field of view of 83 degrees. They capture images at a resolution of $480 \times 752$ or $514 \times 955$ depending on the scenes and on the camera. The exposure settings of the cameras are set to automatic, allowing the device to adapt to illumination changes.
On the Sintel images, GLU-Net-\dicor achieves sharper object boundaries and generally more correct estimated flow fields compared to original GLU-Net. On the ETH3D images, our approach is more robust to illumination changes and light artifacts, leading to better visual outputs. 

\begin{figure*}[t]
\centering
\includegraphics[width=0.93\textwidth]{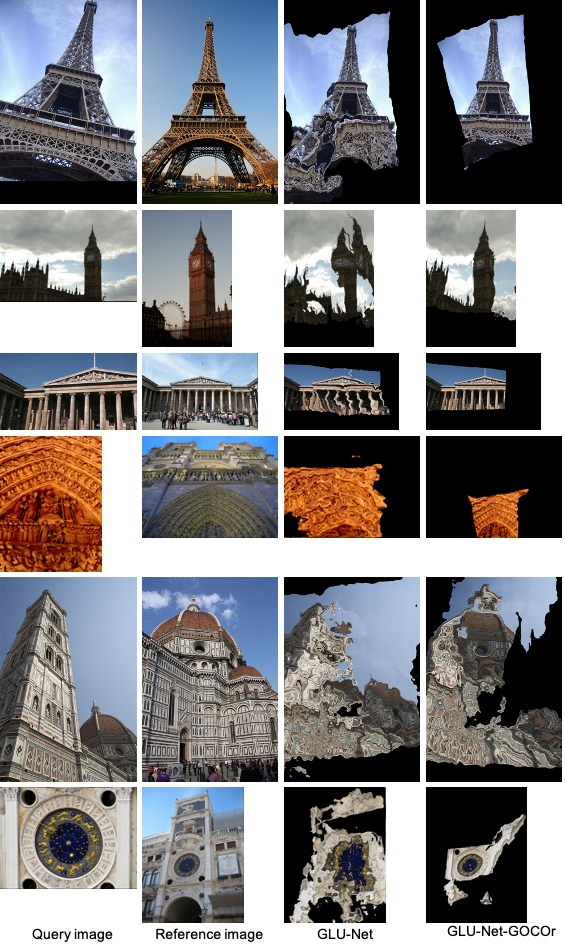}
\vspace{-5mm}\caption{Qualitative examples of GLU-Net and GLU-Net-\dicor applied to images of the MegaDepth dataset. Both models are trained on the \textit{Dynamic} training data.}
\label{fig:mega}
\end{figure*}

Finally, we qualitatively compare the output of GLU-Net-\dicor and GLU-Net on example pairs of the MegaDepth dataset in Figure \ref{fig:mega}. It is obvious that GOCor provides an increased robustness to very large geometric view-point changes, such as large scaling or perspective variations.

\subsection{Results for smooth version of the reference loss}
\label{dualbent-resuls}

In Section \ref{subsec:loss-ref}, we defined our robust and learnable objective function for integrating reference frame information as $\Lr(\wt; \ftr, \theta) = \left\| \sigma_\eta\big(\corr(\wt, \ftr) ; \, \ps, \ns\big) - \y \right\|^2$ (eq. \ref{eq:error-func} of main paper), with $\sigma_\eta(\corr(\wt, \ftr); \ps\!, \ns) = \frac{\ps\! - \ns\!}{2}\! \left(\!\sqrt{\corr(\wt, \ftr)^2 + \eta^2} - \eta\right) + \frac{\ps\! + \ns\!}{2} \corr(\wt, \ftr)$ (eq. \ref{eq:error-func} of main paper).

Here, setting $\eta >0$ enables to avoid the discontinuity in the derivative of $\sigma$ at $\varepsilon=0$. 
We analyze two different settings for $\eta$ when integrated in GLU-Net-\dicor. Specifically, in Table \ref{tab:optical-flow-dualbent}, we compare the results obtained by GLU-Net-\dicor on optical flow data, when setting $\eta=0$ or $\eta=0.1$. Both values obtain very similar results. Therefore, for simplicity and efficiency, we use $\eta = 0$ in all other experiments.

\begin{table}[H]
\centering
\caption{Comparison of different parametrisation for our robust loss formulation $L^r$. Both GLU-Net-\dicor are trained on the \textit{Dynamic} dataset with three optimization iterations and evaluated with three and seven iterations for respectively the global-\dicor and the local-\dicor modules.}
\resizebox{0.99\textwidth}{!}{%
\begin{tabular}{lcc|cc|ccc|ccc}
\toprule
             & \multicolumn{2}{c}{\textbf{KITTI-2012}} & \multicolumn{2}{c}{\textbf{KITTI-2015}} & \multicolumn{3}{c}{\textbf{Sintel Clean}} & \multicolumn{3}{c}{\textbf{Sintel Final}}\\ 
 & AEPE    $\downarrow$          & F1   [\%]  $\downarrow$        & AEPE  $\downarrow$               & F1  [\%]  $\downarrow$  & AEPE  $\downarrow$   & PCK-1  [\%] $\uparrow$ & PCK-5  [\%] $\uparrow$ & AEPE $\downarrow$    & PCK-1  [\%] $\uparrow$ & PCK-5  [\%]  $\uparrow$\\ \midrule
GLU-Net  & 3.14 & 19.76 & 7.49 & 33.83 & 4.25 & 62.08 & 88.40 & 5.50 & 57.85 & 85.10 \\
GLU-Net-\dicor, $\eta=0$ & 2.68 & 15.43 & 6.68 & 27.57 & 3.80 & 67.12 & 90.41 & 4.90 & 63.38 & 87.69 \\
GLU-Net-\dicor, $\eta=0.1$ & 2.70 & 15.51 & 6.92 & 28.06 & 3.74 & 66.72 & 90.42 & 4.81 & 63.04 & 87.69 \\
\bottomrule
\end{tabular}%
}\vspace{1mm}
\label{tab:optical-flow-dualbent}
\end{table} 

\newpage
\begin{figure}[H]
\centering
(A) 
\includegraphics[width=0.99\textwidth]{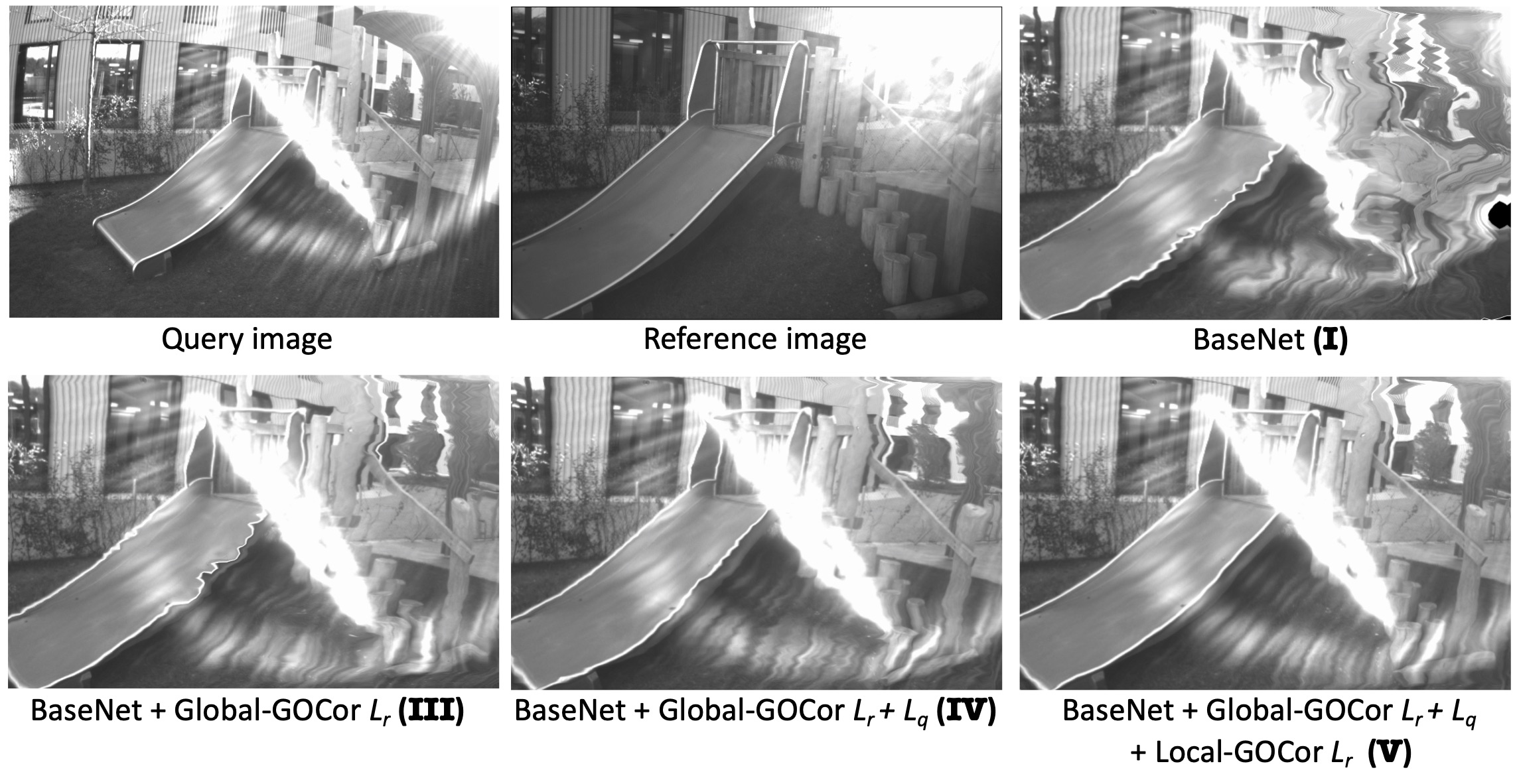}
(B)
\includegraphics[width=0.99\textwidth]{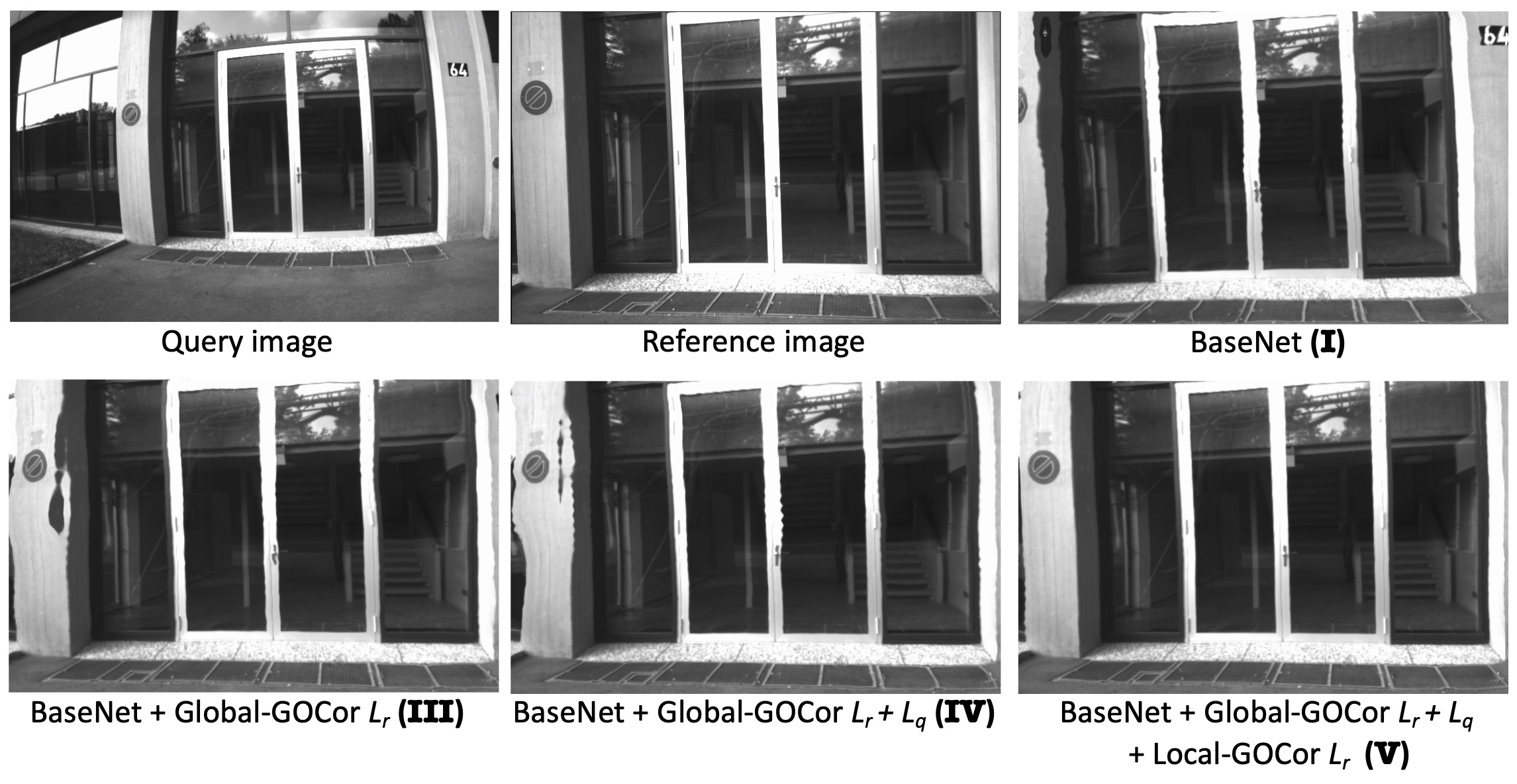}
\vspace{-3mm}\caption{Qualitative analysis of the different components of our approach, when the corresponding networks are applied to images of the ETH3D dataset. All models are trained on the \textit{Dynamic} dataset. We visualize the query images warped according to the flow fields estimated by the networks. The warped query images should resemble the reference images.}
\label{fig:ab-study-eth3d}
\end{figure}

\begin{figure}[H]
\centering
(A) KITTI-2015
\includegraphics[width=0.99\textwidth]{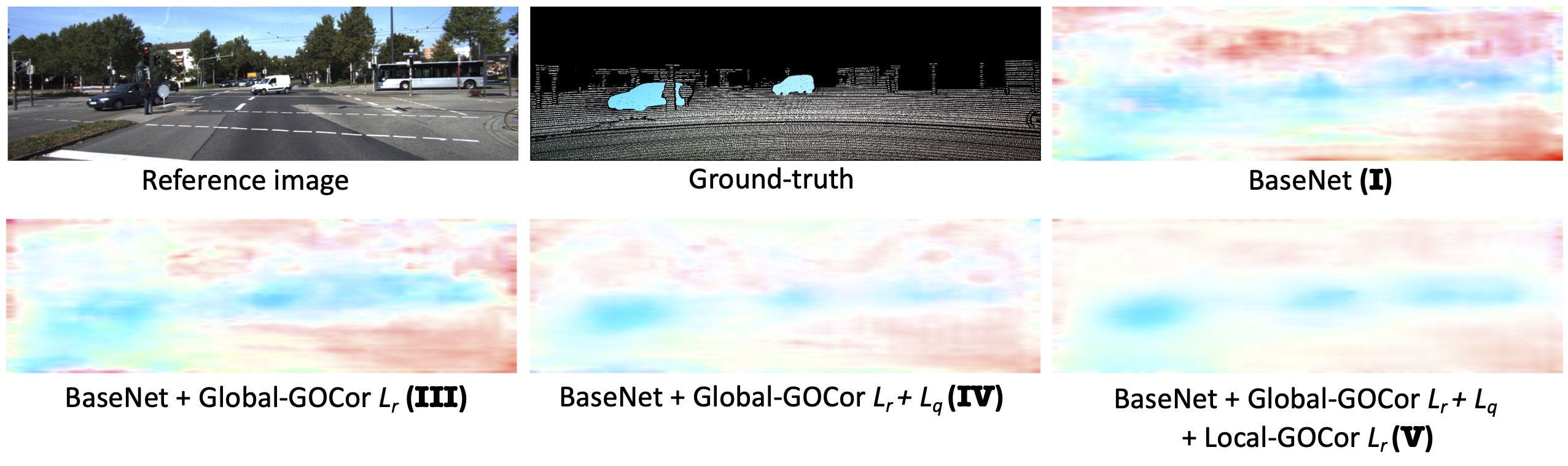}
(B) Sintel-clean
\includegraphics[width=0.99\textwidth]{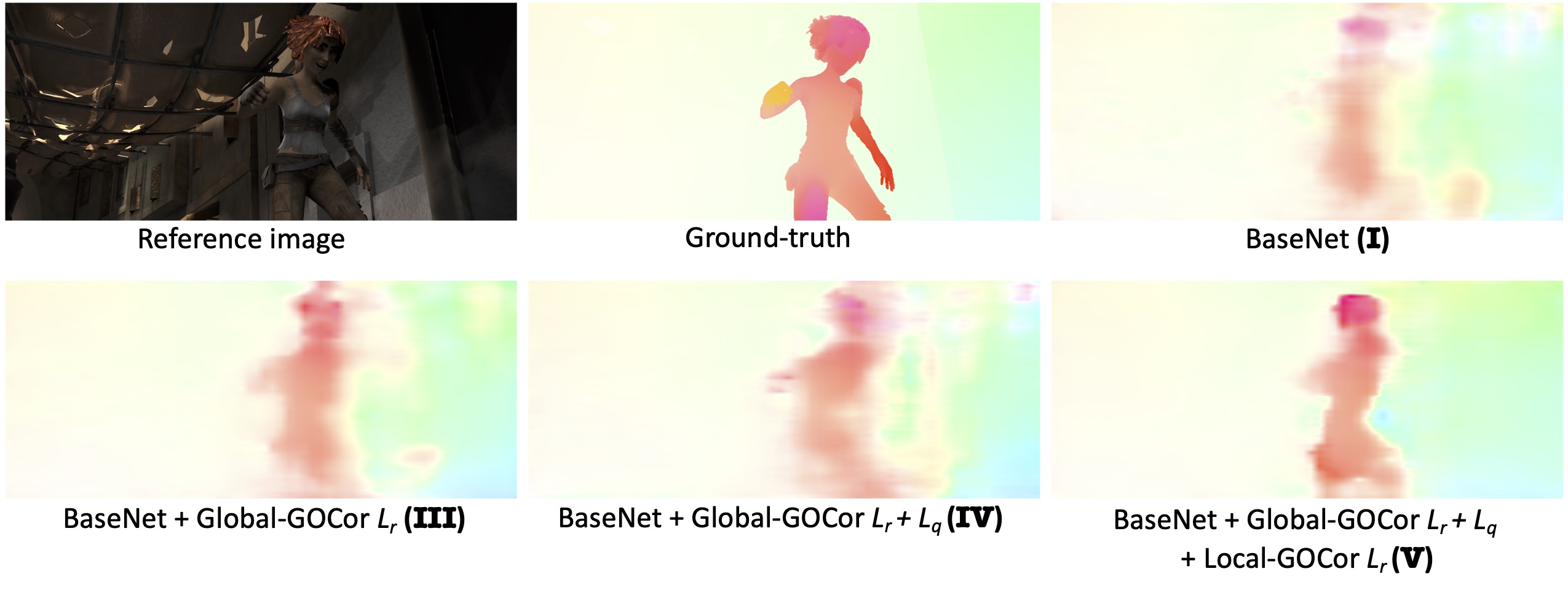}
\vspace{-3mm}\caption{Qualitative analysis of the different components of our approach, when the corresponding networks are applied to images of the Sintel dataset, clean pass. All models are trained on the \textit{Dynamic} dataset. We plot directly the estimated flow field for each image pair. }
\label{fig:ab-study-sintel}
\end{figure}

\section{Additional ablation study}
\label{Sec:sup-ablation}

For completeness, in Table \ref{tab:ablation-on-cad}, we provide a similar ablation study as that of the main paper, when BaseNet is trained on the \textit{Static} data, instead of the \textit{Dynamic} one. The same conclusions apply. 
We thus perform the additional ablative experiments when training all BaseNet variants on the \textit{Static} data. 

\parsection{Qualitative ablation study}  We visualize the quality of the estimated flow fields outputted by BaseNet (\textbf{I}), BaseNet + Global-\dicor $L_r$  (\textbf{III}), BaseNet + Global-\dicor $L_r + L_q$ (\textbf{IV}) and BaseNet + Global-\dicor $L_r + L_q$ + Local-\dicor $L_r$ (\textbf{V}) when applied to images of the ETH3D dataset and of optical flow datasets Sintel and KITTI in respectively Figures \ref{fig:ab-study-eth3d} and \ref{fig:ab-study-sintel}. 
Example (A) of Figure \ref{fig:ab-study-eth3d} shows the benefit of our global \dicor module as compared to the global feature correlation layer. Indeed, BaseNet does not manage to correctly capture the geometric transformation between the query and the reference images. On the other hand, thanks to our global \dicor module, BaseNet + Global-\dicor estimates a much more accurate transformation relating the frames. 
In example (B), we illustrate the gradual improvement from version (III) to (V). While introducing Global-\dicor with only the reference loss $L_r$ (version III) makes the estimated flow more stable than BaseNet, especially in the background, the representation of the slide object on the warped image is still shaky. Adding the query loss $L_q$ (version IV) smooths the estimated flow, and therefore the warped query image according to this flow. The later looks visually much better because of additional smoothness. 
Finally, further substituting the local feature correlation layers with our local \dicor module (version V) finishes to polish the result. The slide object in that case looks almost perfect and artifacts in the background are partially removed. 

The impact of the query frame objective $L_q$ in our global \dicor module is further illustrated in example (A) of Figure \ref{fig:ab-study-sintel}. Introducing $L_q$ enables to smooth the estimated flow field and to remove part of the artifacts. 
Finally, the advantage of our local \dicor module as opposed to the local feature correlation layer is visualized in both examples of Figure \ref{fig:ab-study-sintel}. From version (IV) to (V), the local \dicor module can recover sharper motion boundaries and the estimated flow is generally more accurate. Moreover, remaining artifacts in the background are removed.

\parsection{Comparison with post-processing method NC-Net} Here, we investigate the impact of post-processing method NC-Net \cite{Rocco2018b} and show comparisons to BaseNet and our approach in Table \ref{tab:ablation-on-cad}. When trained on the \textit{Static} dataset, including post-processing module NC-Net following the global correlation layer (version \textbf{II}) leads to better results than original version BaseNet (\textbf{I}) on the HPatches and the KITTI-2012 datasets. However, on the KITTI-2015 images, adding NC-Net results in worse performance. 
This is due to the fact that NC-Net uses correspondences with high confidences to support other uncertain neighboring matches. However, in the case of independently moving objects, neighboring matches can correspond to completely different motions, which breaks the assumption of the neighborhood consensus constraint used in NC-Net. 
Since it cannot cope with independently moving objects,  NC-Net obtains worse results on KITTI-2015, which depicts dynamic scenes. This is contrary to HPatches which present planar scenes with homographies and to KITTI-2012, which is restricted to static scenes. 
This observation is also emphasized by the ablation study in Table 3 of the main paper, where BaseNet + NC-Net is trained on the \textit{Dynamic} dataset. In that case, the performance of the resulting network is much worse than original BaseNet on all datasets. This is again due to the inability of NC-Net to handle moving objects, in that case, present in the training dataset. 
This shows the advantage of our method, which instead of applying 4D convolutions to post-process the correspondence volume, integrates them \textit{before} the correlation operation itself.

\parsection{Impact of objective function in the Local \dicor} In Table \ref{tab:ablation-on-cad}, we analyse the impact of both terms $L_r, L_q$ of our objective function $L$ when used in our Local \dicor module. Comparing versions \textbf{(V)} and \textbf{(VI)}, we found that adding the loss on the query frame $\Lq$ (Sec \ref{subsec:smooth-loss}) for Local-\dicor is harmful for its performance, particularly on the KITTI-datasets. Besides, adding the regularizer loss in the local \dicor level leads to longer training and inference run times. 
We therefore do not include it, our best version of BaseNet-\dicor resulting in \textbf{(V)}.

\begin{table}
\centering
\caption{Ablation study. All networks are trained on the \textit{Static} dataset. The \dicor modules are trained and evaluated with 3 steepest descent iterations.}
\vspace{-1mm}\resizebox{0.99\textwidth}{!}{%
\begin{tabular}{ll cc|cc|cc}
\toprule
           &  & \multicolumn{2}{c}{\textbf{HP}} & \multicolumn{2}{c}{\textbf{KITTI-2012}} & \multicolumn{2}{c}{\textbf{KITTI-2015}} \\ 
          &    & AEPE $\downarrow$     & PCK-5  [\%] $\uparrow$   & AEPE $\downarrow$               & F1    [\%]   $\downarrow$       & AEPE $\downarrow$                & F1  [\%]    $\downarrow$           \\ \midrule
I & BaseNet &  26.73 & 65.30 & 4.95 & 42.49 & 11.52 & 61.90 \\
II & BaseNet + NC-Net  & 24.59 & 66.62 & 5.00 & 39.80 & 12.44 & 62.96 \\
III & BaseNet + Global-\dicor $L_r$  &  22.80 & 70.00 & 4.43 & 34.81 & 10.93 & 55.73 \\
IV & BaseNet + Global-\dicor $L_r + L_q$  & 22.16 & 70.54 &  4.36 & 34.15 & 10.97 & 55.62 \\
V & BaseNet + Global-\dicor $L_r + L_q$ + Local-\dicor  $L_r$ & 22.00 & 74.80 &   \textbf{4.02} &   \textbf{31.24} &  \textbf{9.92} &  \textbf{50.54} \\
VI & BaseNet + Global-\dicor $L_r + L_q$ + Local-\dicor  $L_r+ L_q$  & \textbf{21.96} &  \textbf{75.26} &  4.24 & 33.43 & 10.20 & 53.53 \\
\bottomrule
\end{tabular}%
}\vspace{-5mm}
\label{tab:ablation-on-cad}
\end{table} 

\parsection{Impact of number of training optimization iterations} Here, we investigate the influence of the number of training steepest descent iterations. We train multiple BaseNet-\dicor networks, gradually increasing the number of training optimization iterations within the global and the local \dicor modules. All networks are trained on the \textit{Static} dataset. We evaluate all variants on the HPatches, KITTI and Sintel datasets and present the results in Table \ref{tab:nbr-iteration}. We use the same number of optimization iteration during evaluation as during training. We additionally measure the inference run-time of each network, computed as the average over the 194 KITTI-2012 images on an NVIDIA Titan X GPU. 

Training and evaluating with more steepest descent iterations consistently leads to better performances on all metrics and all datasets. It is also interesting to note that BaseNet + Global-\dicor without going through the optimizer (0 iteration) already outperforms the original BaseNet. In that case, the improvement is solely due to our powerful initialization $\wt^0$. 

Nevertheless, the improvement in performance when increasing the number of optimization iterations comes at the expense of inference and training time. As a result, we trained all our \dicor modules with three iterations which presented a satisfactory trade-off between inference time and accuracy. 
Besides, it must also be noted that for time-critical applications, using a single optimization iteration for both local and global \dicor modules already leads to significant improvements over the standard feature correlation layer.

\begin{table}
\centering
\caption{Analysis of the number of training optimization iterations. Both Local-\dicor and Global-\dicor layers are trained and evaluated with the same number of iterations. All networks are trained on the \textit{Static} dataset.}
\vspace{-1mm}\resizebox{0.99\textwidth}{!}{%
\begin{tabular}{lccc|ccc|ccc}
\toprule
             &  \multicolumn{3}{c}{\textbf{KITTI-2012}} & \multicolumn{3}{c}{\textbf{HP}} & \multicolumn{3}{c}{\textbf{Sintel-clean}} \\ 
             & Run-time [ms] & AEPE     $\downarrow$          & F1   [\%]  $\downarrow$   & AEPE  $\downarrow$    & PCK-1  [\%]$\uparrow$  & PCK-5  [\%] $\uparrow$  &  AEPE   $\downarrow$      & PCK-1  [\%]   $\uparrow$      & PCK-5  [\%]        $\uparrow$       \\ \midrule
BaseNet & 63.20  & 4.95 & 42.49 & 26.73 & 12.02 & 65.30 &  7.78 & 12.46 & 74.52 \\
BaseNet-\dicor, optim-iter = 0 & 66.80 & 4.87 & 37.42 & 26.47 & 16.40 & 66.42 &  6.94 & 27.63 & 77.20 \\
BaseNet-\dicor, optim-iter = 1  &  70.52 & 4.18 & 32.95 &  21.91 &  20.73 & 72.82 &  6.49 & 31.03 & 79.87 \\
BaseNet-\dicor, optim-iter = 3 & 82.42 &  4.02 &  31.24 & 22.00 & 23.68 & 74.80 & 6.32 & 33.81 & 80.72 \\
BaseNet-\dicor, optim-iter = 5 & 94.85 & 3.83 & 29.14 & 20.68 & 25.99 & 76.88 & 6.34 & 38.10 & 80.91 \\
\bottomrule
\end{tabular}%
}\vspace{1mm}
\label{tab:nbr-iteration}
\end{table} 

\parsection{Impact of performing global correlation with interchanging query and reference features}
We also experimented with interchanging the query and reference frames at the global level, and then fusing the two resulting GOCor correspondence volumes before passing them to the flow estimation decoder. However, we only observed marginal improvements. e.g., on KITTI-2015 it obtains an EPE of 11.07 and an F1 of 54.68\% compared to 10.97 EPE and 55.62\% F1 for the baseline BaseNet + Global-\dicor $L_r + L_q$ (IV). Considering that computing the GOCor correspondence volume twice increases the inference time of the model and the limited improvement of performance brought by fusing the two resulting correspondence volumes, we did not include this alternative in the final model.

\parsection{Impact of filter initializer $w^0$} As detailed in Sec. \ref{sec:sup-initializer}, we introduced several versions of our filter initializer module. Here, we train BaseNet with standard local feature correlation layers and our Global-\dicor module (using both our loss on the reference and on the query images) integrated in place of the global correlation layer. We experiment with different variants of the initializer and present the corresponding evaluation results in Table \ref{tab:initialization}. 
The version ZeroInitializer initializes $\wt^0$ to a zero tensor. Compared to all others, this initialization lacks accuracy for the final network, particularly on the optical flow datasets. 
In the SimpleInitializer versions which do not include global context information, adding more flexibility in the form of a learnt vector does not seem to help. However, in the case where context information is included, compared to ContextAwareInitializer, the Flexible variant significantly gains from increased flexibility. 
Initializer module Flexible-ContextAwareInitializer appears to be the best alternative for our Global-\dicor module. 

\begin{table}
\centering
\caption{Analysis of the impact of the initializer for the Global-\dicor. All networks are trained on the \textit{Static} with three optimization iterations. They are evaluated with the same number of iterations.}
\vspace{-1mm}\resizebox{0.99\textwidth}{!}{%
\begin{tabular}{lccc|cc|ccc}
\toprule
             & \multicolumn{3}{c}{\textbf{HP}} & \multicolumn{2}{c}{\textbf{KITTI-2012}} & \multicolumn{3}{c}{\textbf{Sintel-clean}} \\ 
             & AEPE $\downarrow$    & PCK-1  [\%] $\uparrow$ & PCK-5  [\%] $\uparrow$  & AEPE  $\downarrow$             & F1   [\%]  $\downarrow$          & AEPE  $\downarrow$       & PCK-1  [\%]    $\uparrow$    & PCK-5  [\%]  $\uparrow$            \\ \midrule
BaseNet + Global-\dicor, ZeroInitializer & 23.22 & 14.05 & 67.53 & 4.77 & 38.67 & 7.56 & 14.94 & 74.96  \\
BaseNet + Global-\dicor, SimpleInitializer  &  24.69 & 13.17 & 66.82  & 4.81 & 37.86 &  7.32 & 20.31 & 76.12 \\
BaseNet + Global-\dicor, Flexible-SimpleInitializer  & 25.53 & 12.82 & 66.97 & 4.83 & 38.71 & 7.30 & 20.48 & 76.24 \\
BaseNet + Global-\dicor, ContextAwareInitializer  & 25.13 & 15.01 & 67.14 & 4.87 & 39.14 & 7.43 & 19.19 & 75.60 \\
BaseNet + Global-\dicor, Flexible-ContextAwareInitializer & \textbf{22.16} & \textbf{17.03} & \textbf{70.54} &  \textbf{4.36} & \textbf{34.15} & \textbf{7.21} & \textbf{20.68} & \textbf{77.09} \\
\bottomrule
\end{tabular}%
}\vspace{-2mm}
\label{tab:initialization}
\end{table} 

\end{document}